\documentclass[journal]{journal}

\usepackage{graphicx}
\usepackage{caption}
\usepackage{subcaption}
\usepackage[hidelinks]{hyperref}
\usepackage{amsfonts}
\begin{document}
%
\title{Free-riders in Federated Learning: Attacks and Defenses}
\author{Jierui Lin,
        Min Du,
        and~Jian Liu\\University of California, Berkeley

\thanks{J. Lin, M. Du and J. Liu are with the Department
of Electrical Engineering and Computer Science, University of California, Berkeley,
CA, 94704, USA e-mail: jerrylin0928, min.du@berkeley.edu, jian.liu@eecs.berkeley.edu.}}

\maketitle
\thispagestyle{empty}

\begin{abstract}
Federated learning is a recently proposed paradigm that enables multiple clients to collaboratively train a joint model. It allows clients to train models locally, and leverages the parameter server to generate a global model by aggregating the locally submitted gradient updates at each round.
Although the incentive model for federated learning has not been fully developed, it is supposed that participants are able to get rewards or the privilege to use the final global model, as a compensation for taking efforts to train the model.
Therefore, a client who does not have any local data has the incentive to construct local gradient updates in order to deceive for rewards.
In this paper, we are the first to propose the notion of free rider attacks, to explore possible ways that an attacker may construct gradient updates, without any local training data.
Furthermore, we explore possible defenses that could detect the proposed attacks, and propose a new high dimensional detection method called STD-DAGMM, which particularly works well for anomaly detection of model parameters. We extend the attacks and defenses to consider more free riders as well as differential privacy, which sheds light on and calls for future research in this field.
\end{abstract}


%
\IEEEpeerreviewmaketitle

\section{Introduction}

\IEEEPARstart{F}{ederated} learning~\cite{DBLP:journals/corr/McMahanMRA16,DBLP:journals/corr/KonecnyMRR16,7447103} has been proposed to facilitate a joint model training leveraging data from multiple clients, where the training process is coordinated by a parameter server. In the whole process, clients' data stay local, and only model parameters are communicated among clients through the parameter server. A typical training iteration works as follows. First, the parameter server sends the newest global model to each client. Then, each client locally updates the model using local data and reports updated gradients to the parameter server. Finally, the server performs model aggregation on all submitted local updates to form a new global model, 
which has better performance than models trained using any single client's data.

Compared with an alternative approach which simply collects all data from the clients and trains a model on those data, federated learning is able to save the communication overhead by only transmitting model parameters, as well as protect privacy since all data stay local. As a result, federated learning has attracted much attention and is being widely adopted for model training utilizing data from multiple users and/or organizations.

Federated learning has many practical applications, such as next-word prediction in a virtual keyboard for smartphones~\cite{DBLP:journals/corr/abs-1811-03604}, prediction of customers' usage of IoT devices ~\cite{2019arXiv190610893Z} and classification of hospitalizations for cardiac events~\cite{article}, to achieve a well-trained model. 
Although the profit model~\cite{Weng2018DeepChainAA} for federated learning has not been fully developed, one would assume that a client who contributes data to the model training process could either get money or the privilege to use the final trained model as a reward. However, a question raised by this is: \textit{how could one tell whether the model updates uploaded by a client are actually trained using local data, or artificially constructed weights which adds no value to the model?}

{\em Free-riders} are generally referred as some individuals who benefit from resources, public goods, or services of a communal nature, but do not pay for them~\cite{Baumol2004}. 
The free-riders problem has been extensively studied in peer-to-peer systems~\cite{cs5696, Feldman:2005:OFB:1120717.1120723, 1626427}.
The aim of this paper is to set out the research of free-riders in federated learning.
In a federated learning setting where each contributing client could get a reward, there may exist clients that pretend to be contributing to trick rewards. We refer such clients as free-riders, and the procedure of generating fake weights to report to the parameter server as \textit{free-rider attacks}. 
There could be two main incentives for free riders to submit fake updates. For one, a client may not have the required data, or is concerned about data privacy, so that local data are not available for model training. For another, a client may want to save local CPU cycles or other computing resources.
Note that in the first case, where local data are not available but a client is willing to consume computing resources, the client may try to submit gradients trained with another dataset in order to evade detection. However, the central server could easily catch such a free rider by verifying the model accuracy on a holdout validation dataset. Thus, in both cases, free riders will have to fake their weight submissions, through clever manipulations of the information provided by the central server.
With that, we explore a set of free-rider attacks and propose 
a powerful defense mechanism - STD-DAGMM, that is able to detect most free riders and prevent them from getting the final model and monetary rewards.
We leave it as an open question to explore more sophisticated attacks and defences.

We start from a simple attack where the free-rider simply uses randomly generated values as the gradient updates. 
We empirically show that this attack can evade a popular detection approach, i.e., Autoencoder~\cite{10.1007/978-3-642-82657-3_24,Bourlard1988}, but be detected by a recently proposed high-dimensional anomaly detection method - DAGMM~\cite{zong2018deep}.
Then, we propose a new attack named {\em delta weights attack}, which generates gradient updates by subtracting the two global models received in previous two rounds. We show that such an attack is able to evade DAGMM detection.
To defend against such attacks, we propose a STD-DAGMM, which incorporates the standard deviation metric of a local gradient update matrix with DAGMM.
We extensively evaluate STD-DAGMM under stronger attacks and with varying real-world settings.
We believe our findings could be inspiring for the research in federated learning, and initiates a new field of free rider detection in such an environment. 
Our contributions are summarized as follows.



\begin{enumerate}
    \item As far as we know, we are the first to explore free-rider attacks in federated learning, and propose possible defenses against such attacks.
    \item We explore the best attack a free rider may exploit, and propose a novel high-dimensional anomaly detection approach, STD-DAGMM, that is particularly useful for free rider detection. The proposed method is potentially useful for other model weights anomaly detection as well.
    \item We extend our findings under more real world settings, such as in the face of differential privacy. We observe that differential privacy in federated learning could potentially mitigate the effectiveness of the proposed attacks and make them more easily to be detected.
\end{enumerate}

\section{Preliminary}

\subsection{Federated learning}
\label{sec:pre-fed}
In a typical federated learning setting, 
a central parameter server coordinates a network of nodes, e.g. smartphones, tablets and laptops, to train a global model utilizing data from all the participants. 
Suppose there are $n$ clients $C_1,$ $C_2,$ $...,$ $C_n$ collaboratively training a global model $M$ in the setting of federated learning. 
In the $j$-th round, each client $C_i$ receives the newest global model $M_{j}$ from the parameter server, and produces local gradient $G_{i,j}$ by training the model using its local data; the parameter server aggregates each $G_{i,j}$ to be a new global model, using~\cite{DBLP:journals/corr/McMahanMRA16}
\begin{equation}
\label{eq:global-model}
M_{j+1} = M_{j} - \eta \cdot \frac{1}{n} \sum_{i=1}^n G_{i,j}
\end{equation}
where $\eta$ is a global learning rate applied by the parameter server, to control learning speed etc. In a simple case, $\eta=1$, where for each round, a new global model is generated by simply subtracting the average of all local gradient updates from the previous global model.


\subsection{DAGMM}
\label{sec:dagmm}
DAGMM~\cite{zong2018deep}, short for Deep Autoencoding Gaussian Mixture Model, uses a deep Autoencoder to generate a low-dimensional representation, as well as calculate the reconstruction error for each input data point. It further concatenates the low-dimensional embedding with the distances between input and reconstructed output by Autoencoder, including Euclidean distance and Cosine distance. The concatenated vector is fed into a Gaussian mixture model (GMM), which adopts an estimation network and trains the parameters to make the input fit the Gaussian mixture model. DAGMM jointly optimizes the parameters for deep Autoencoder and the Gaussian mixture model in an end-to-end fashion, with the goal of minimizing Autoencoder input-output error and fitting the GMM model in the same time.
According to~\cite{zong2018deep},
the end-to-end training is beneficial to density estimation, 
since we can have more freedom to adjust the dimension of encoded vector to favor the subsequent density estimation tasks. To perform anomaly detection, DAGMM takes mini batches of high dimensional vectors as an inputs, trains to optimize the model parameters, and finally outputs an energy value for each vector.
The vectors having energy values higher than the majority 
are considered as suspicious.
\section{Problem setting}
\label{sec:problem}

For federated learning, we adopt the notions and symbols presented in Section~\ref{sec:pre-fed}.
In the $j$-th round, a client $C_i$ receives a global model $M_j$, and submits a gradient update $G_{i,j}$.

\paragraph{Free-riders.} We define a {\em free-rider} as a client (say $C_i$) who fakes $G_{i,j}$ and pretends it to be generated through local data training. $C_i$ may not have access to local data that could be used to 
train the model.
Rather, $C_i$ simply generates $G_{i,j}$ using the available knowledge (e.g., global model architecture, matrix values of each received global model), with a goal of fooling the parameter server, in order to enjoy the benefits of participating in global model training. 
We use $C_i^f$ to denote that a client $C_i$ is a free rider.

\paragraph{Assumptions.}
We restrain our problem as: if a client is a free-rider, it will be a free rider in all rounds, rather than being free-riders in some rounds while not in others. The intuition is that if a client has no local data, it cannot train a real model in any round. Also, free riders are independent that they can't collude to evade detection. 
We further assume that the parameter server does not keep a history of local clients updates. With this, to detect if a client is a free rider in round $j$, one can only use the updates available at round $i$, e.g., [$C_{1,j}$, $C_{2,j}$, $C_{3,j}$, ...], without having access to the history. 

\paragraph{Detection goals.} 
In this problem, there are multiple series of local gradient updates, e.g., [$G_{1,1},$ $G_{1,2},$ $...,$ $G_{1,t}$], [$G_{2,1},$ $G_{2,2},$ $...,$ $G_{2,t}],$ $...,$ [$G_{n,1},$ $G_{n,2},$ $...,$ $G_{n,t}$], and a series of global models [$M_1,$ $M_2,$ $...,$ $M_t$]. The goal of free-rider detection is to detect gradient series [$G_{i,1},$ $G_{i,2},$ $...,$ $G_{i,t}$], 
where all gradients $G_{i,j}$ are faked instead of being updated through local data.




\section{Attack I: Random Weights}
In this section, we present a basic attack that a naive attacker may exploit, and a possible detection mechanism for this attack.

\subsection{Attack specification}
\label{sec:random-attack}
For the federated learning process, in the $j$-th round, a free-rider client $C_i^f$ receives the newest global model $M_{j}$. The gradient update $G_{i,j}$ to submit should have the same dimension with global model $M_{j}$, otherwise would be rejected by the parameter server immediately. 
A straightforward method to construct $G_{i,j}$, is to simply copy $M_{j}$ as $G_{i,j}$, and replace each value in $G_{i,j}$ with 0, which however, would be trivial to detect. 
A better way is to replace each value in $G_{i,j}$ with a randomly generated value, 
which we detail below.


\paragraph{Random weights attack.}
For this attack, a free rider will attempt to construct a gradient update matrix that has the same dimension with the received global model, through randomly sampling each value from a uniform distribution within range $[-R, R]$. The attacker may have previous model training experience such that $R$ is close to mimic other normal clients' updates.
For this attack, the attacker may only vary one parameter, which is listed as below.

\textit{Random weights range $R$.}
This parameter decides the maximum and minimum range that a free rider may draw a random value from. Since an attacker may have previous model training experience, we may assume this range could be arbitrarily close to the real case.


\subsection{Defense strategy}
\label{sec:random-defense}
Our defense strategy in this section would also focus on existing anomaly detection methods that a parameter server may explore. Firstly, we show that this naively generated attack is possible to evade 
probably
the most popular deep learning based outlier detection approach
- Autoencoder anomaly detection. Secondly, we show that a recent work dedicated for high-dimensional anomaly detection called DAGMM, as presented in Section~\ref{sec:dagmm}, is able to detect such attacks.

\paragraph{Autoencoder detection.}
Autoencoder~\cite{10.1007/978-3-642-82657-3_24,Bourlard1988} is a deep neural network model architecture that has been widely adopted for outlier detection~\cite{Sakurada:2014:ADU:2689746.2689747,DBLP:journals/corr/abs-1811-05269}. It contains an encoder which aims to reduce the input data dimension, and a decoder which attempts to reconstruct the input data. Given a data sample, the learning objective is to minimize the error between an input and its corresponding output. Since the dimensionality reduction inevitably brings information loss, an Autoencoder model which is trained to minimize the average loss of the entire training data, is trained to preserve the most common information. As a result, outliers would have larger input-output error compared with others, since they may contain patterns that are not learned by Autoencoder. Therefore, we could do outlier detection by measuring the reconstruction error for each data sample.
To use Autoencoder for free rider detection, we first concatenate all rows of a client update matrix (e.g., having dimension $m\times n$) to be a single vector (e.g., having dimension $1 \times mn$). Then, all clients' update vectors are fed into Autoencoder in mini-batches, to train the Autoencoder with a learning objective to minimize the average reconstruction error for each mini-batch.

\paragraph{DAGMM detection.}
As introduced in Section~\ref{sec:dagmm}, DAGMM is a recently proposed anomaly detection method that is particularly useful for high-dimensional data. 
Note that deep learning model weight matrices could easily contain millions of parameters. 
For a simple 2-layer fully connected network we use for MNIST digit classification, the parameter size is close to 
0.2 million.
Thus, we believe DAGMM could be a more suitable model for our purpose due to the high dimensional nature of weight matrices.
To use DAGMM for free rider detection, we do the same thing for each client updates as for Autoencoder - concatenate all rows in a weight matrix to construct one single vector, and then feed all client update vectors into the DAGMM model in mini-batches.

\begin{figure}[hbtp]
     \centering
         \includegraphics[width=\linewidth]{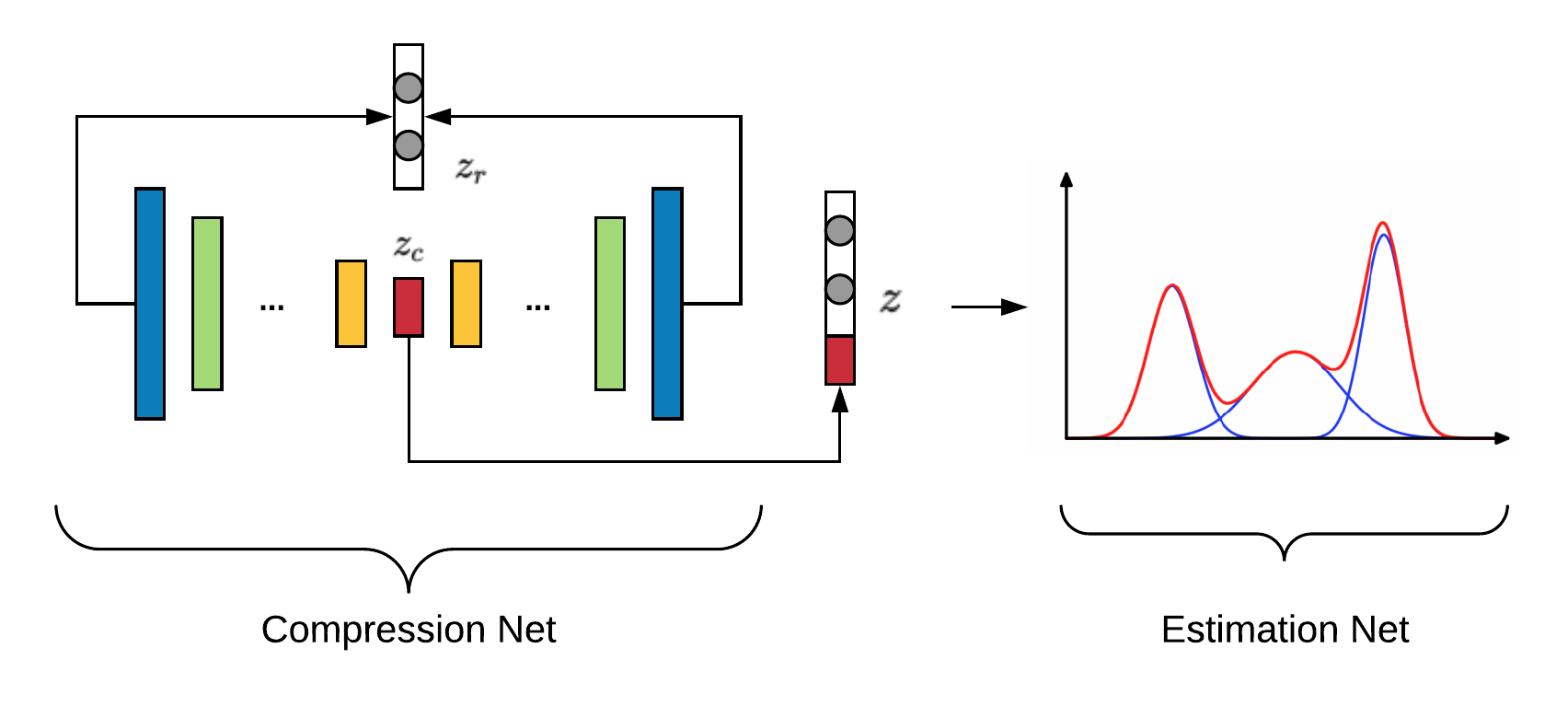}
 \caption{DAGMM model architecture}
         \label{fig:dagmm}
\end{figure}


\subsection{Experimental validation}
\label{sec:random-eval}
Without loss of generality, we assume there are 100 clients in total, and only 1 free rider among all clients. Extension on more free riders will be presented in Section~\ref{sec:more-free-riders}.
In this set of experiments, we use MNIST dataset~\cite{deng2012mnist}, and explore two cases in terms of whether each client has similar data distribution. 
Note that the clients having similar data are expected to have similar gradient updates as well. In this case, the free rider could be easier to detect since the randomly constructed weight matrix is different from the majority. On the other hand, if each client may have different data with others, a free-rider could be harder to detect, since all other clients may have different gradient updates after training on local data.

We explore both cases in terms of 
different $R$ for the random selection range $[-R, R]$. 
Also, note that 
the performance of the detection method is highly dependent on the training period,
i.e., a free rider is easier to be detected when the model is (close to) converged, than at the beginning of a training period.
Although early detection is preferred, we also show the detection performance when model is close to converged, as an additional reference.

\subsubsection{Each client has similar local data distribution.} 
To mimic the case where each client has similar local data distribution with others, we randomly distribute the MNIST dataset to each client. Therefore, we can assume that each client possesses all 10 classes in MNIST, and different clients have similar local datasets.
As explained in Section~\ref{sec:random-attack}, an attacker may have previous training knowledge to choose range $R$, to make the constructed weights close to other clients' gradient updates.

\begin{figure}[hbtp]
     \centering
     \begin{subfigure}[b]{\linewidth}
         \centering
         \includegraphics[width=\textwidth]{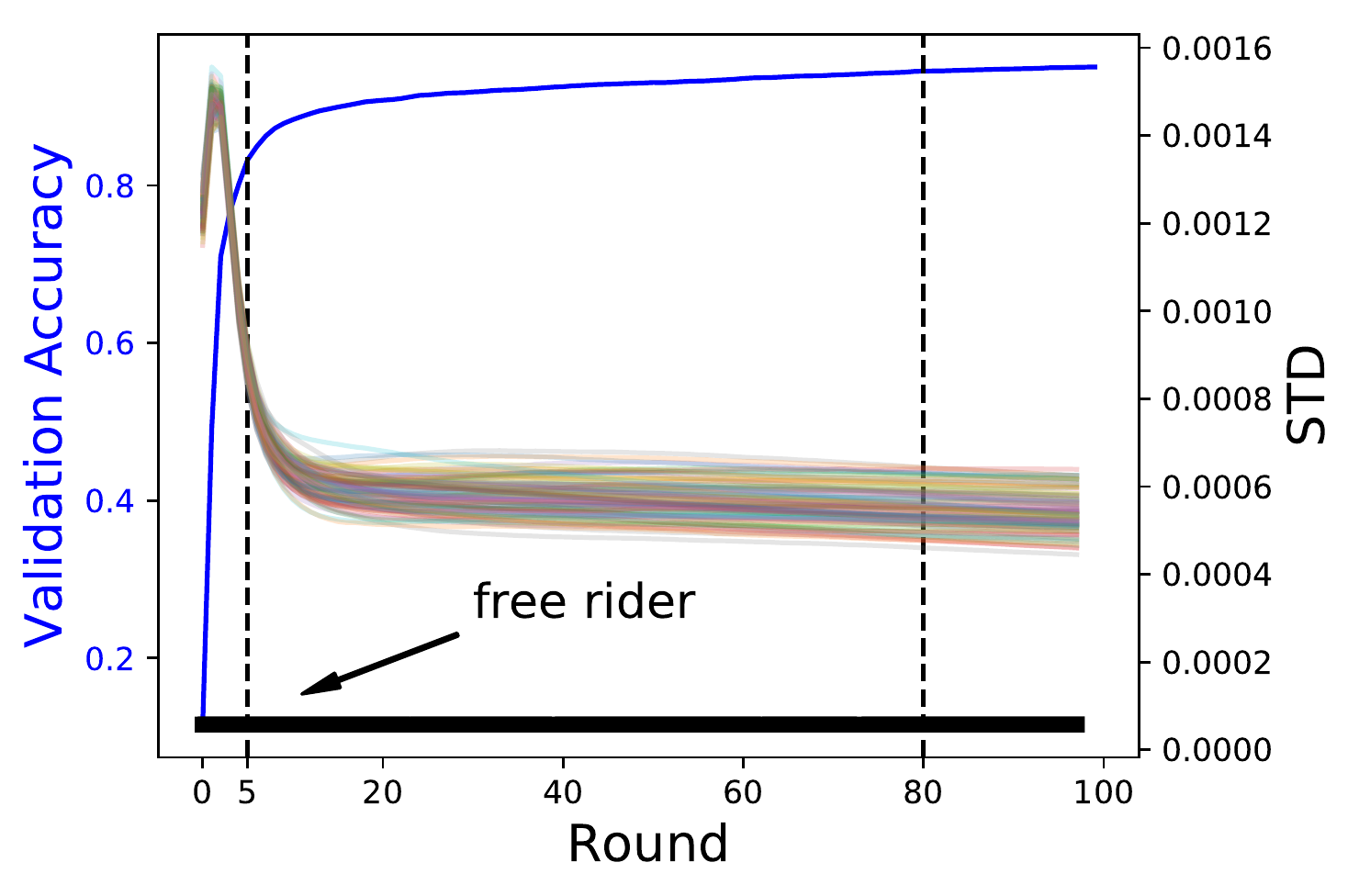}
         \caption{Standard deviation (STD) of the updates submitted by each client, with the change of training process}
         \label{fig:random10e-4-even-std}
     \end{subfigure}
     \begin{subfigure}[b]{\linewidth}
         \centering
         \begin{subfigure}[b]{0.49\linewidth}
             \includegraphics[width=\linewidth]{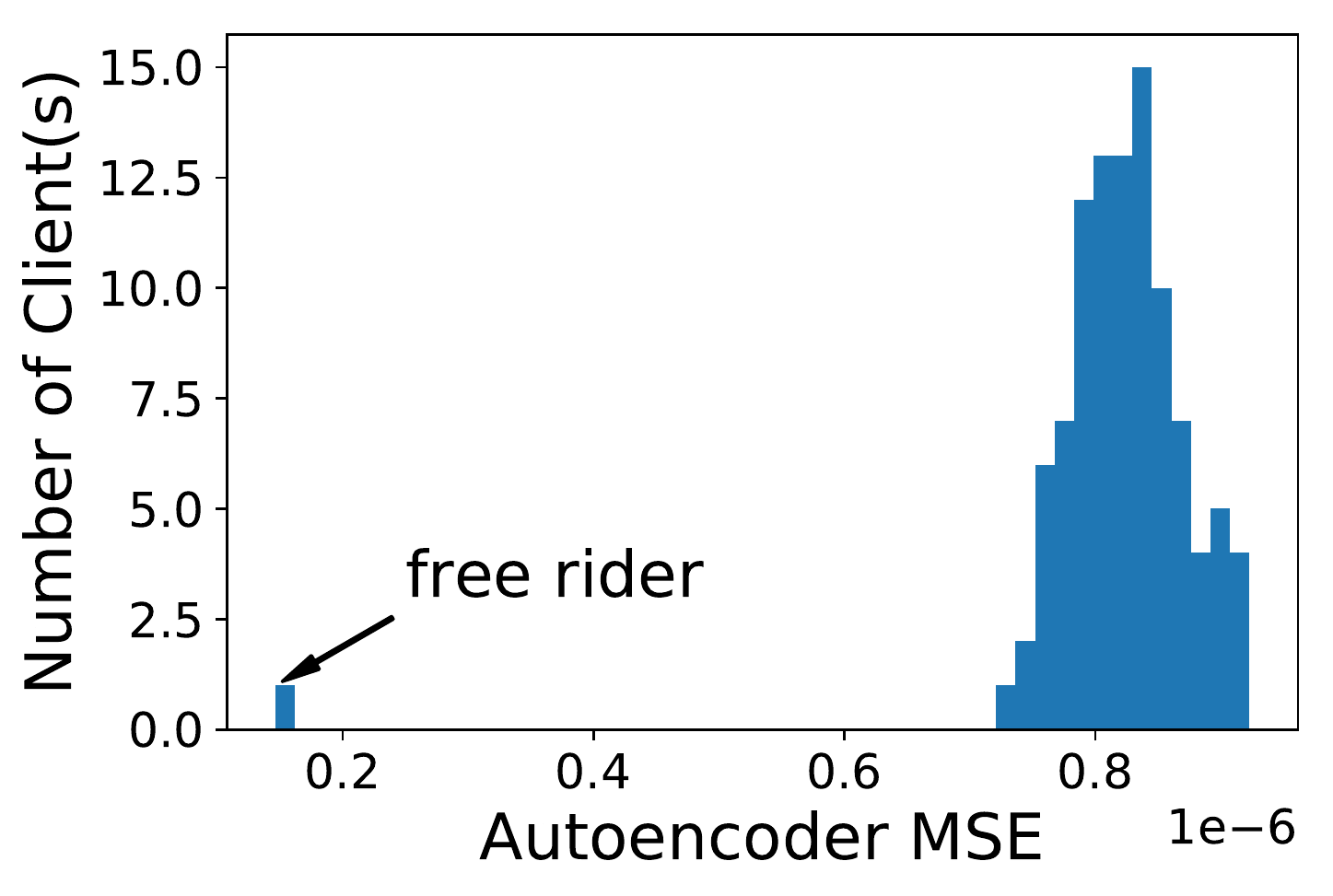}\hfill%
             \caption{Round 5}
         \end{subfigure}
         \begin{subfigure}[b]{0.49\linewidth}
             \includegraphics[width=\linewidth]{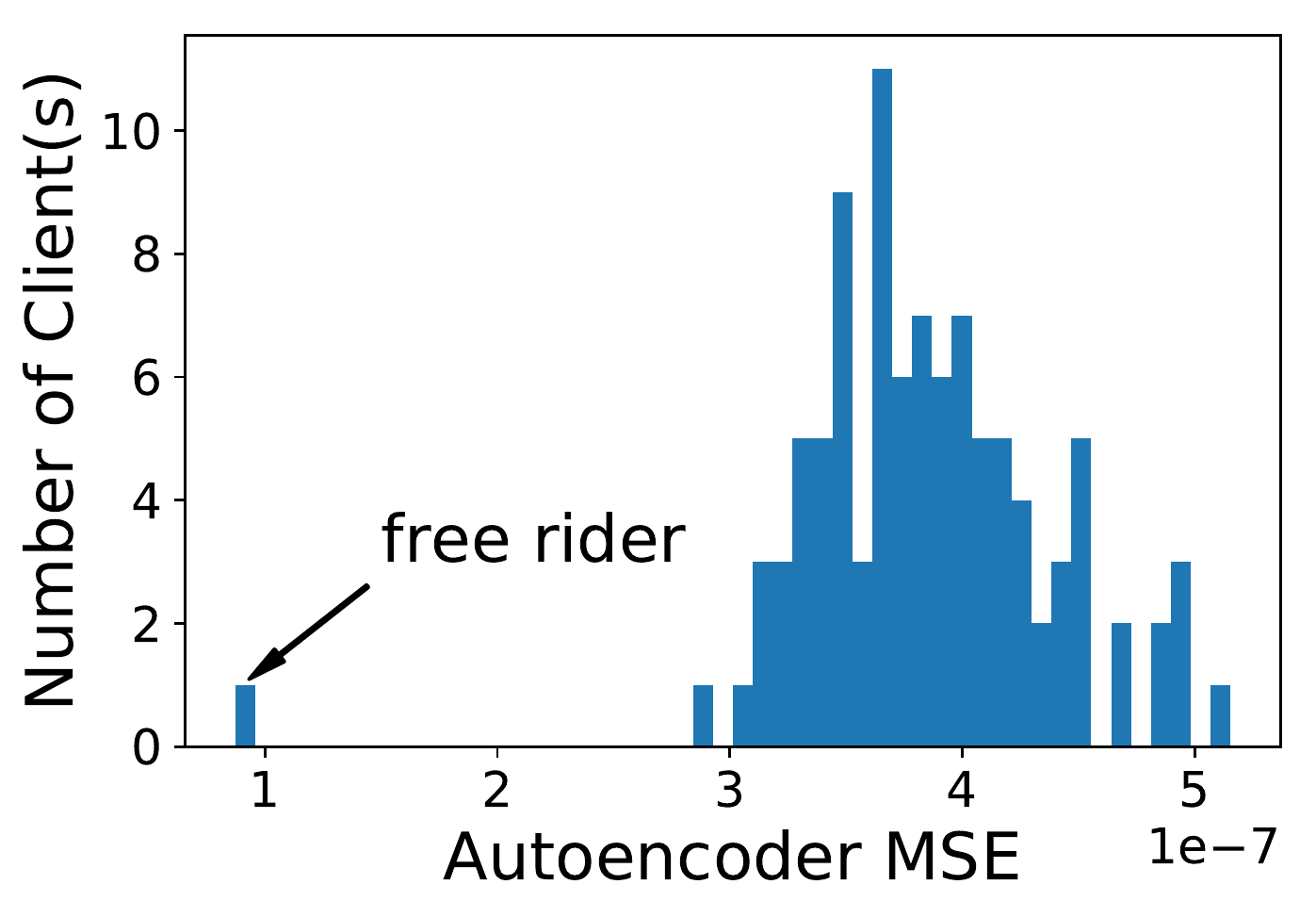}
             \caption{Round 80}
         \end{subfigure}
         \caption{Autoencoder - not detected}
         \label{fig:random10e-4-even-autoencoder}
     \end{subfigure}
     \begin{subfigure}[b]{0.48\textwidth}
         \centering
         \begin{subfigure}[b]{0.49\linewidth}
             \includegraphics[width=\linewidth]{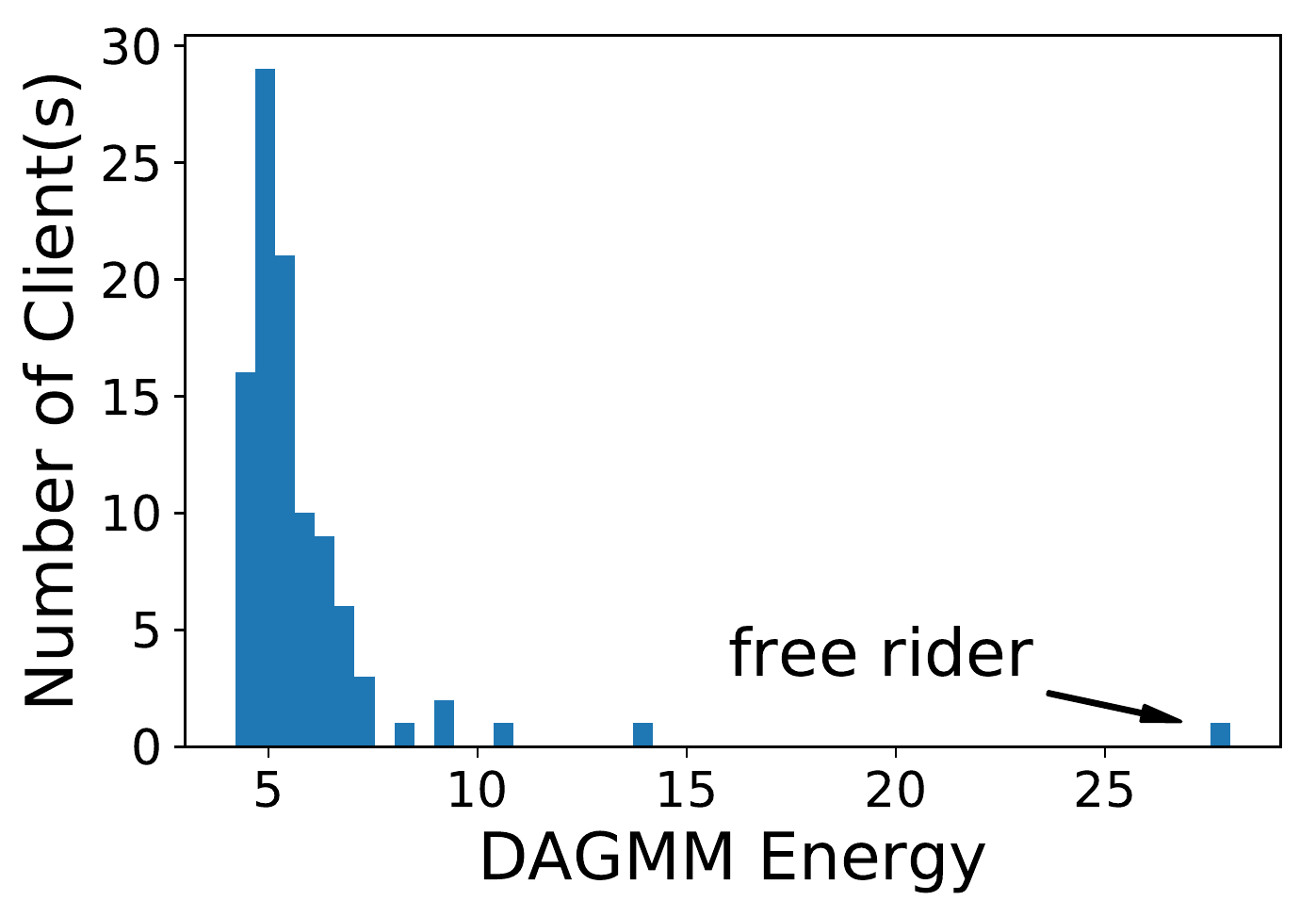}\hfill%
             \caption{Round 5}
         \end{subfigure}
         \begin{subfigure}[b]{0.49\linewidth}
             \includegraphics[width=\linewidth]{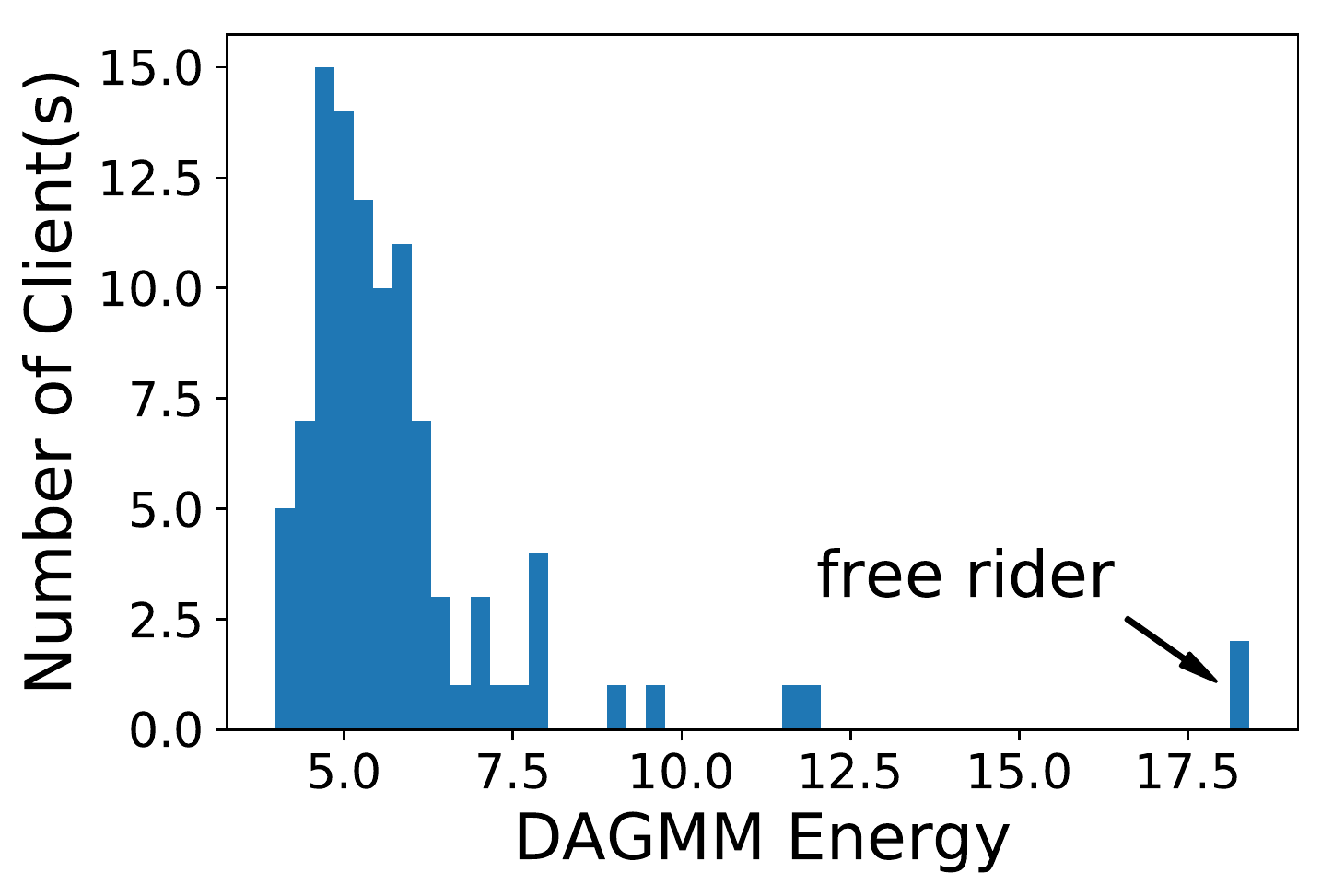}
             \caption{Round 80}
         \end{subfigure}
         \caption{DAGMM - detected}
         \label{fig:random10e-4-even-dagmm}
     \end{subfigure}
        \caption{Similar local data distribution, randomly sampled updates from [$-10^{-4}$, $10^{-4}$]}
        \label{fig:random10e-4-even}
\end{figure}

\paragraph{Random weights range $\mathbf{R=10^{-4}}$.}
We first consider a case where $R$ is not properly selected.
In this experiment, for the gradient updates $G_{i,j}$ submitted by $C_i^f$, each value is randomly sampled from a uniform distribution within range [$-10^{-4}$, $10^{-4}$]. 
Figure~\ref{fig:random10e-4-even-std} shows the standard deviation (STD) of each client's local gradient updates, with the increase of rounds in federated learning. The thicker, black line indicates the STD of the free rider, while lighter lines having various other colors indicate all other 99 clients' STD statistics.
We pick two rounds for free rider detection: round 5 which represents the beginning of federated learning process, and round 80 which represents the case when the model is almost converged,
which
are indicated in Figure~\ref{fig:random10e-4-even-std} as two vertical dashed lines.

Figure~\ref{fig:random10e-4-even-autoencoder} shows the free rider detection result using Autoencoder. 
Presumably, Autoencoder should be able to learn some important features from the similar weights submitted by benign clients and thus, produce a higher reconstruction error
for the free rider. However, surprisingly, as in Figure~\ref{fig:random10e-4-even-autoencoder}, Autoencoder has the smallest error for the free rider among all clients, and thus fails to detect such attack.
Based on
the statistics in Figure~\ref{fig:random10e-4-even-std}, our assumption is that the free rider update has much smaller standard deviation compared with other clients' updates, which could be easier to memorize for Autoencoder, producing a small reconstruction error.


On the other hand, DAGMM 
is able to detect such attack, as shown in Figure~\ref{fig:random10e-4-even-dagmm}.
Besides the reason that DAGMM considers the reduced dimensional embeddings, we think that the Consine distance DAGMM takes into account also helps to identify free riders, since its input-output angle could be different from those of normal clients.
After DAGMM, the free rider has the highest energy value, and a wide range of thresholds suffice to separate the free rider from all other clients.

\begin{figure}[hbtp]
     \centering
     \begin{subfigure}[b]{\linewidth}
         \centering
         \includegraphics[width=\textwidth]{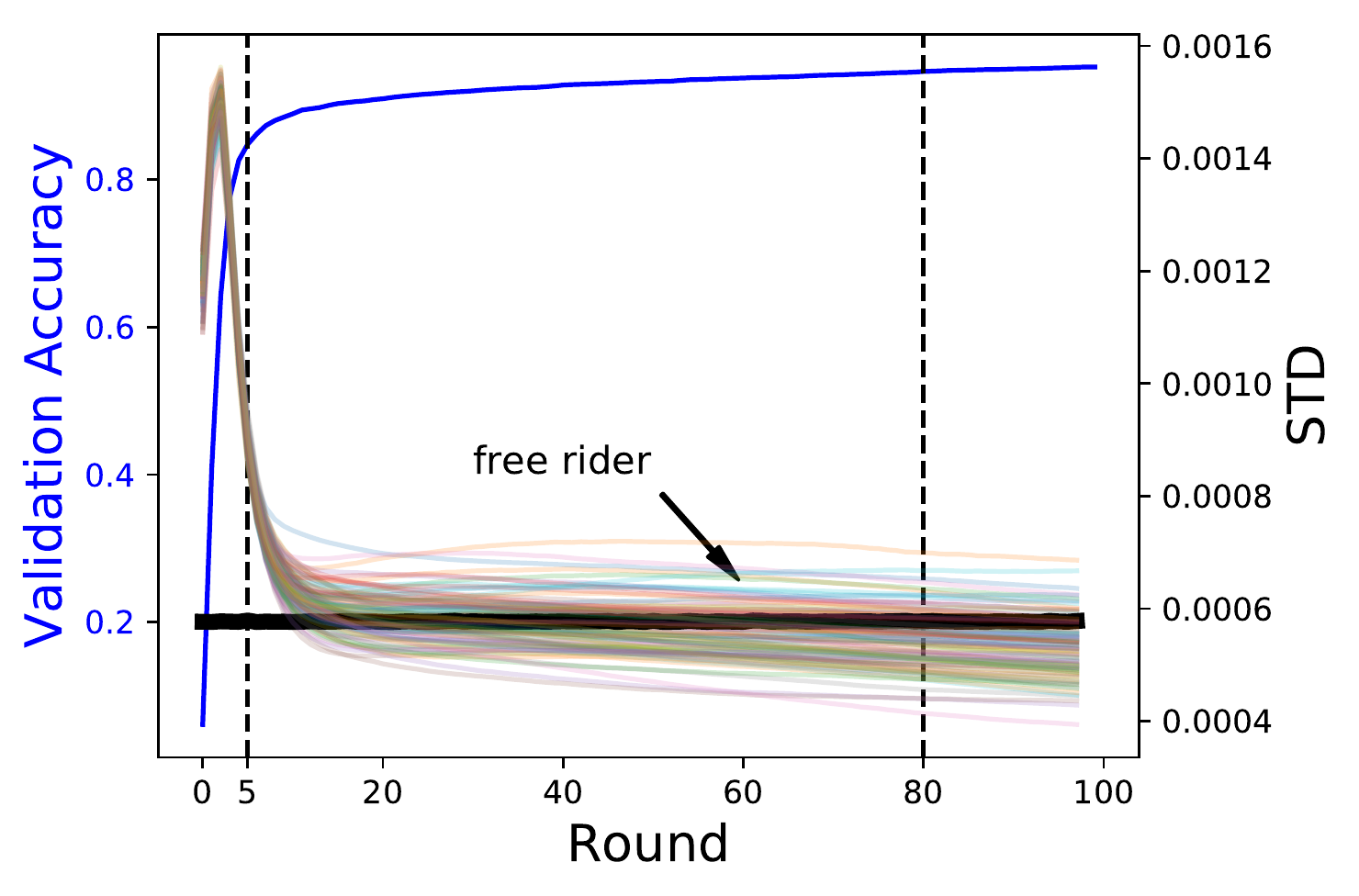}
         \caption{Standard deviation (STD) of the updates submitted by each client, with the change of training process}
         \label{fig:random10e-3-even-std}
     \end{subfigure}
     \begin{subfigure}[b]{\linewidth}
         \centering
         \begin{subfigure}[b]{.48\textwidth}
             \includegraphics[width=\linewidth]{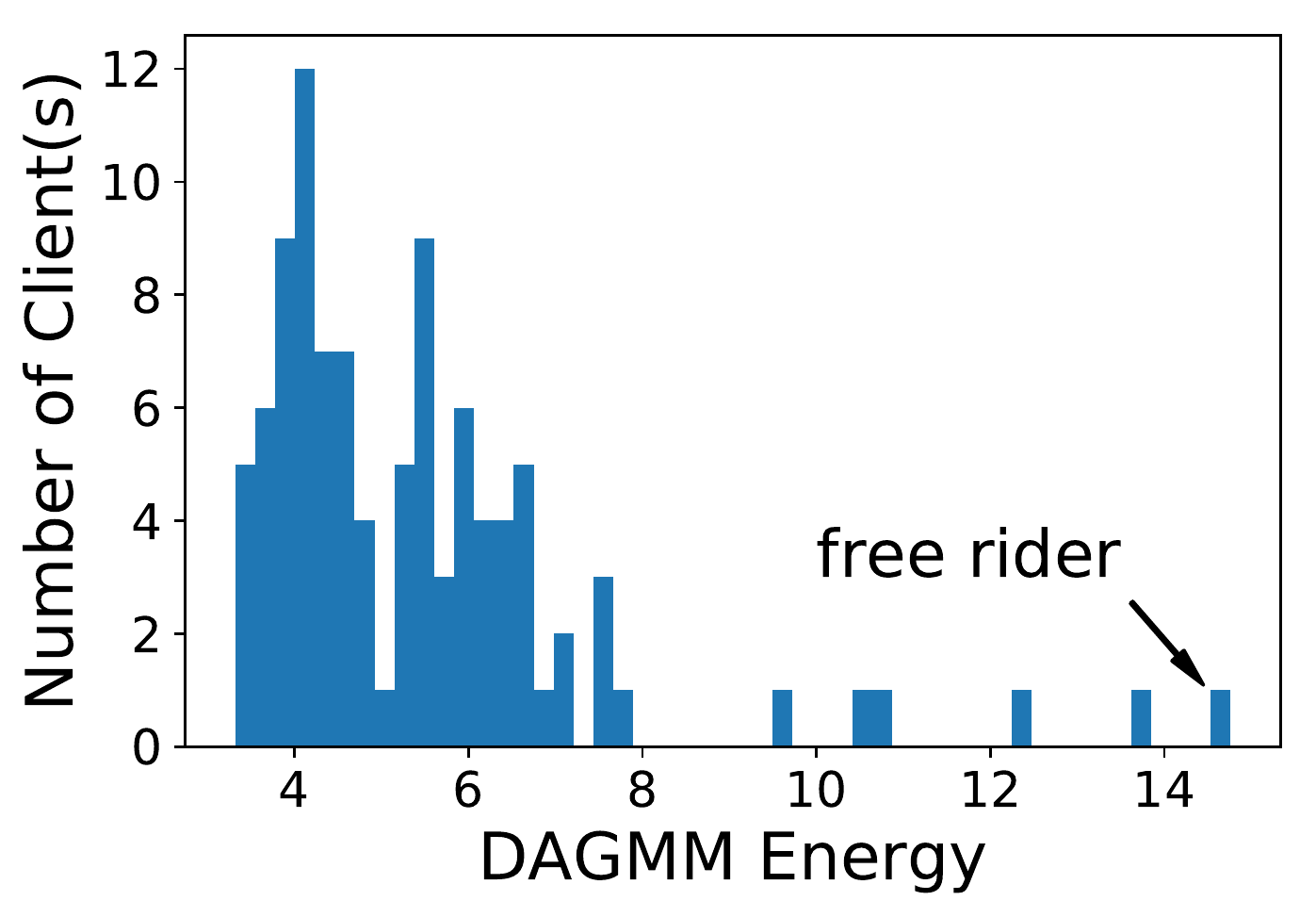}\hfill%
             \caption{Round 5}
         \end{subfigure}
         \begin{subfigure}[b]{.48\textwidth}
             \includegraphics[width=\linewidth]{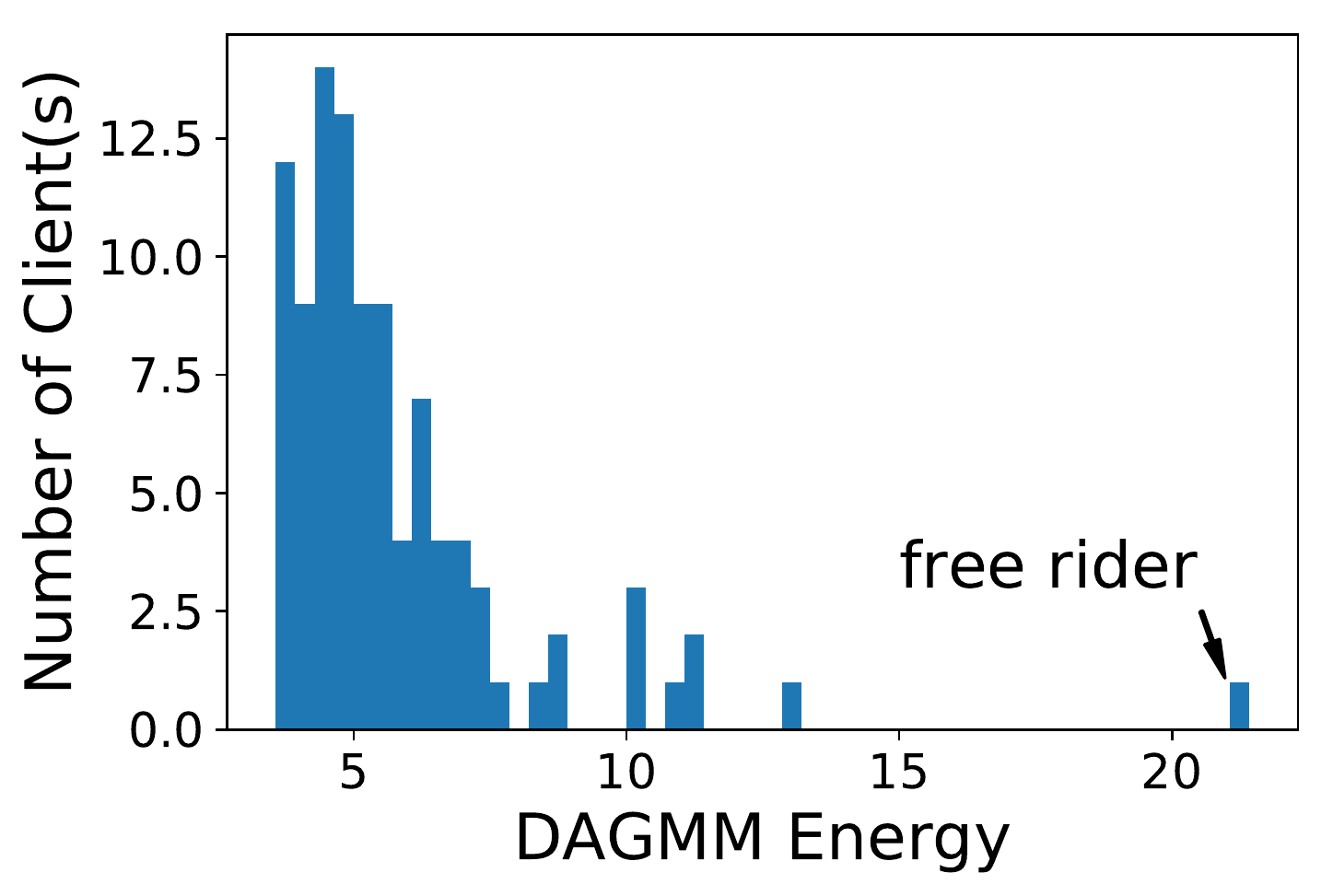}
             \caption{Round 80}
         \end{subfigure}
         \caption{DAGMM - detected}
         \label{fig:random10e-3-even-dagmm}
     \end{subfigure}
        \caption{Similar local data distribution, randomly sampled updates from [$-10^{-3}$, $10^{-3}$]}
        \label{fig:random10e-3-even}
\end{figure}

\paragraph{Random weights range $\mathbf{R=10^{-3}}$.}
In this part, we explore the case where range $R$ is carefully selected such that the STD statistics for the free rider is similar to that of all other clients.
In this experiment, for the gradient updates $G_{i,j}$ submitted by $C_i^f$, each value is randomly sampled from a uniform distribution within range [$-10^{-3}$, $10^{-3}$].
The STD statistics for all clients with the increase of 
learning rounds are shown in Figure~\ref{fig:random10e-3-even-std}. Note that the STD of the free rider and other clients are basically inseparable after round 10. However, our experimental results (in Figure~\ref{fig:random10e-3-even-dagmm}) suggest that DAGMM is still able to detect the free rider, both at the beginning training period (round 5), and when close to converge (round 80).


\subsubsection{Each client has different local data distribution.} 
\label{sec:random-eval-uneven}
To simulate the case where each client may have different local data with others, we first sort the MNIST dataset based on the labels each image has (i.e., from class 0 to class 9), and then sequentially distribute the sorted dataset to all clients.
As a result, a majority number of clients only have one class of MNIST data images locally (e.g., only have images with number 9), and the rest have up to two classes.
In this case, the free rider could be harder to detect by analyzing submitted gradient updates, because all other clients' updates are also different with each other.
As previous section has indicated that $R=10^{-3}$ generates a random update matrix that has similar STD with other clients, in this section we only show the results in this case.

\begin{figure}[hbtp]
     \centering
     \begin{subfigure}[b]{\linewidth}
         \centering
         \includegraphics[width=\textwidth]{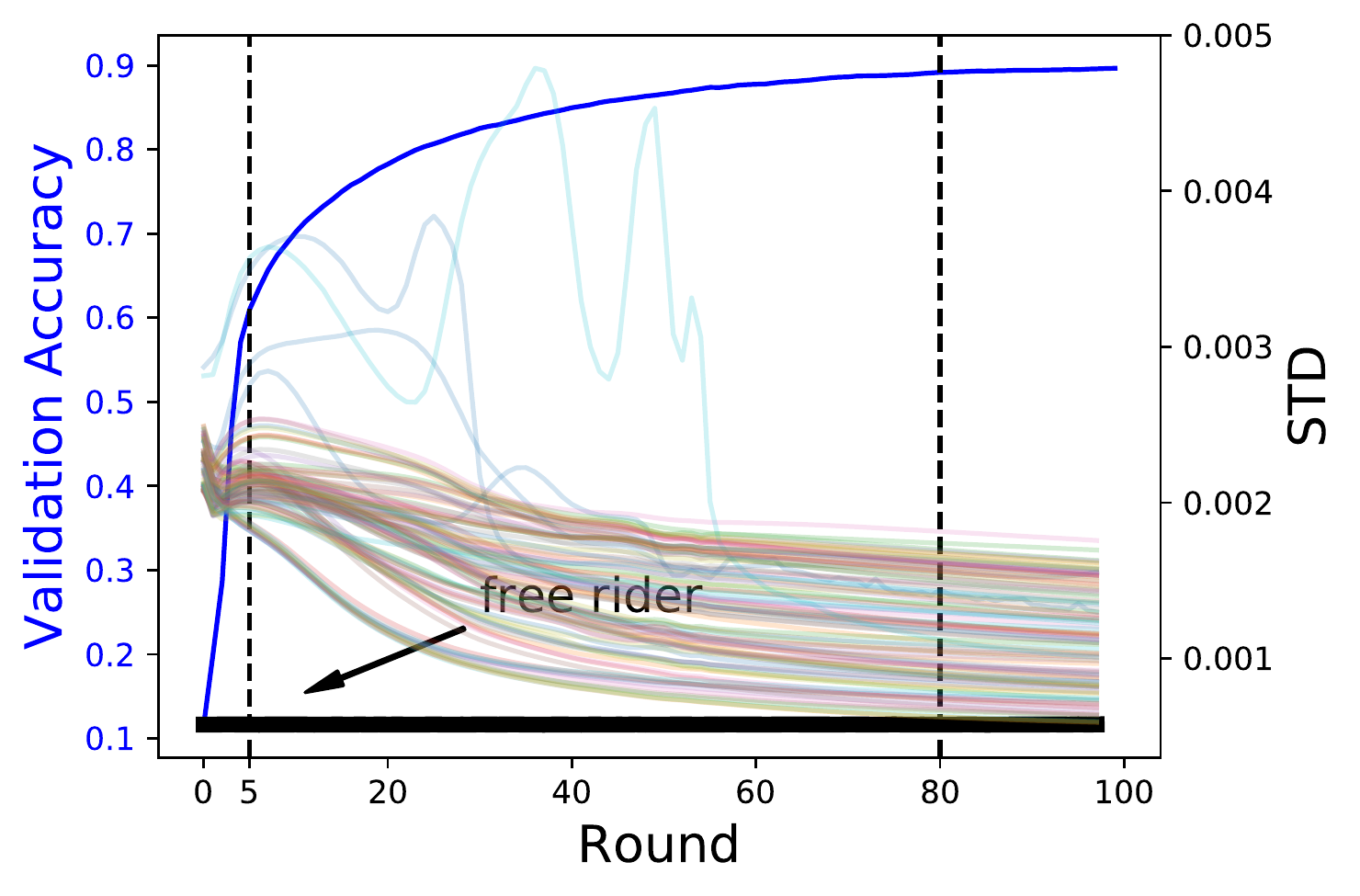}
         \caption{Standard deviation (STD) of the updates submitted by each client, with the change of training process}
         \label{fig:random10e-3-uneven-std}
     \end{subfigure}
     \begin{subfigure}[b]{\linewidth}
         \centering
         \begin{subfigure}[b]{.48\textwidth}
             \includegraphics[width=\linewidth]{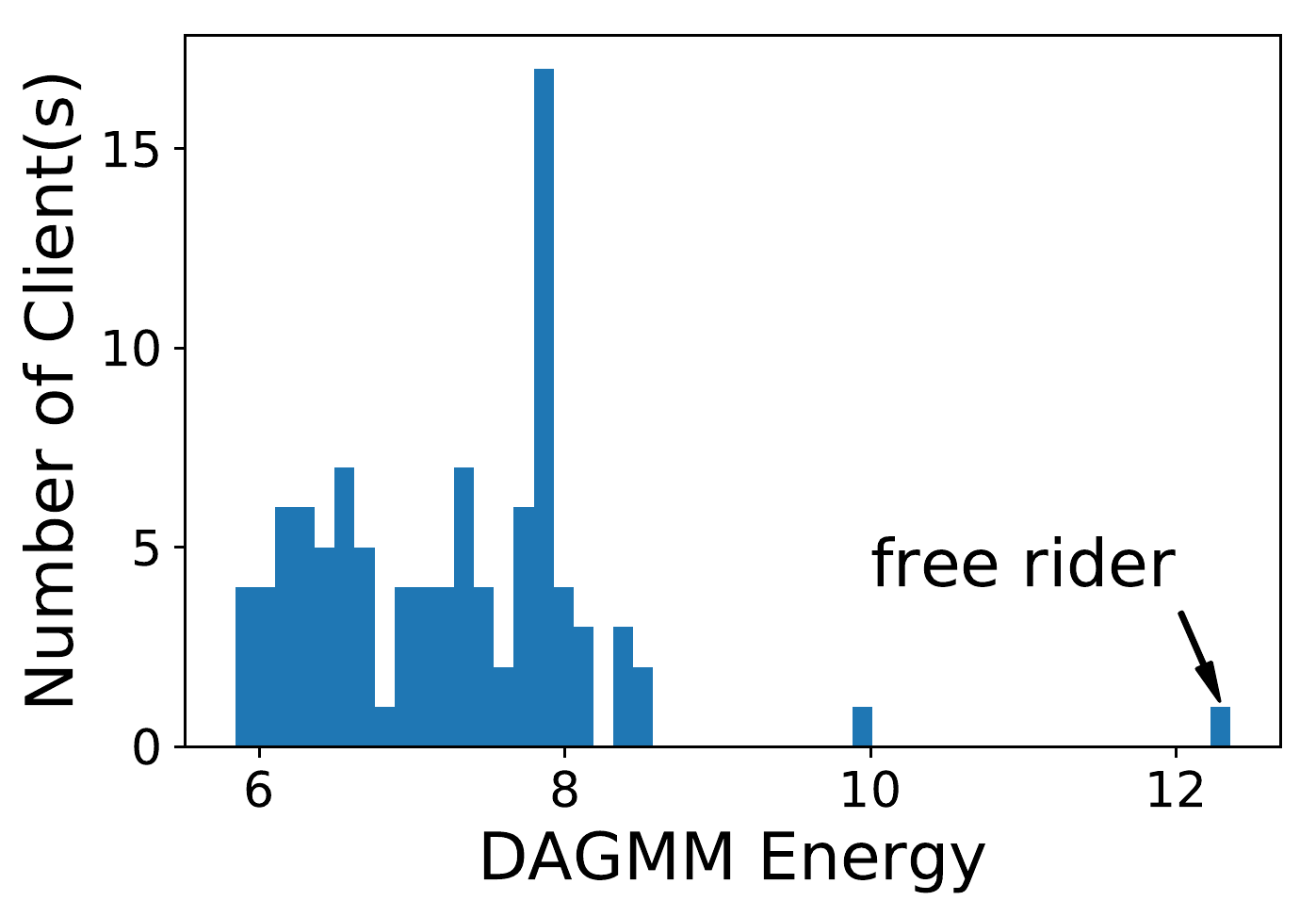}\hfill%
             \caption{Round 5}
         \end{subfigure}
         \begin{subfigure}[b]{.48\textwidth}
             \includegraphics[width=\linewidth]{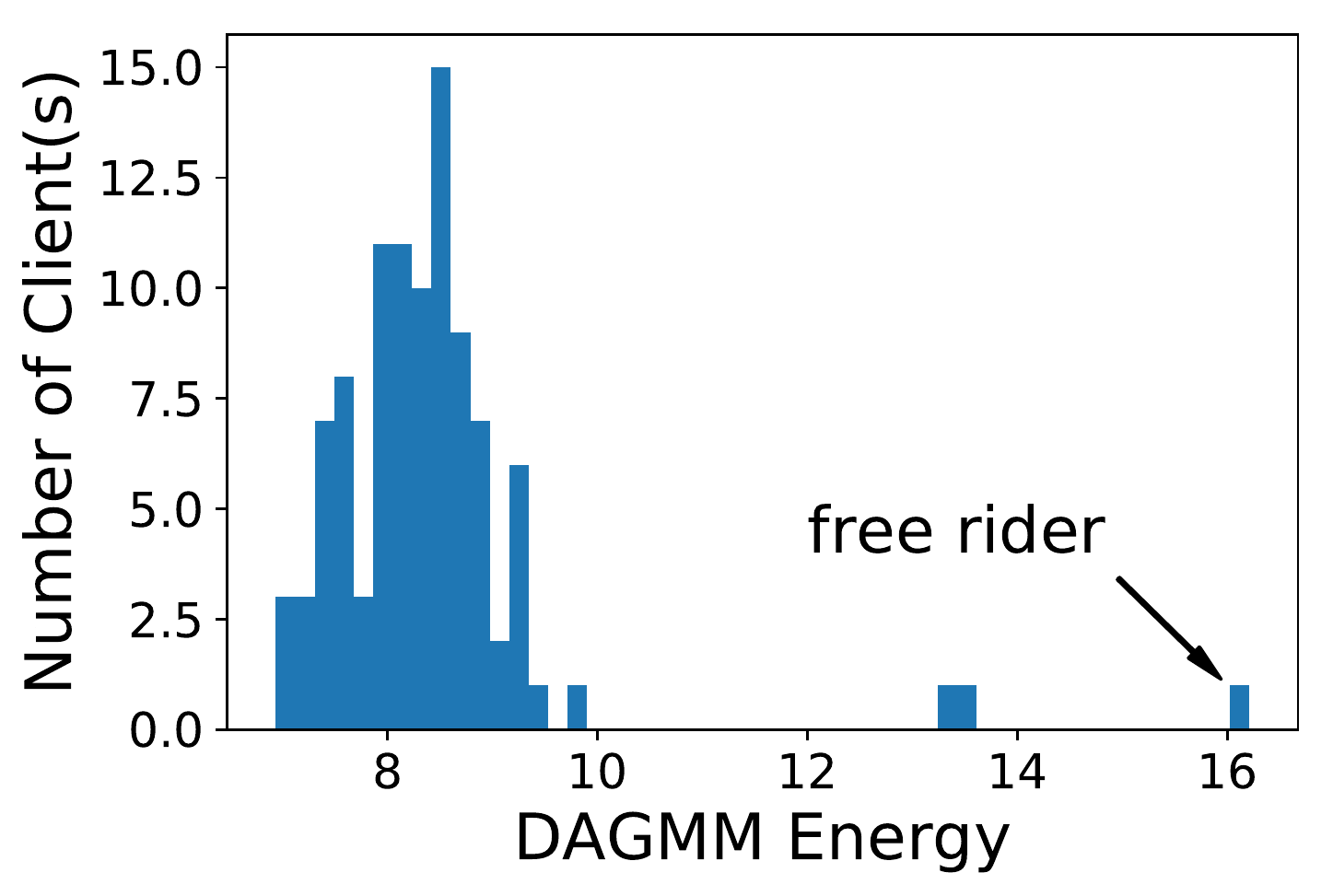}
             \caption{Round 80}
         \end{subfigure}
         \caption{DAGMM - detected}
         \label{fig:random10e-3-uneven-dagmm}
     \end{subfigure}
        \caption{Different local data distribution, randomly sampled updates from [$-10^{-3}$, $10^{-3}$]}
        \label{fig:random10e-3-uneven}
\end{figure}

\paragraph{Random weights range $\mathbf{R=10^{-3}}$.}
The STD statistics of each client's updates are presented in Figure~\ref{fig:random10e-3-uneven-std}. Note here the model 
accuracy increases slower,
and the STD for each client's gradient update is also bigger, which are all due to the different local data distribution on each client.

We demonstrate that DAGMM is still able to successfully detect such attacks, as shown in Figure~\ref{fig:random10e-3-uneven-dagmm}. 

\section{Attack II: Delta Weights}
\label{sec:delta}
\subsection{Attack specification}

As presented in Section~\ref{sec:pre-fed}, in the $j$-th round of federated learning, a client $C_i$ receives the global model $M_j$ and submits a local gradient update $G_{i,j}$.
With a received global model at each round, a free rider is able to construct more sophisticated fake gradient updates, than merely generating a random update matrix that has the same dimension with the global model. In this section, we consider a sophisticated attacker, which generates the fake gradient updates by subtracting two previously received global model. We call such an attack as \textit{delta weights attack}.

\paragraph{Delta weights attack.}
Suppose a free rider receives global model $M_{j-1}$ at round $j-1$ and $M_{j}$ at round $j$.
In the delta weights attack, the free rider $C_i^f$ constructs gradient updates $G_{i,j}^f$ by
\begin{equation}
\label{eq:delta}
G_{i,j}^f = M_{j-1} - M_{j} = \eta \cdot \frac{1}{n} \sum_{x=1}^n G_{x,j-1} \;\;\;\;\;\;\mathrm{(ref. Equation~\ref{eq:global-model})}
\end{equation}
This suggests that the fake gradient update $G_{i,j}^f$ constructed at a round $j$, is essentially the average gradient update submitted by all clients at previous round $j-1$, times a scaling factor $\eta$.

Note that for machine learning training, except for the first few epochs, the parameter change at each round is very small. As a result, the constructed fake gradients could be very similar to the ones submitted by other clients, and could most likely evade any detection mechanism that uses a validation dataset to detect free riders by checking if utility drop happens.


\subsection{Defense strategy}
For this attack,
the effectiveness of the defense strategy could be affected by global learning rate $\eta$. In the typical case, when $\eta=1$, the fake gradient update generated by the free rider is simply the average of all other clients' updates in previous rounds.
When $\eta$ is small, free rider's fake updates would be much smaller compared with other clients' updates.
Interestingly, we find that DAGMM works well when $\eta$ is close to 1, but fails to detect the free rider when $\eta$ is small. Another observation is that, when $\eta$ is small, the STD statistics of the fake gradients is also much smaller than other clients' updates. 
The detailed results will be presented in the following experiment section.
Based on the two observations, we propose a general free-rider detection approach that combines both STD and DAGMM, aiming to provide a general free rider detection mechanism that works under different learning rates $\eta$. We refer this method as \textit{STD-DAGMM detection}.

\begin{figure}[hbtp]
     \centering
         \includegraphics[width=\linewidth]{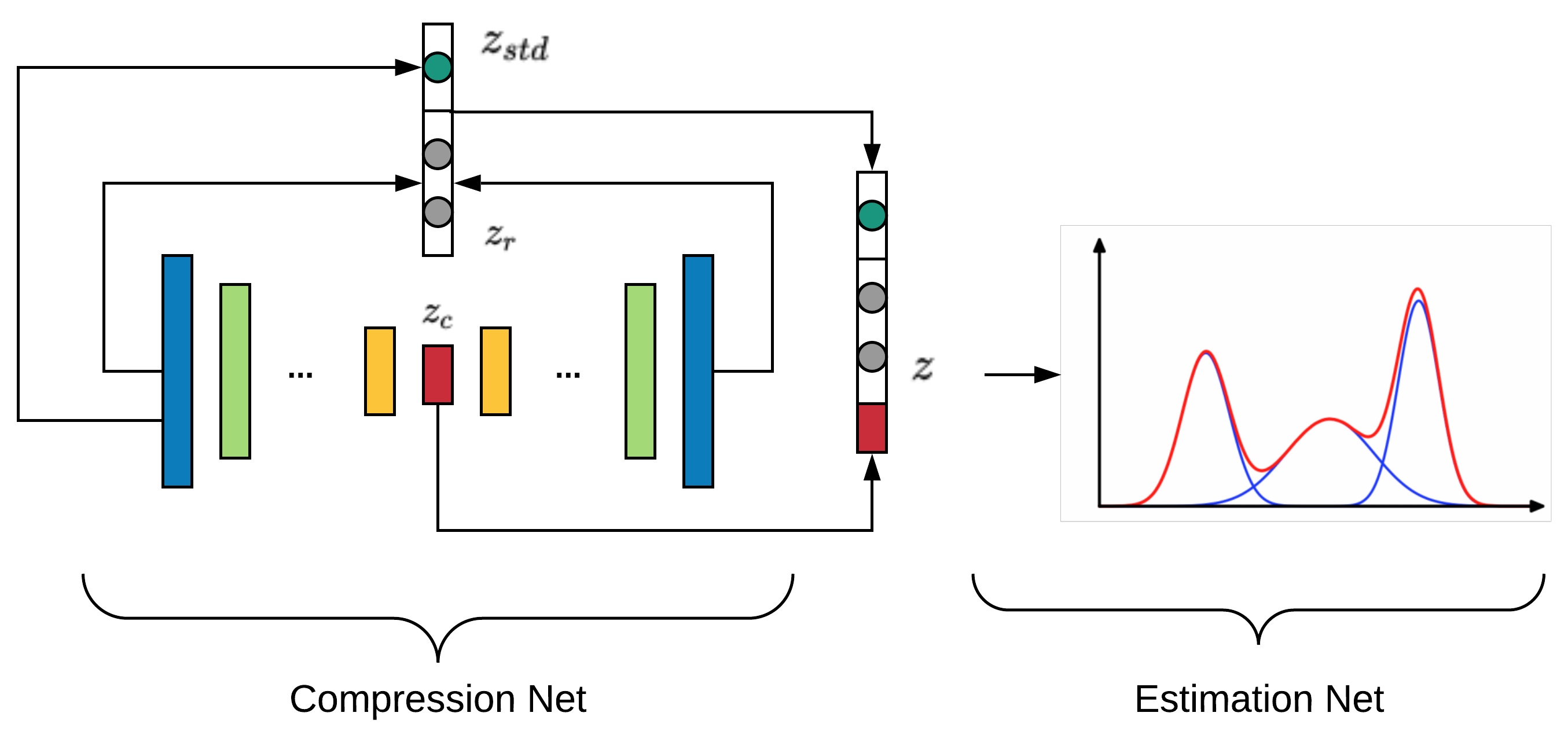}
 \caption{STD-DAGMM model architecture}
         \label{fig:std-dagmm}
\end{figure}

\paragraph{STD-DAGMM detection.}
In Section~\ref{sec:dagmm}, we have demonstrated that for DAGMM, the input vector of the estimation net concatenates both the reduced dimensional embedding vector in Autoencoder, and the distance metrics (e.g., Euclidean and Cosine distances) between Autoencoder input and output. To add STD metric into DAGMM network, we could append this metric into the input vector of estimation net.
We refer this new approach as STD-DAGMM, as shown in Figure~\ref{fig:std-dagmm}.
Specifically, given a gradient update matrix, we first concatenate all rows in the matrix to create a one-dimensional vector as described in Section~\ref{sec:random-defense}, and then feed it into the Autoencoder (compression network) of STD-DAGMM. We then calculate the standard deviation (STD) $z_{std}$ of this one-dimensional input vector, as well as the Euclidean and Cosine distance metrics as discussed in Section~\ref{sec:random-defense}, and further stack the 3 metrics to create a vector $z_r$ of size 3.
Finally, this vector $z_r$ is concatenated with the low-dimensional representation $z_c$ learned by Autoencoder (compression network). The output concatenated vector $z$ is fed into the estimation network for multivariate Gaussian estimation.


The learning objective for STD-DAGMM is the same as DAGMM, which is to simultaneously minimize Autoencoder reconstruction error to better preserve the important feature of input samples, and maximize the likelihood of observing input samples.

We find that STD-DAGMM is extremely effective towards free rider detection, even for very sophisticated attacks which show no clear separation boundary between basic statistics of fake and true gradient updates, as will be presented in Section~\ref{sec:advanced-delta}.


\subsection{Experimental validation}
\label{sec:delta-eval}
As in previous section, for this set of experiments, we consider a total of 100 clients and only 1 free rider. Remember that this attack is simply constructed by subtracting two previous global models. If there are multiple attackers doing such attack, these free riders will be trivial to detect as they all submit the same update matrix.

Similar to Section~\ref{sec:random-eval}, we adopt MNIST dataset and a 2-layer fully connected deep neural network, and present our results under two cases, in terms of whether each client has similar local data distribution.

\subsubsection{Each client has similar local data distribution.}
In this section, we demonstrate how previously proposed detection mechanism may fail, and show the effectiveness of our proposed STD-DAGMM method under various learning rates $\eta$.

\begin{figure}[hbtp]
     \centering
     \begin{subfigure}[b]{\linewidth}
         \centering
         \includegraphics[width=\textwidth]{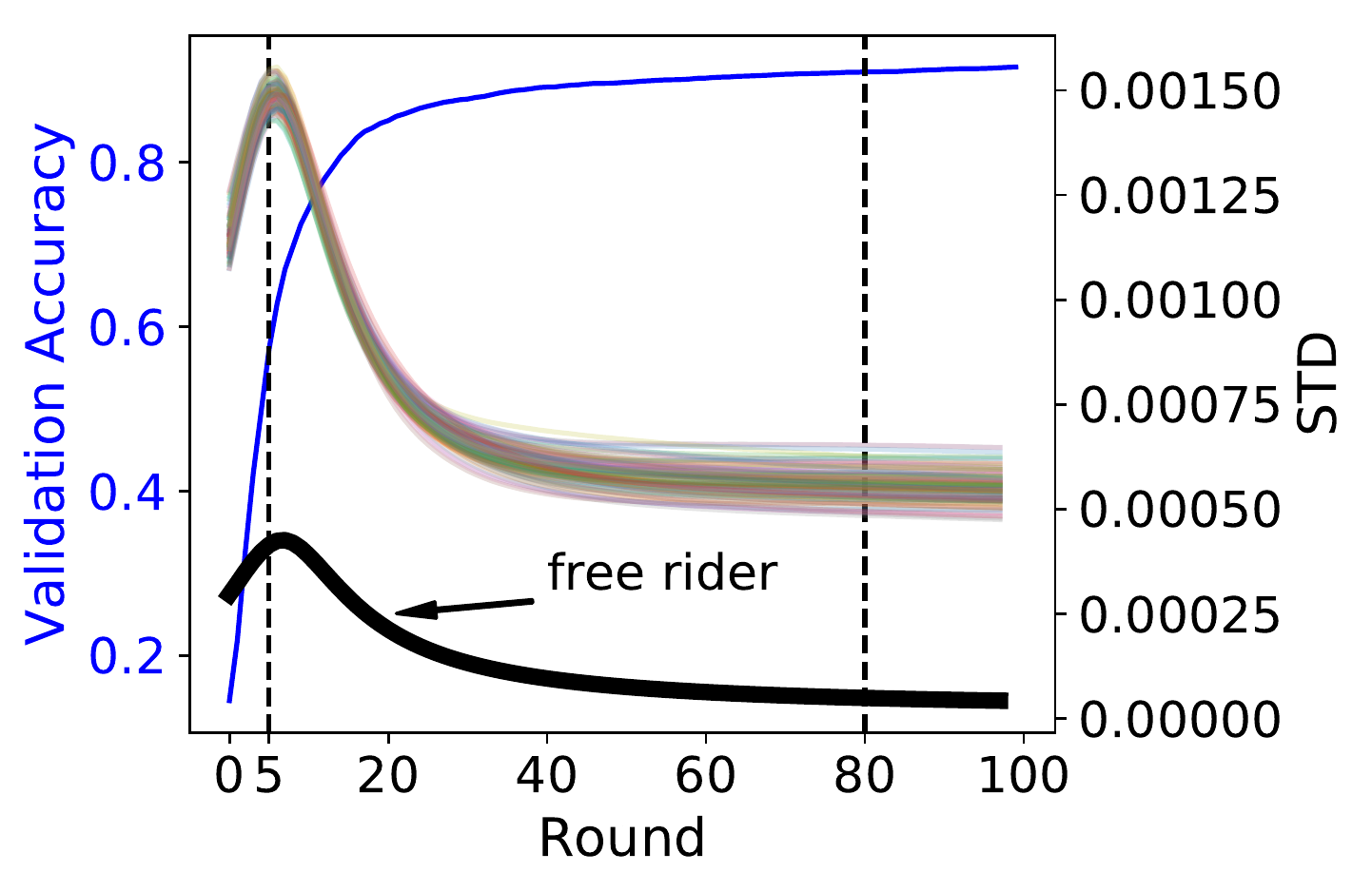}
         \caption{Standard deviation (STD) of the updates submitted by each client, with the change of training process}
         \label{fig:delta0.3-even-std}
     \end{subfigure}
     \begin{subfigure}[b]{\linewidth}
         \centering
         \begin{subfigure}[b]{0.49\linewidth}
             \includegraphics[width=\linewidth]{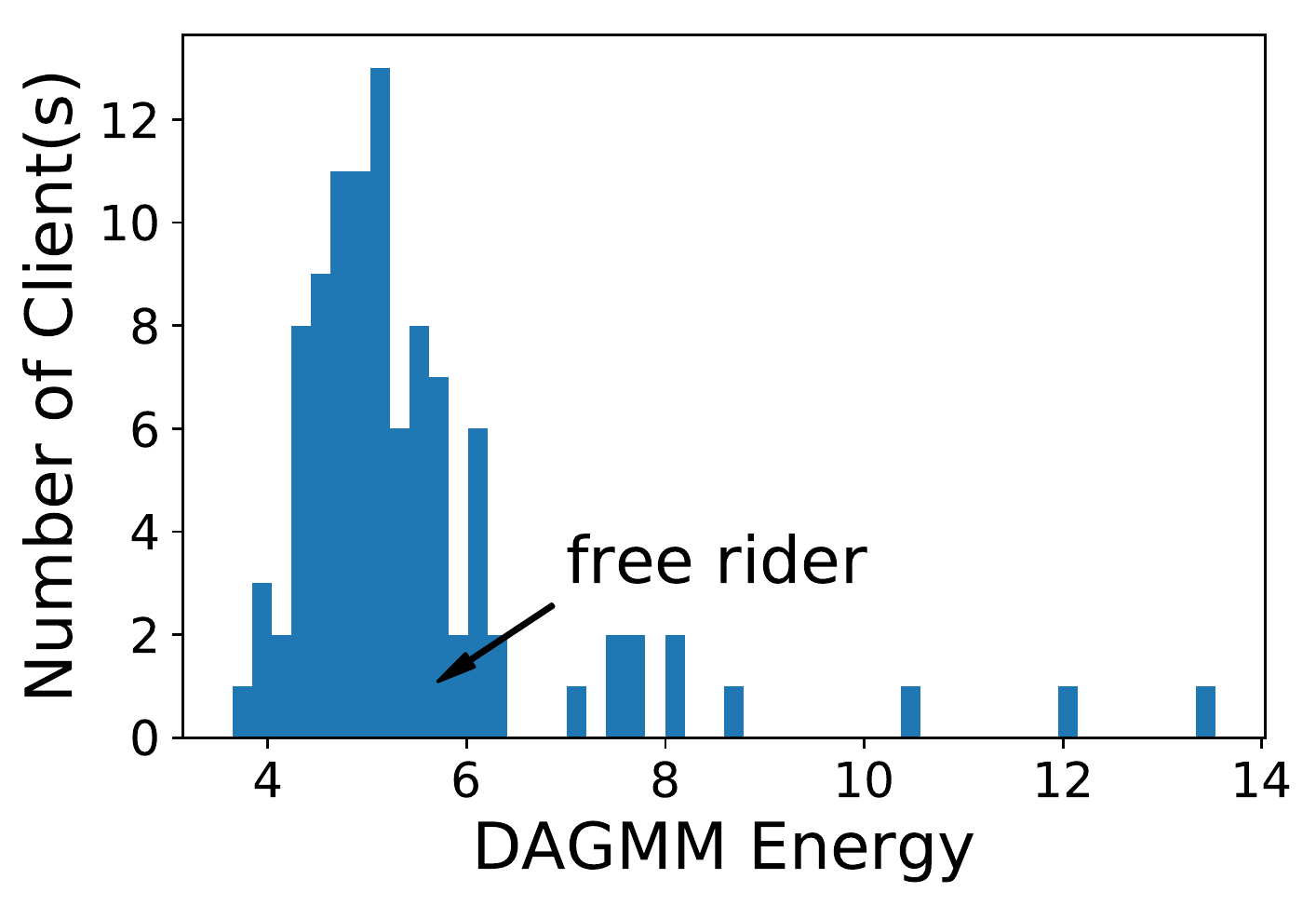}\hfill%
             \caption{Round 5}
             \label{fig:delta0.3-even-dagmm-round5}
         \end{subfigure}
         \begin{subfigure}[b]{0.49\linewidth}
             \includegraphics[width=\linewidth]{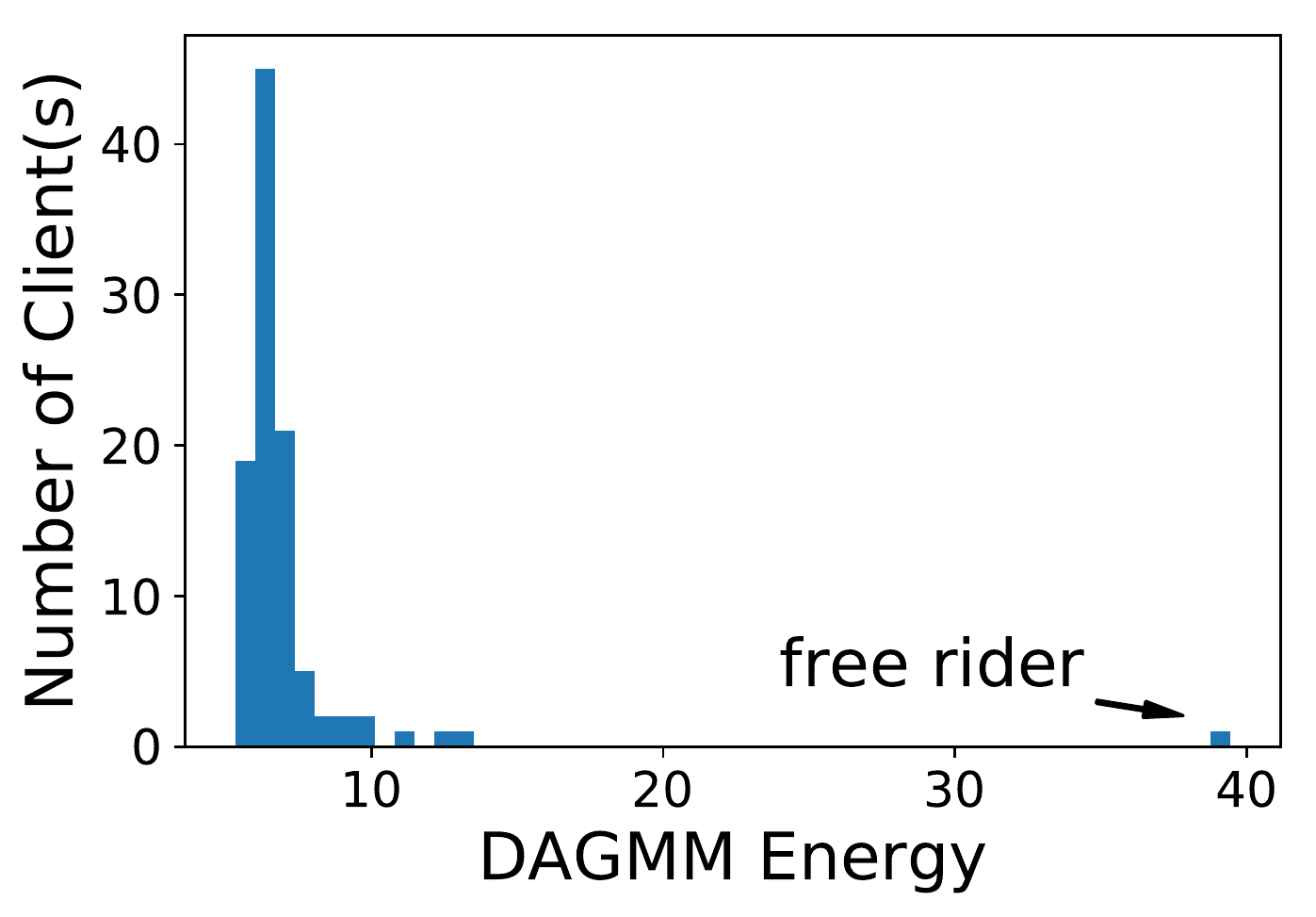}
             \caption{Round 80}
         \end{subfigure}
         \caption{DAGMM - not detected}
         \label{fig:delta0.3-even-dagmm}
     \end{subfigure}
     \begin{subfigure}[b]{\linewidth}
         \centering
         \begin{subfigure}[b]{0.49\linewidth}
             \includegraphics[width=\linewidth]{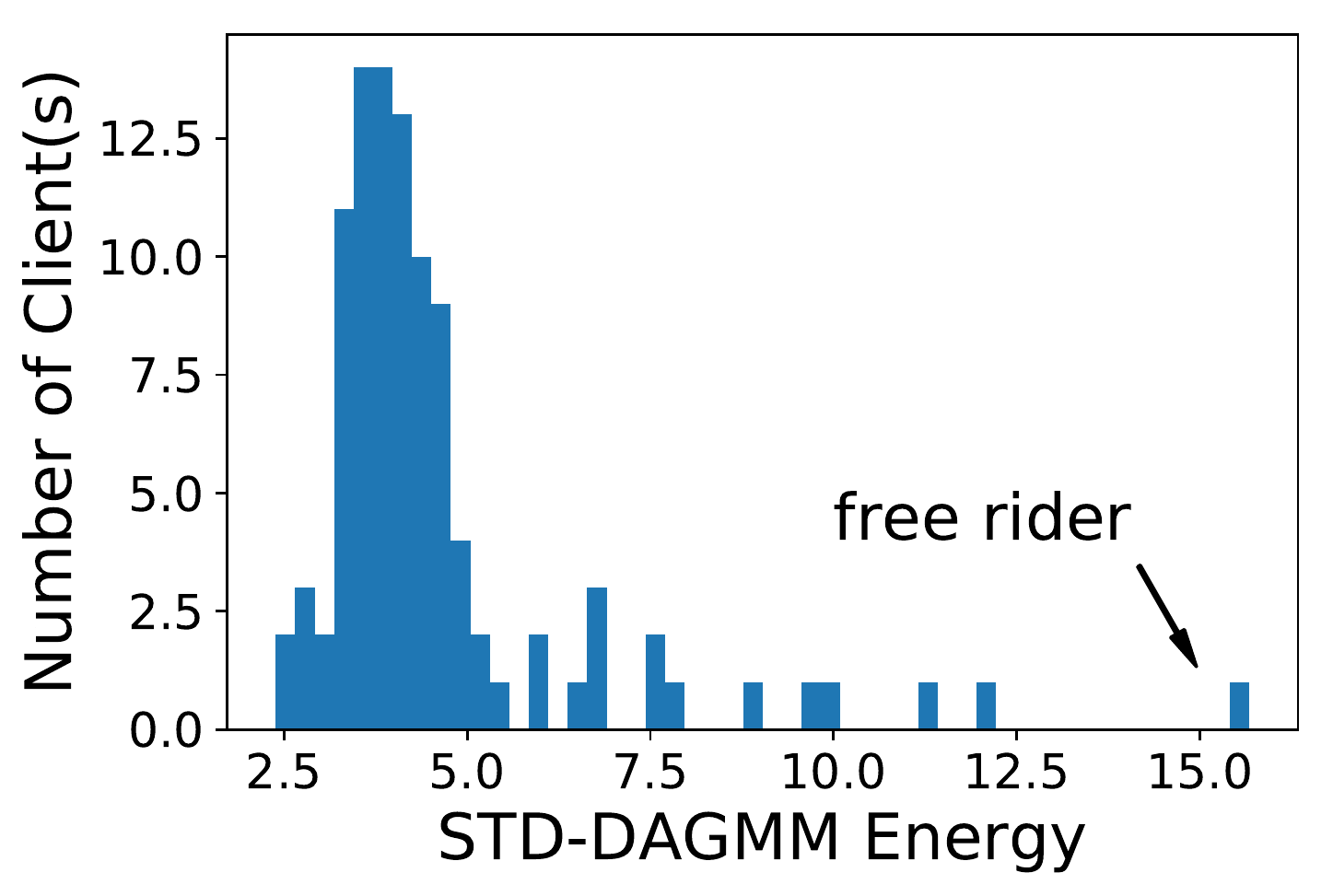}\hfill%
             \caption{Round 5}
         \end{subfigure}
         \begin{subfigure}[b]{0.49\linewidth}
             \includegraphics[width=\linewidth]{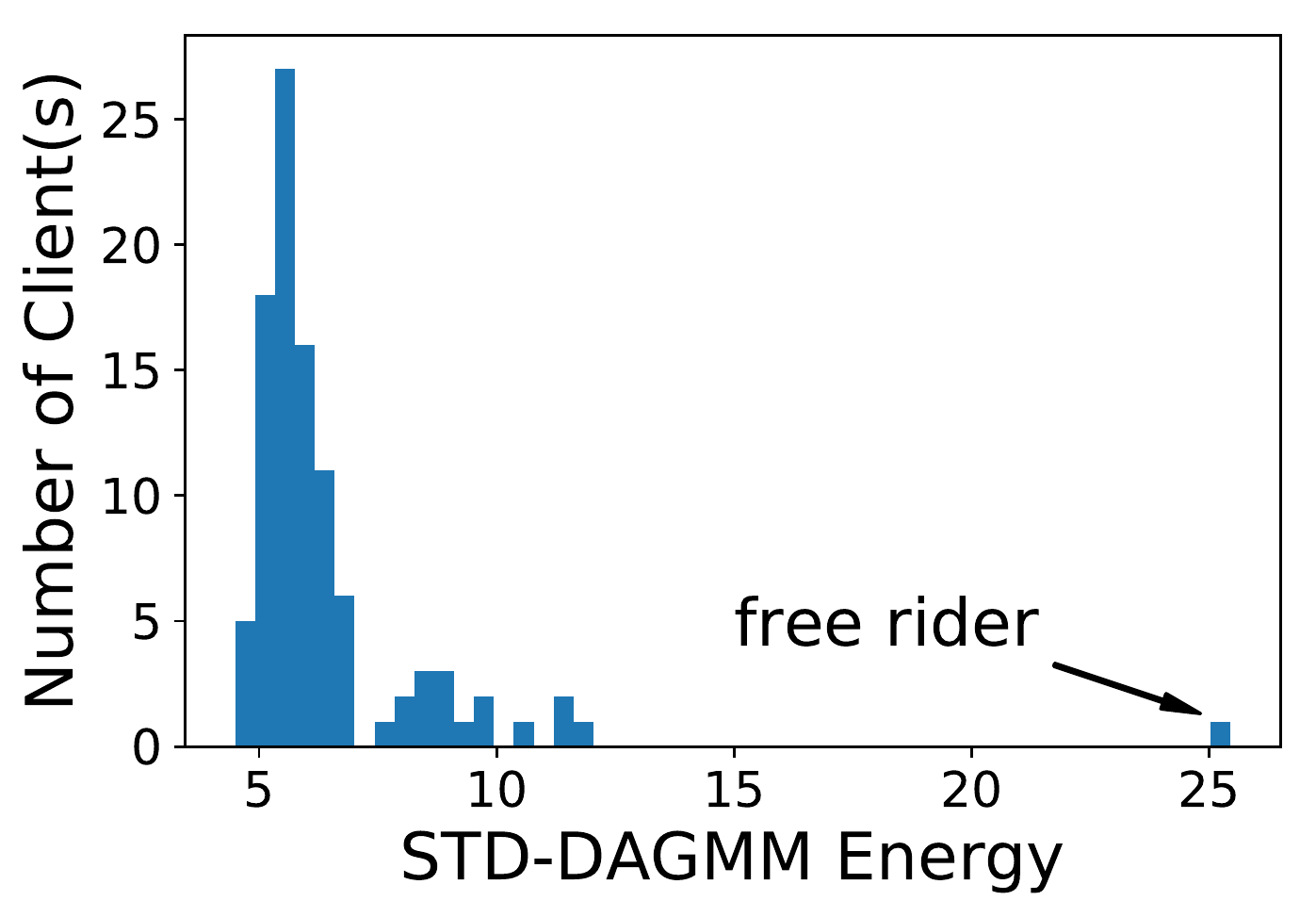}
             \caption{Round 80}
         \end{subfigure}
         \caption{STD-DAGMM - detected}
         \label{fig:delta0.3-even-stddagmm}
     \end{subfigure}
        \caption{Similar local data distribution, delta weights attack with $\eta=0.3$}
        \label{fig:delta0.3-even}
\end{figure}

We find that when the global learning rate $\eta$ is small (less than 0.3), DAGMM fails to detect the free rider at the beginning of the training period (e.g., round 5). This could happen because: 1) with a low global learning rate $\eta$, the gradient update a free rider constructs is also degraded by $\eta$, which is closer to $0$ and easier for the network to memorize; and 2) compared with Figure~\ref{fig:random10e-4-even-std}, the weight submitted by free riders and benign clients have similar directions, which makes the Consine distance metric in DAGMM not work.
However, we observe that when DAGMM fails,
the STD of the free rider gradient updates is much smaller than those of other clients, due to the scaling factor $\eta$.
This inspires us to add the STD metric into DAGMM network, referred as STD-DAGMM detection approach. The detection result of this method is shown in Figure~\ref{fig:delta0.3-even-stddagmm}. The free rider has an energy value much larger than other clients, and thus could be easily detected by a wide range of thresholds.

\begin{figure}[hbtp]
     \centering
     \begin{subfigure}[b]{\linewidth}
         \centering
         \includegraphics[width=\linewidth]{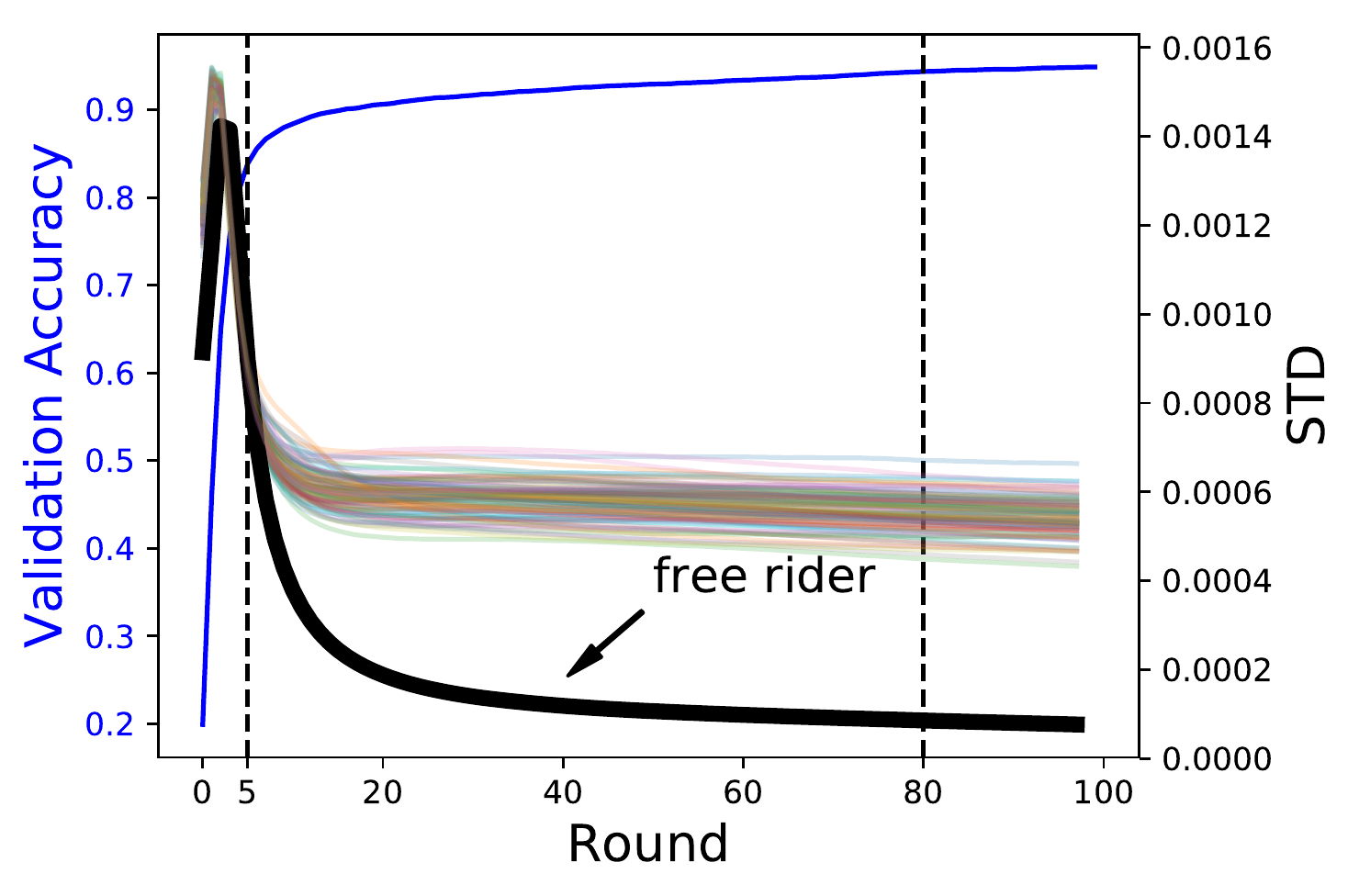}
         \caption{Standard deviation (STD) of the updates submitted by each client, with the change of training process}
         \label{fig:delta1-even-std}
     \end{subfigure}
     \begin{subfigure}[b]{\linewidth}
         \centering
         \begin{subfigure}[b]{0.49\linewidth}
             \includegraphics[width=\linewidth]{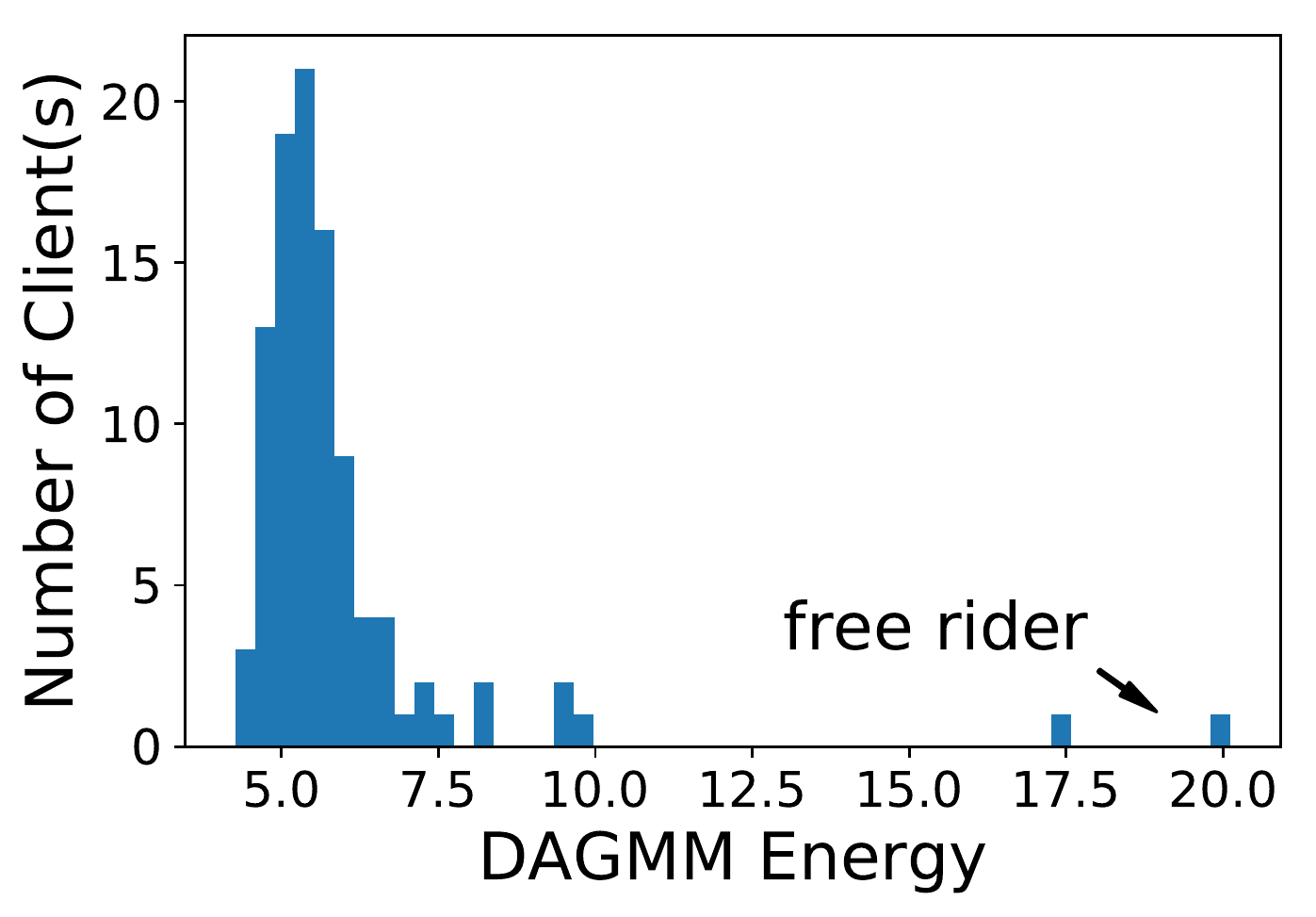}\hfill%
             \caption{Round 5}
             \label{fig:delta1-even-dagmm-round5}
         \end{subfigure}
         \begin{subfigure}[b]{0.49\linewidth}
             \includegraphics[width=\linewidth]{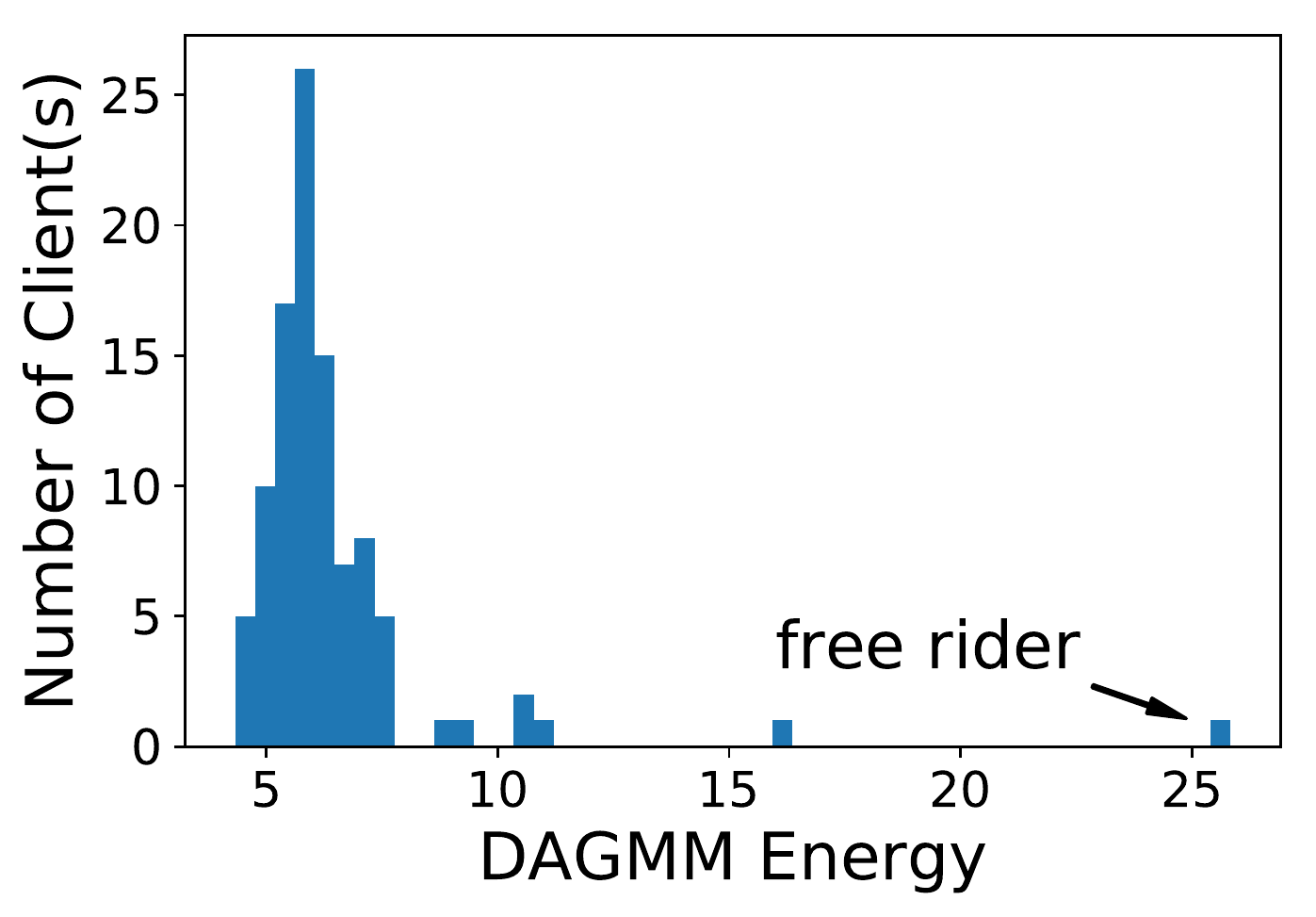}
             \caption{Round 80}
         \end{subfigure}
         \caption{DAGMM - detected}
         \label{fig:delta1-even-dagmm}
     \end{subfigure}
     \begin{subfigure}[b]{\linewidth}
         \centering
         \begin{subfigure}[b]{0.49\linewidth}
             \includegraphics[width=\linewidth]{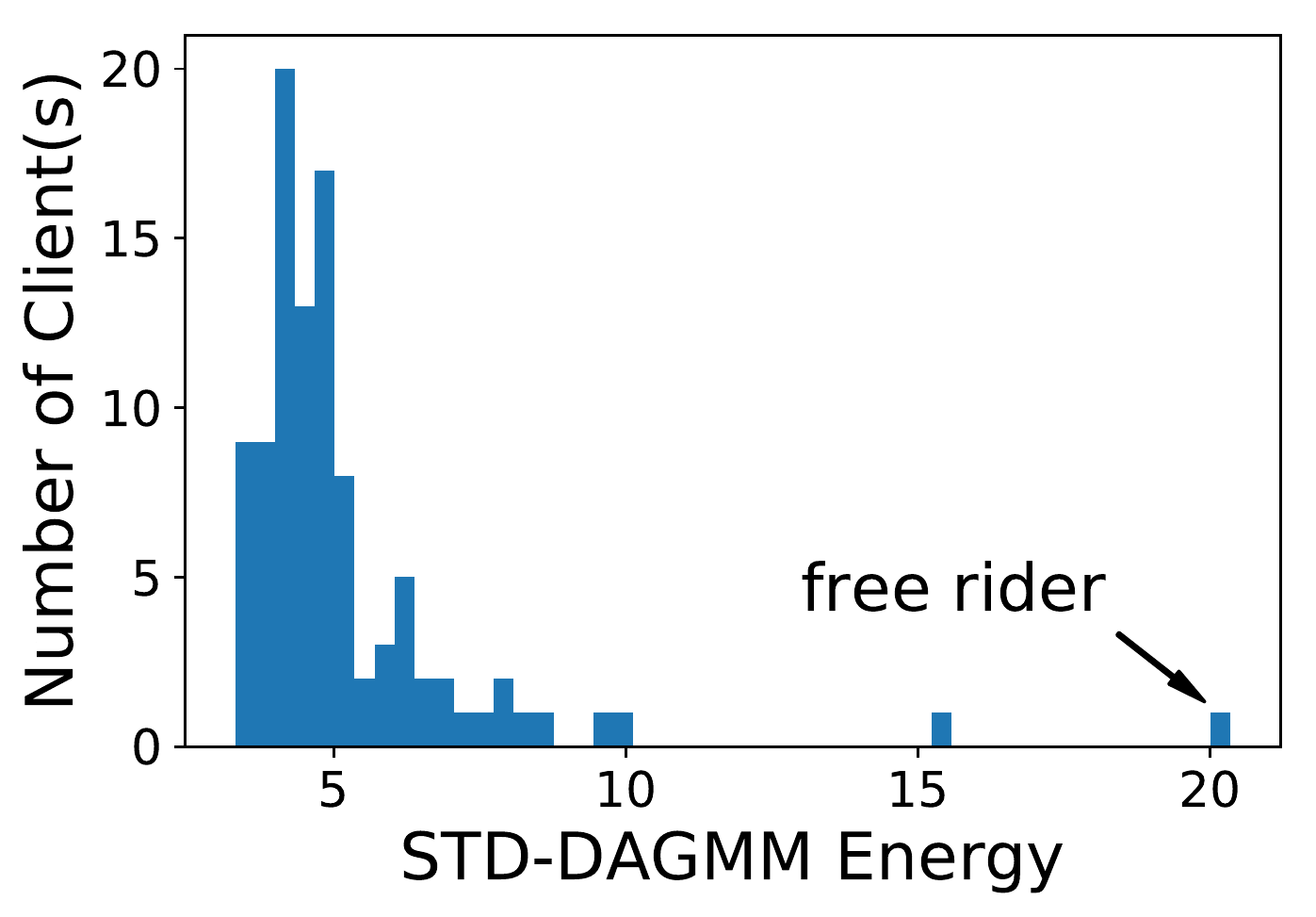}\hfill%
             \caption{Round 5}
         \end{subfigure}
         \begin{subfigure}[b]{0.49\linewidth}
             \includegraphics[width=\linewidth]{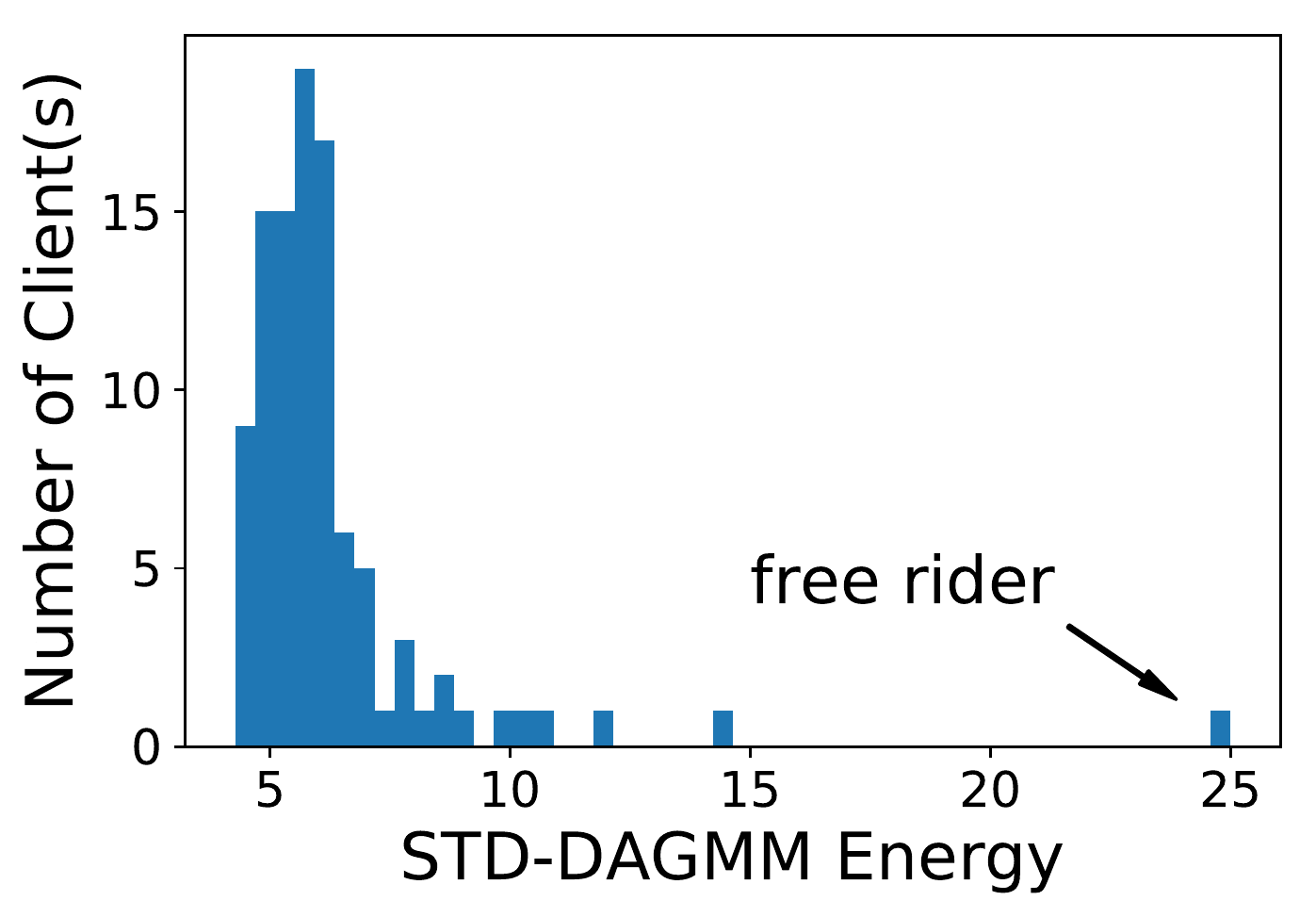}
             \caption{Round 80}
         \end{subfigure}
         \caption{STD-DAGMM - detected}
         \label{fig:delta1-even-stddagmm}
     \end{subfigure}
        \caption{Similar local data distribution, delta weights attack with $\eta=1$}
        \label{fig:delta1-even}
\end{figure}

Based on the statistics in Figure~\ref{fig:delta0.3-even-std}, one may think that the STD metric itself could be a suitable indicator for free rider detection.
However, we show that it is not enough. Specifically, we demonstrate the results when $\eta=1$ in
Figure~\ref{fig:delta1-even}. Note in Figure~\ref{fig:delta1-even-std}, before round 10, the STD curve of free rider fake updates mostly overlaps with other clients. Hence, using STD metric is not able to separate the free rider from others. 
However, DAGMM is effective in this case with its ability to capture high-dimensional patterns, as shown in Figure~\ref{fig:delta1-even-dagmm}.
Since the proposed STD-DAGMM method combines STD and DAGMM, it also suffices to detect the free rider in this case, as in Figure~\ref{fig:delta1-even-stddagmm}.
Hence, we believe the proposed STD-DAGMM method is a general technique that is able to detect the free rider utilizing delta weights attack, under different 
learning rates $\eta$.


\begin{figure}[hbtp]
     \centering
     \begin{subfigure}[b]{1.05\linewidth}
         \centering
         \includegraphics[width=\textwidth]{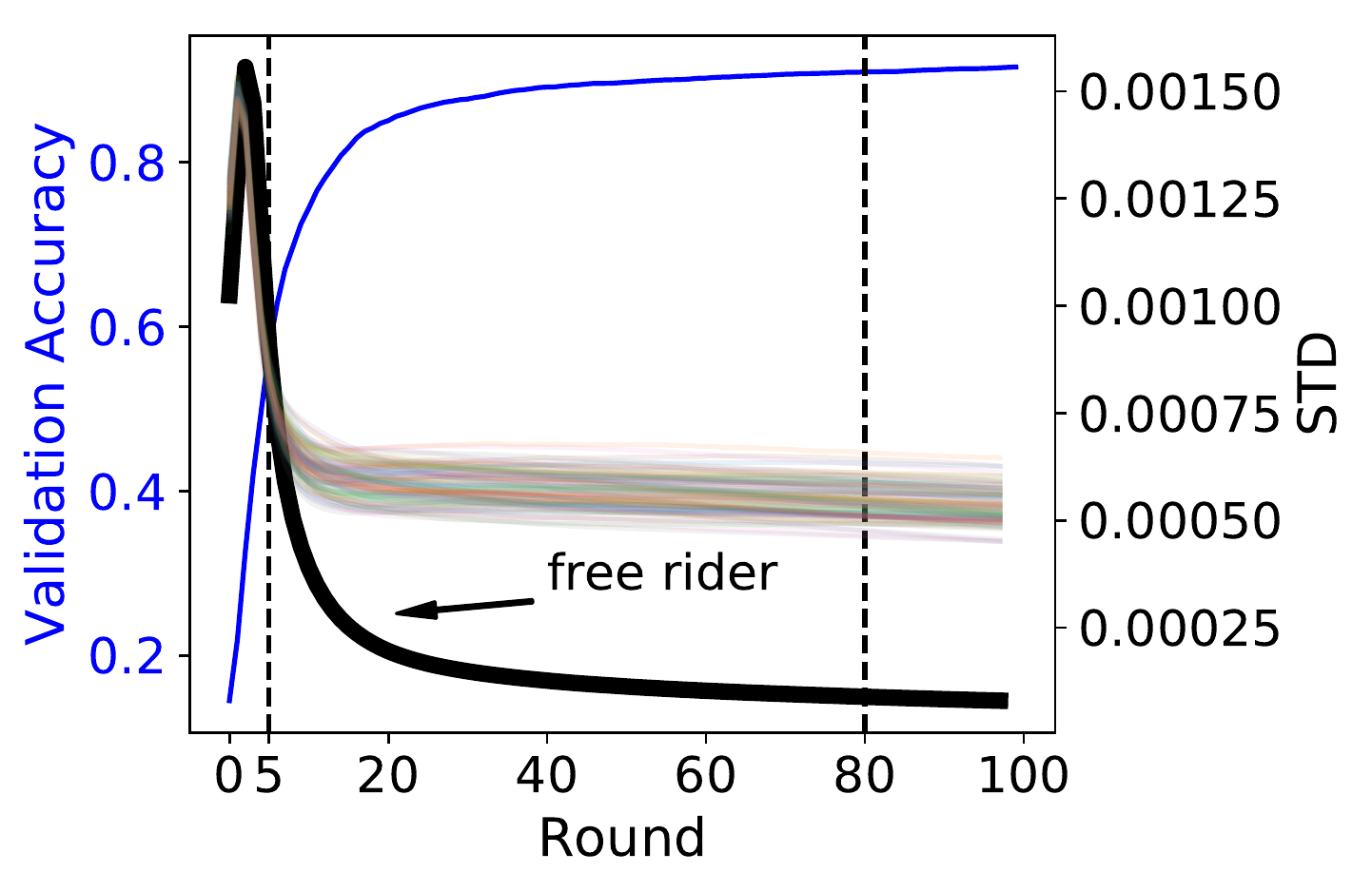}
         \caption{Standard deviation (STD) of the updates submitted by each client, with the change of training process}
         \label{fig:delta0.7-even-std}
     \end{subfigure}
     \begin{subfigure}[b]{\linewidth}
         \centering
         \begin{subfigure}[b]{0.49\linewidth}
             \includegraphics[width=\linewidth]{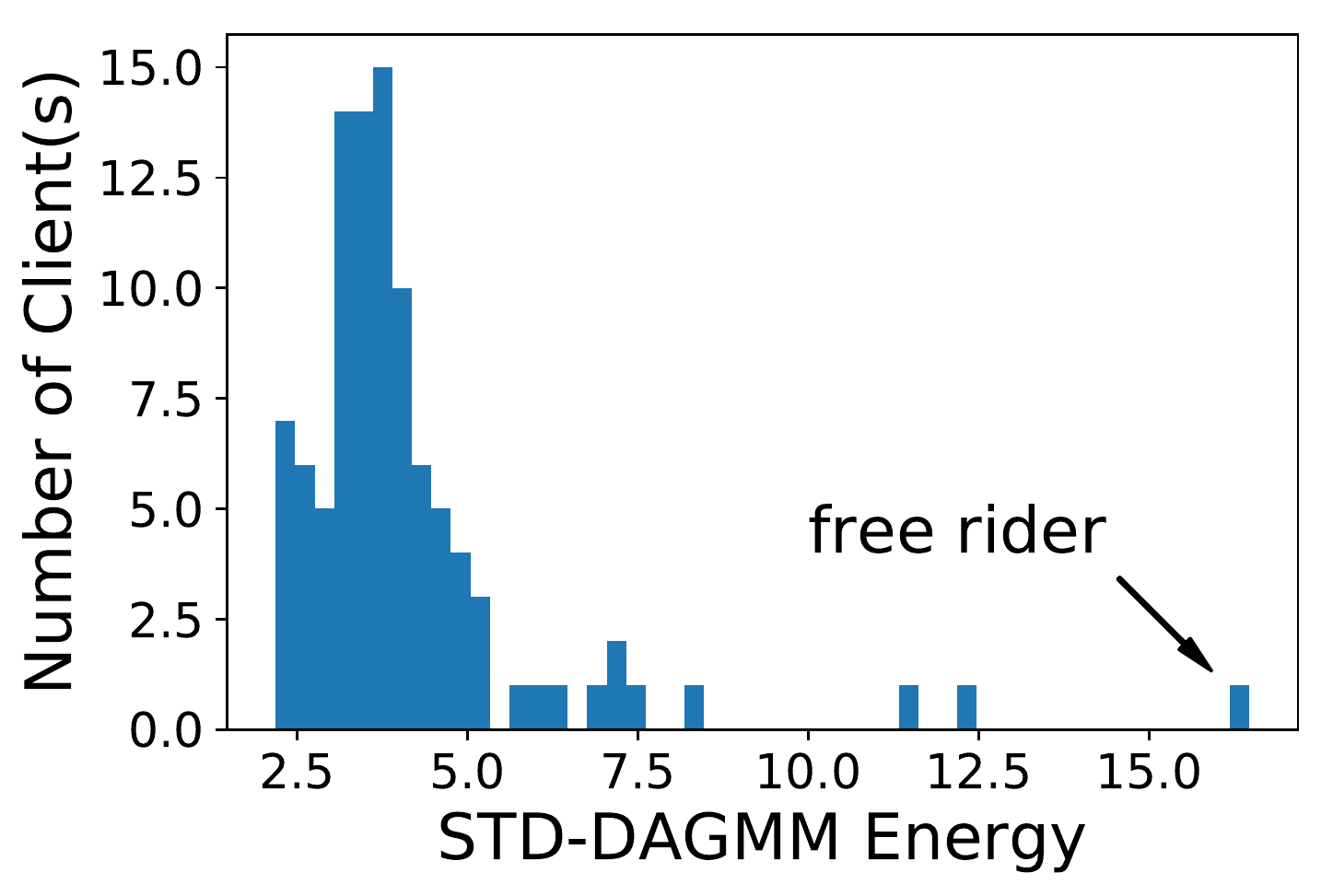}\hfill%
             \caption{Round 5}
         \end{subfigure}
         \begin{subfigure}[b]{0.49\linewidth}
             \includegraphics[width=\linewidth]{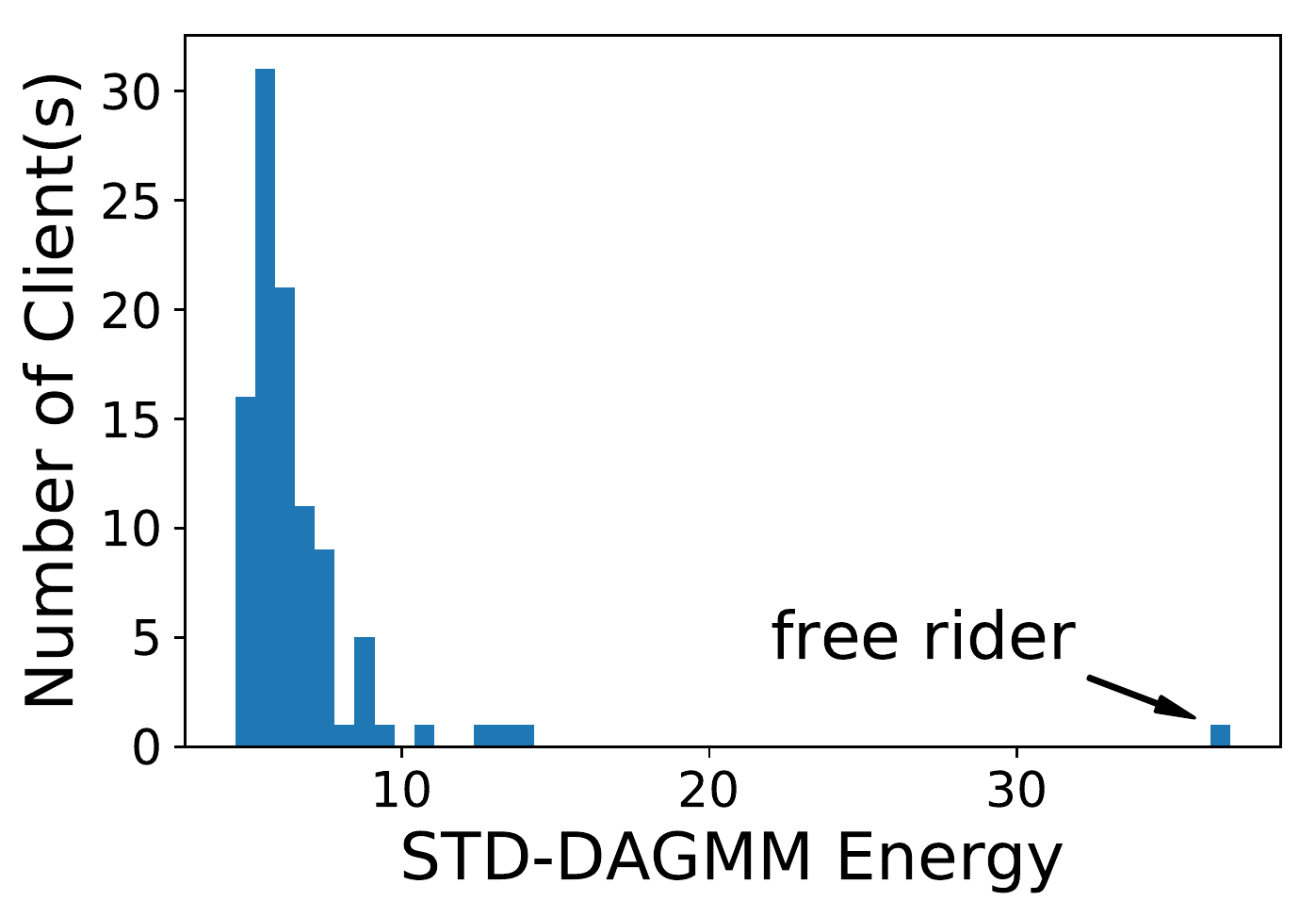}
             \caption{Round 80}
         \end{subfigure}
         \caption{STD-DAGMM - detected}
         \label{fig:delta0.7-even-stddagmm}
     \end{subfigure}
        \caption{Similar local data distribution, delta weights attack with $\eta=0.7$}
        \label{fig:delta0.7-even}
\end{figure}

\begin{figure}[hbtp]
     \centering
     \begin{subfigure}[b]{\linewidth}
         \centering
         \includegraphics[width=\textwidth]{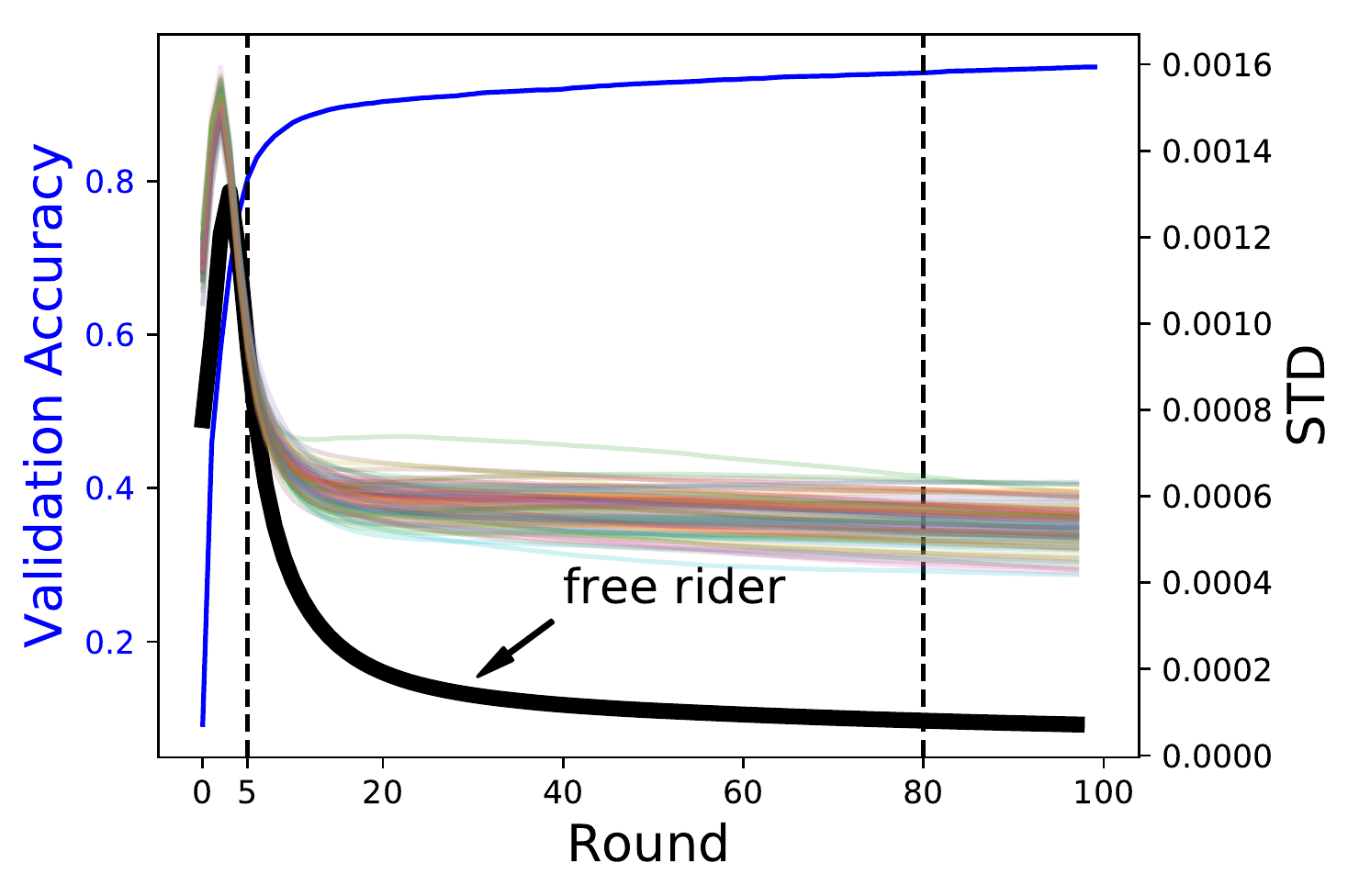}
         \caption{Standard deviation (STD) of the updates submitted by each client, with the change of training process}
         \label{fig:delta0.9-even-std}
     \end{subfigure}
     \begin{subfigure}[b]{\linewidth}
         \centering
         \begin{subfigure}[b]{0.49\linewidth}
             \includegraphics[width=\linewidth]{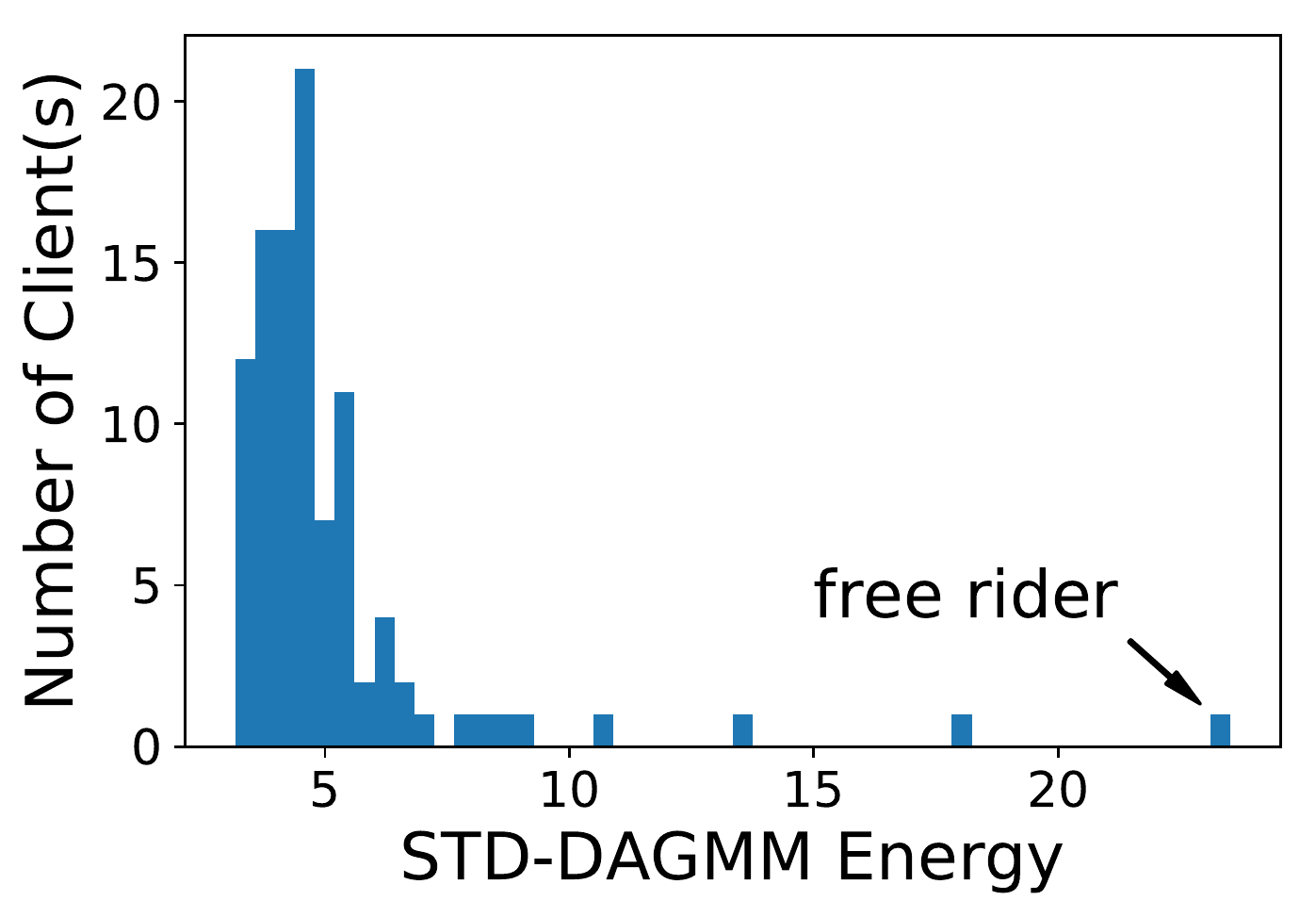}\hfill%
             \caption{Round 5}
         \end{subfigure}
         \begin{subfigure}[b]{0.49\linewidth}
             \includegraphics[width=\linewidth]{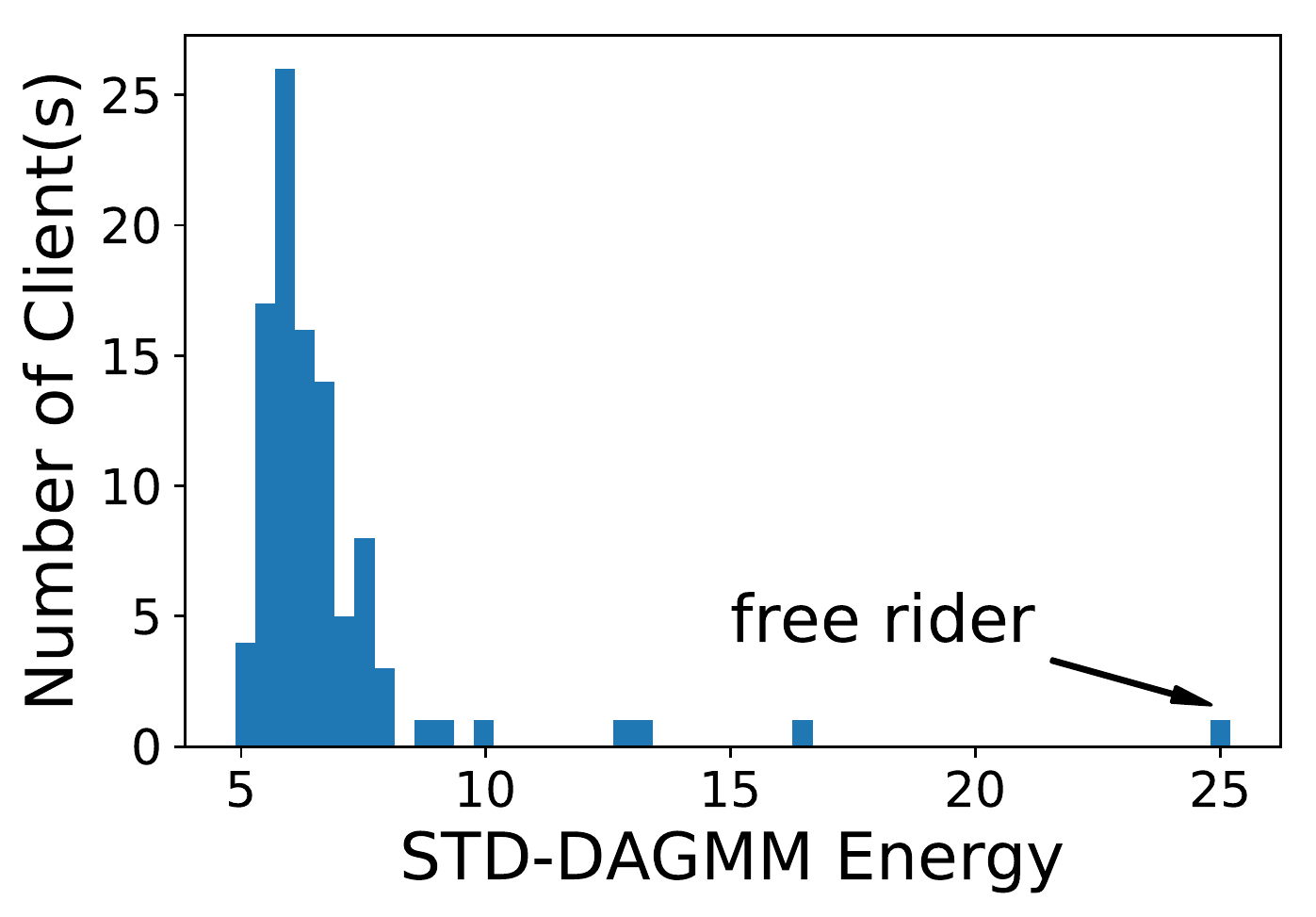}
             \caption{Round 80}
         \end{subfigure}
         \caption{STD-DAGMM - detected}
         \label{fig:delta0.9-even-stddagmm}
     \end{subfigure}
        \caption{Similar local data distribution, delta weights attack with $\eta=0.9$}
        \label{fig:delta0.9-even}
\end{figure}

To further validate the generality of our proposed STD-DAGMM approach, we further show its effectiveness under multiple other learning rates, i.e., $\eta=0.7$ as shown in Figure~\ref{fig:delta0.7-even}, and $\eta=0.9$ as in Figure~\ref{fig:delta0.9-even}. Both results demonstrate that STD-DAGMM is a general approach for free rider detection with delta weights generation under different global learning rates $\eta$.

\subsubsection{Each client has different local data distribution.} 
\begin{figure}[hbtp]
     \centering
     \begin{subfigure}[b]{\linewidth}
         \centering
         \includegraphics[width=\textwidth]{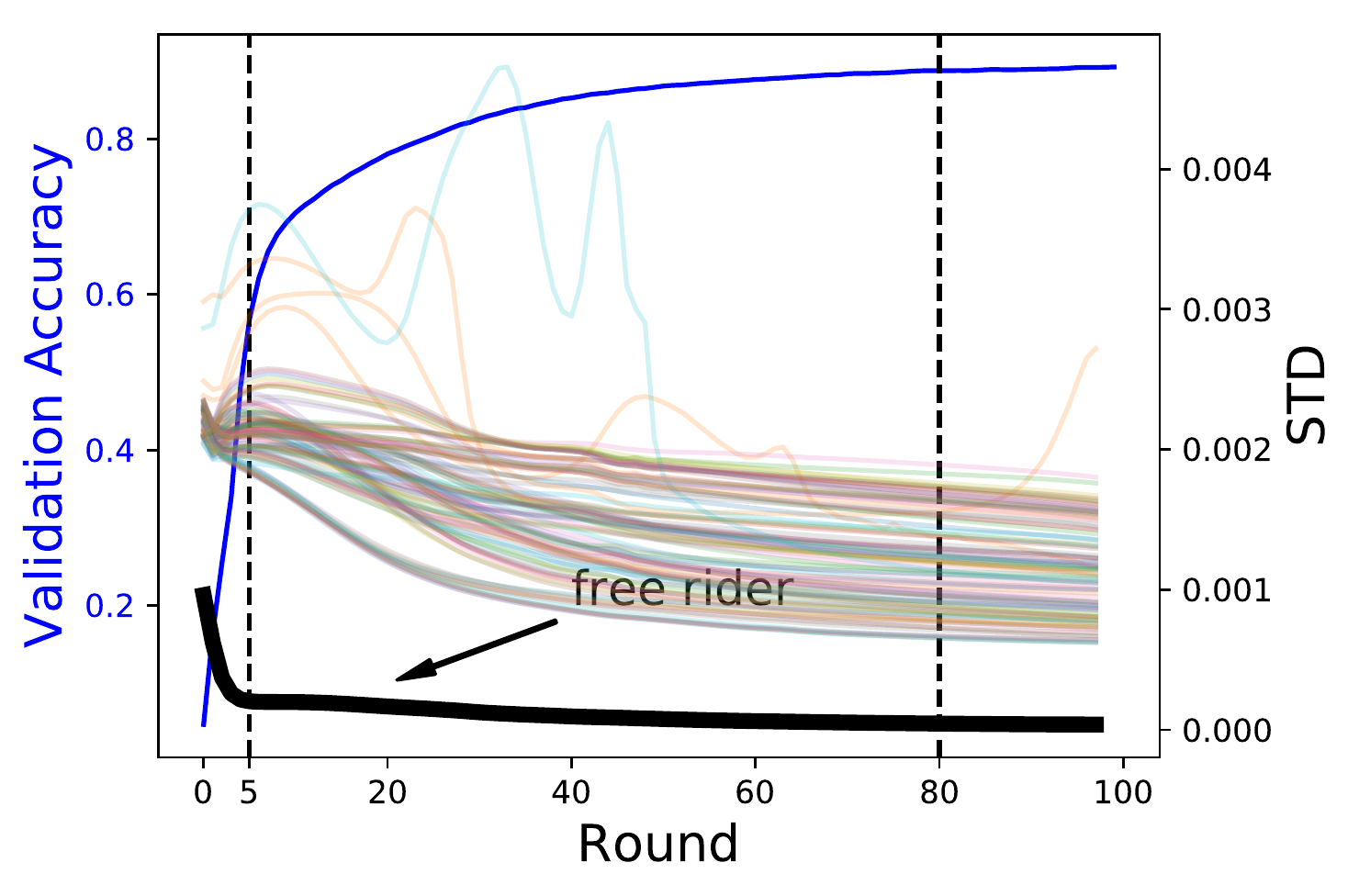}
         \caption{Standard deviation (STD) of the updates submitted by each client, with the change of training process}
         \label{fig:delta1-uneven-std}
     \end{subfigure}
     \begin{subfigure}[b]{\linewidth}
         \centering
         \begin{subfigure}[b]{0.49\linewidth}
             \includegraphics[width=\linewidth]{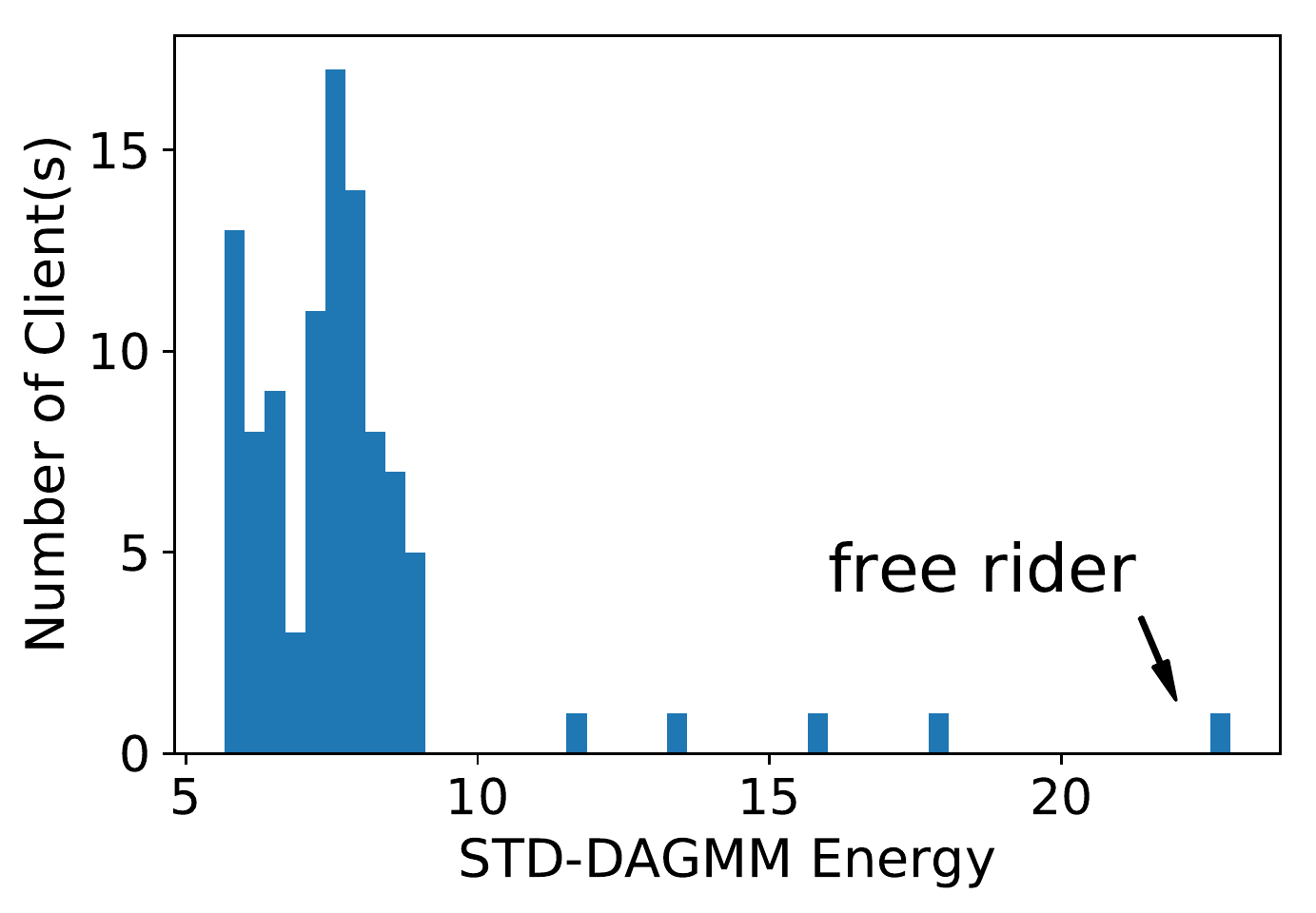}\hfill%
             \caption{Round 5}
         \end{subfigure}
         \begin{subfigure}[b]{0.49\linewidth}
             \includegraphics[width=\linewidth]{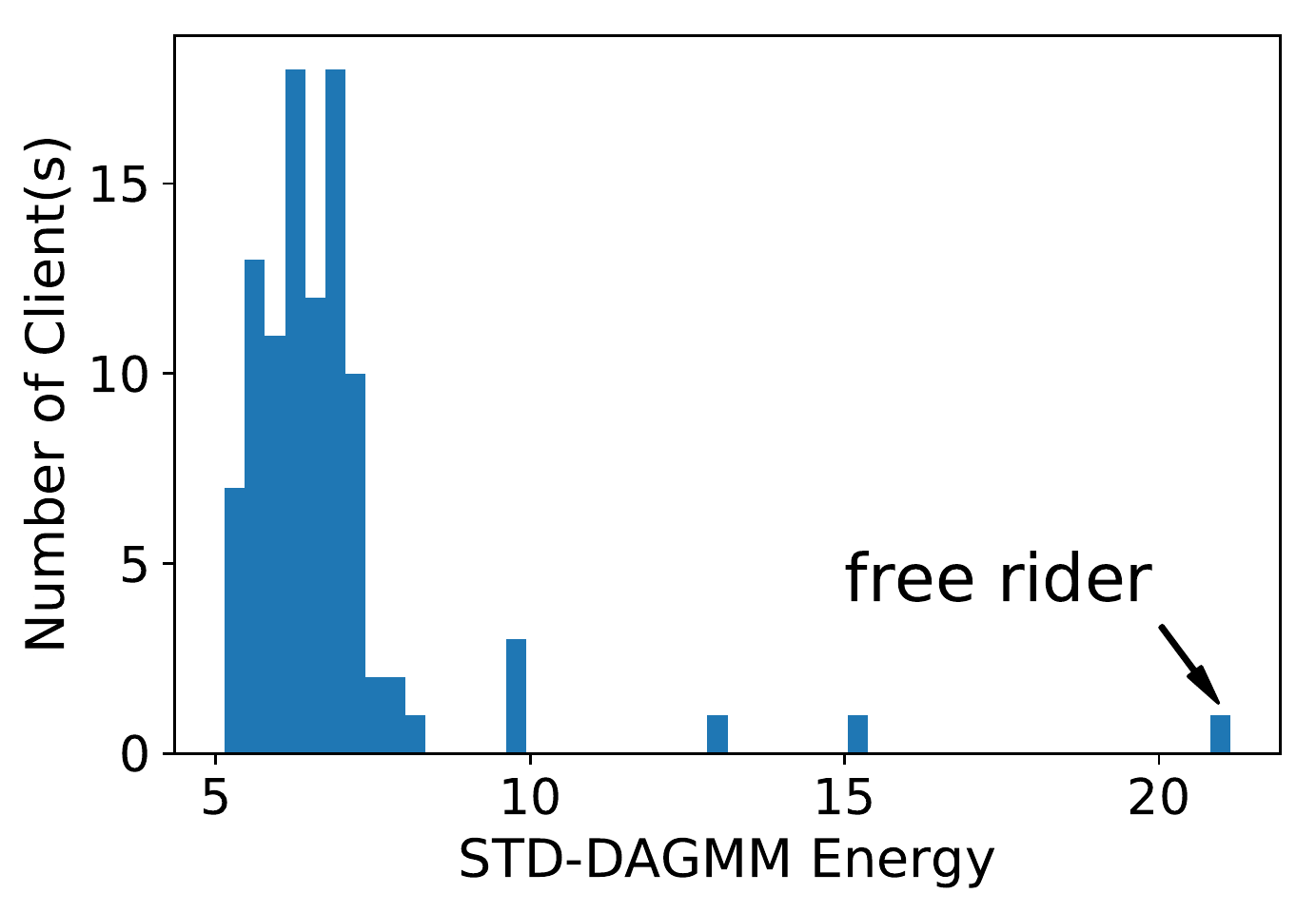}
             \caption{Round 80}
         \end{subfigure}
         \caption{STD-DAGMM - detected}
         \label{fig:delta1-uneven-stddagmm}
     \end{subfigure}
        \caption{Different local data distribution, delta weights attack with $\eta=1$}
        \label{fig:delta1-uneven}
\end{figure}

In this section, we distribute MNIST dataset into each local client, such that each client only has 1 or 2 classes, using the method described in Section~\ref{sec:random-eval-uneven}.
Without loss of generality, we show the statistics and detection results when $\eta=1$, as in Figure~\ref{fig:delta1-uneven}. Note that the STD statistics in Figure~\ref{fig:delta1-uneven-std} show that the free rider fake updates in this case have lower standard deviation compared with other clients. We think that is because all other clients only have a specific set of parameters to update, thus the submitted gradient vector is sparse and has larger STD, while the free rider, by averaging all others' updates, reduces the STD.
Figure~\ref{fig:delta1-uneven-stddagmm} indicates that our proposed STD-DAGMM method is also effective.

\subsubsection{STD-DAGMM for random weights attack detection}
\begin{figure}[hbtp]
     \centering
     \begin{subfigure}[b]{\linewidth}
         \centering
         \begin{subfigure}[b]{0.49\linewidth}
             \includegraphics[width=\linewidth]{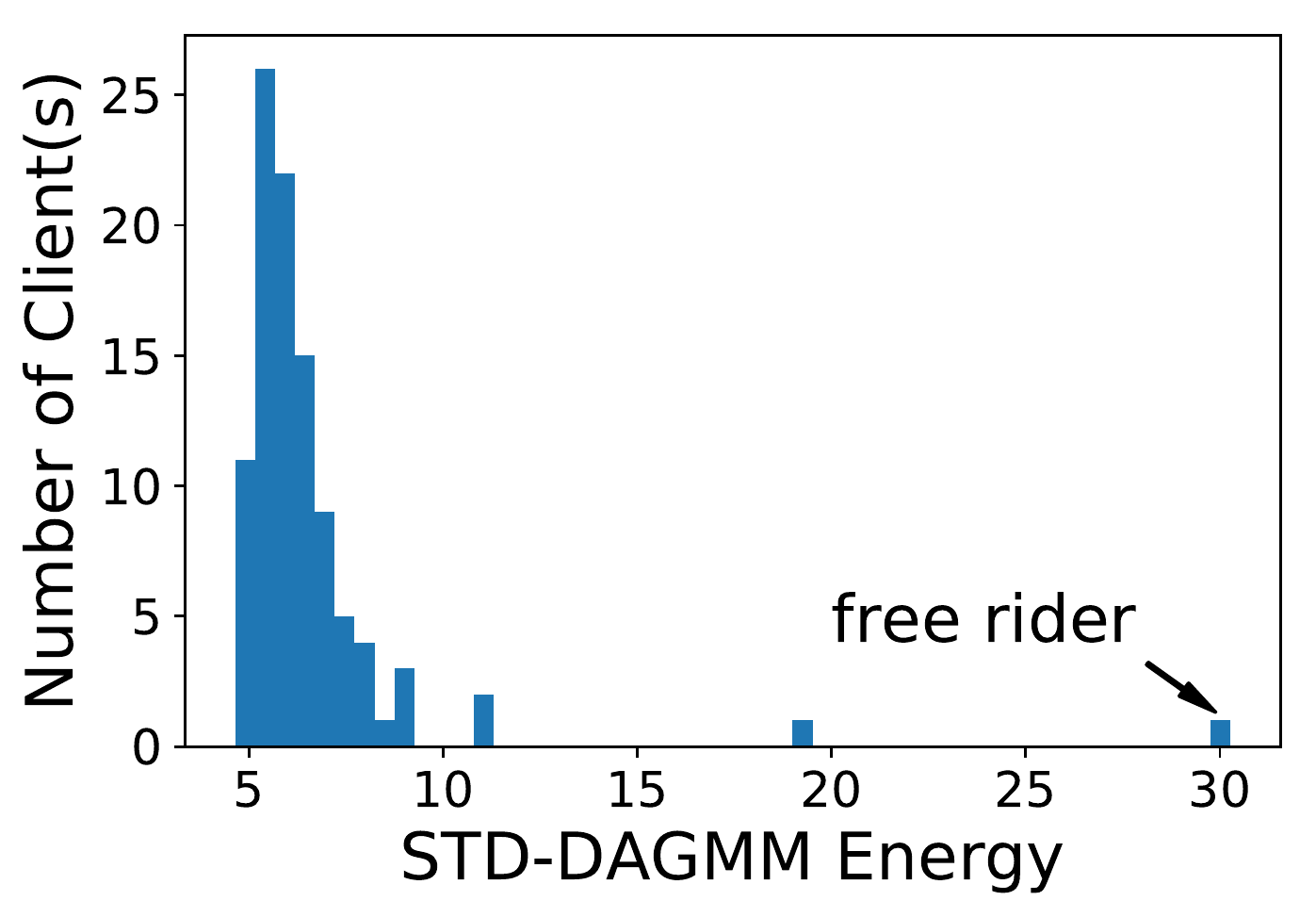}\hfill%
             \caption{Round 5}
         \end{subfigure}
         \begin{subfigure}[b]{0.49\linewidth}
             \includegraphics[width=\linewidth]{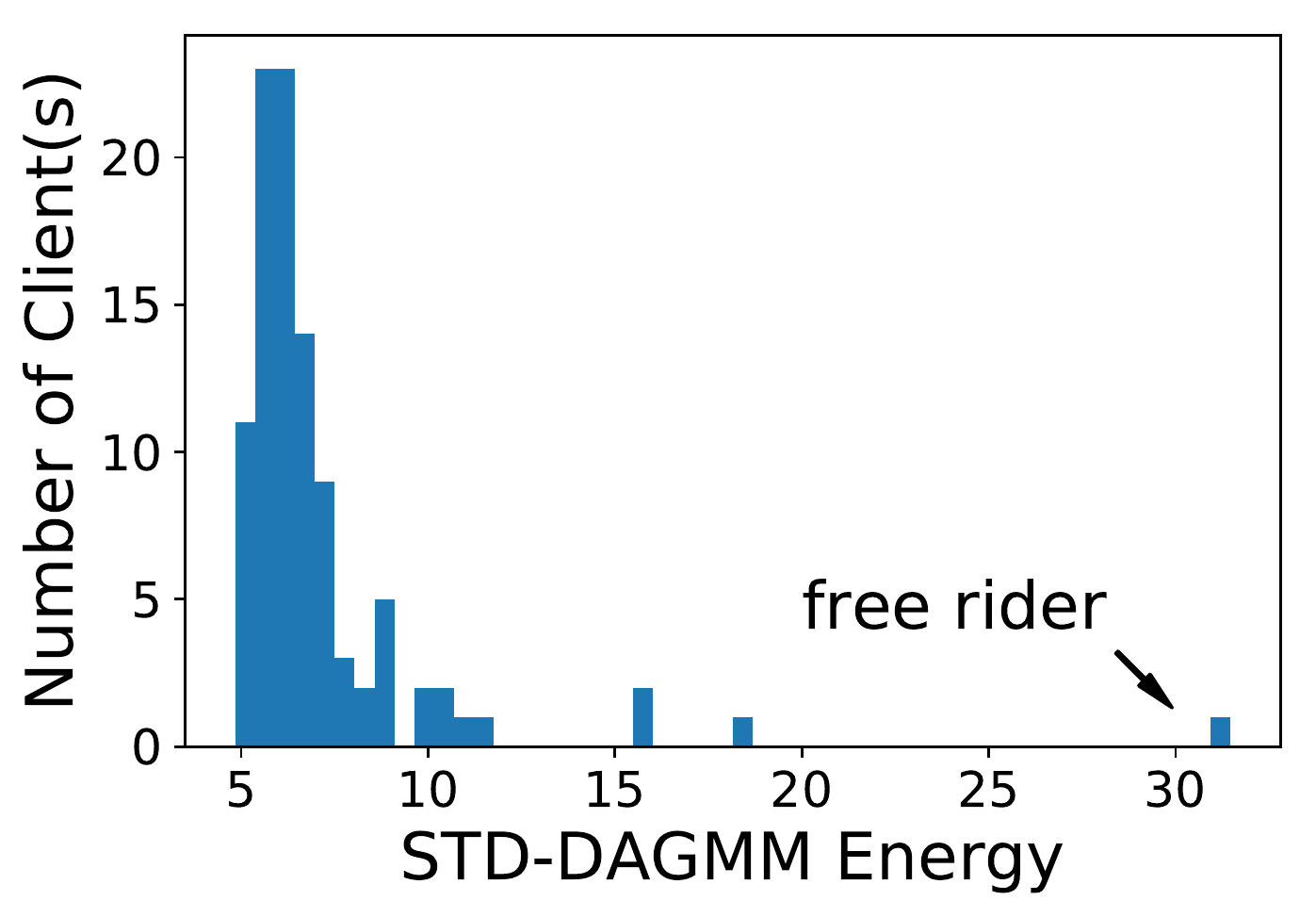}
             \caption{Round 80}
         \end{subfigure}
         \caption{STD-DAGMM - detected attack in Figure~\ref{fig:random10e-4-even}}
         \label{fig:random1e-4-even-stddagmm}
     \end{subfigure}
     \begin{subfigure}[b]{\linewidth}
         \centering
         \begin{subfigure}[b]{0.49\linewidth}
             \includegraphics[width=\linewidth]{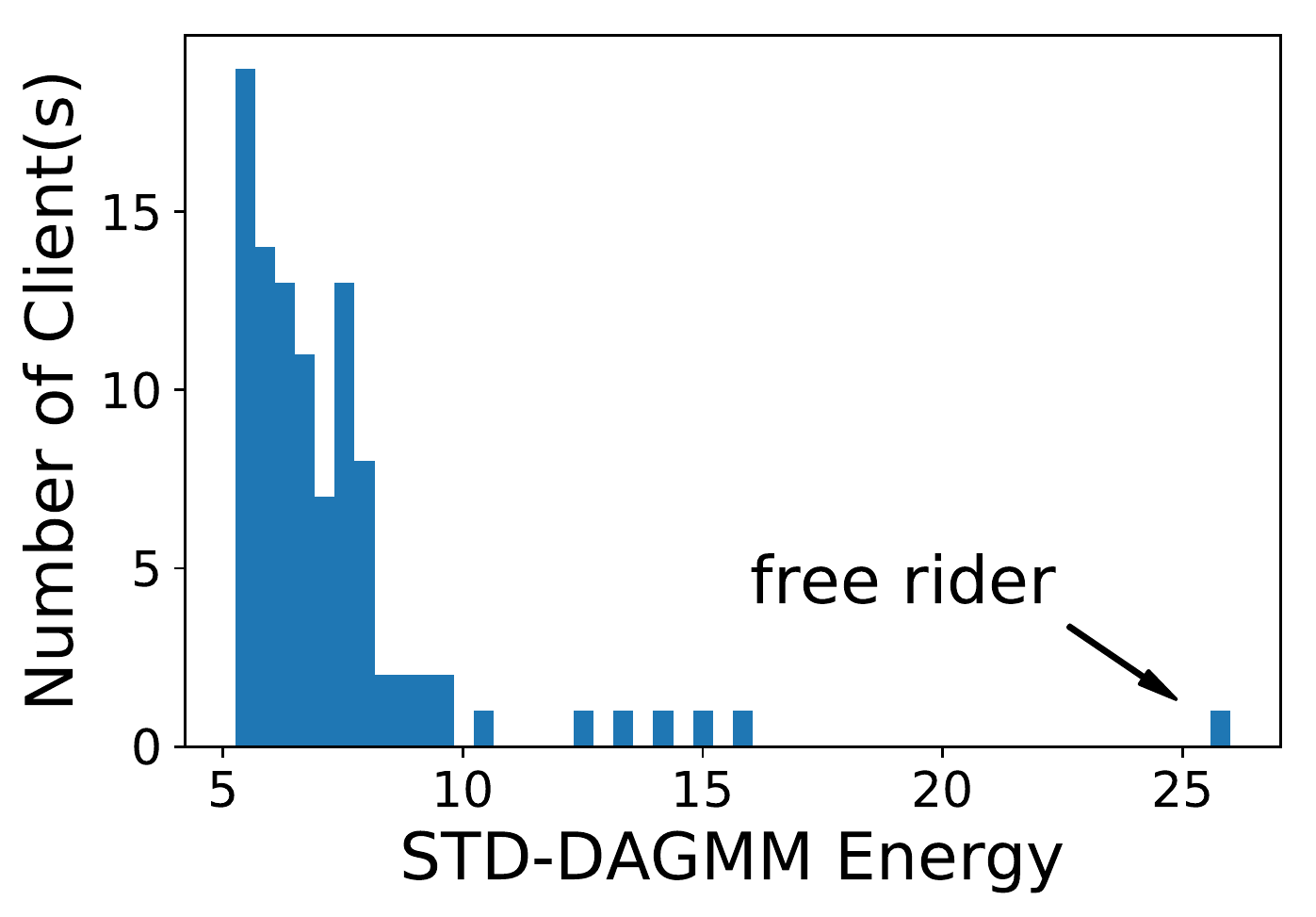}\hfill%
             \caption{Round 5}
         \end{subfigure}
         \begin{subfigure}[b]{0.49\linewidth}
             \includegraphics[width=\linewidth]{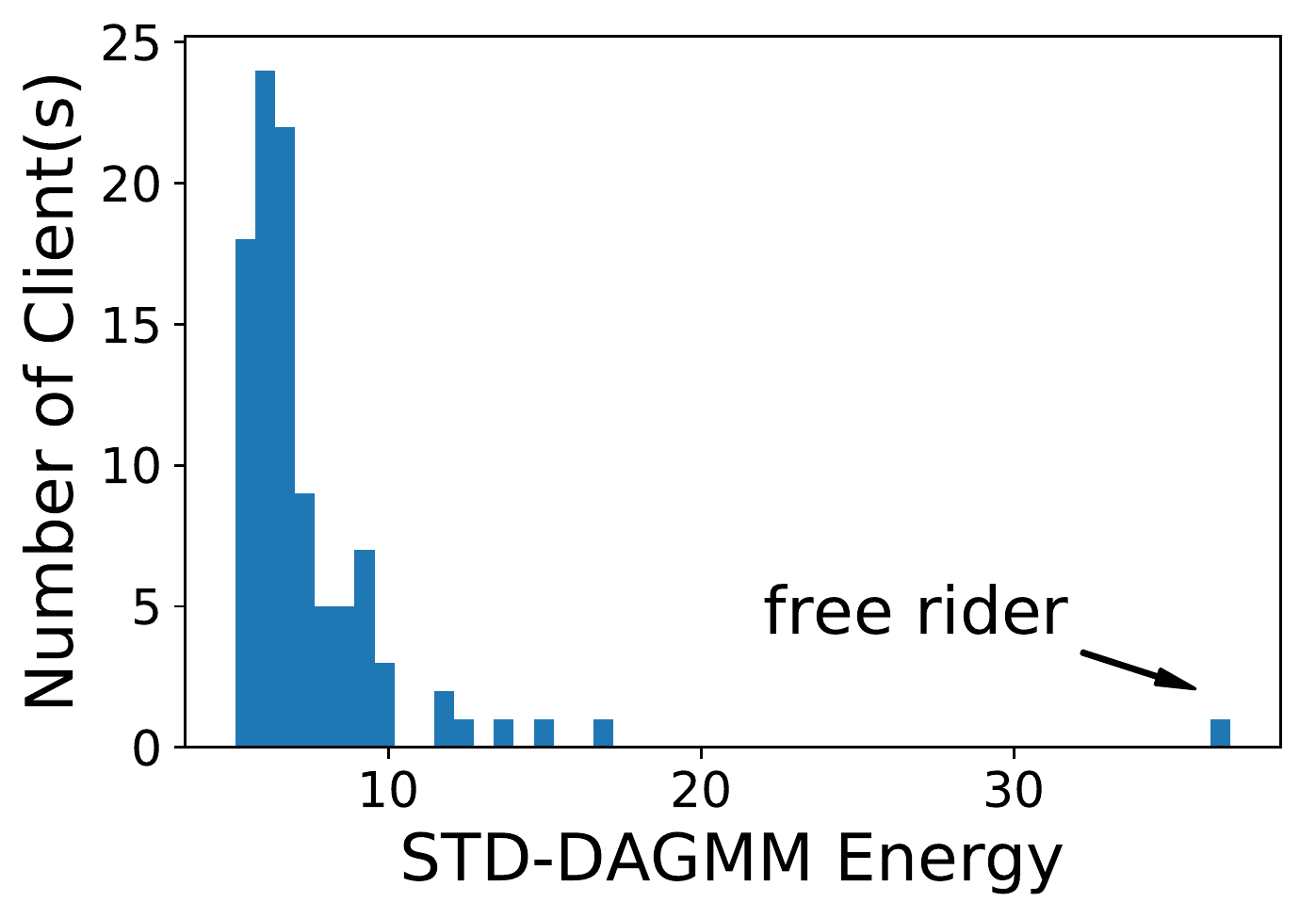}
             \caption{Round 80}
         \end{subfigure}
         \caption{STD-DAGMM - detected attack in Figure~\ref{fig:random10e-3-even}}
         \label{fig:random1e-3-even-stddagmm}
     \end{subfigure}
     \begin{subfigure}[b]{\linewidth}
         \centering
         \begin{subfigure}[b]{0.49\linewidth}
             \includegraphics[width=\linewidth]{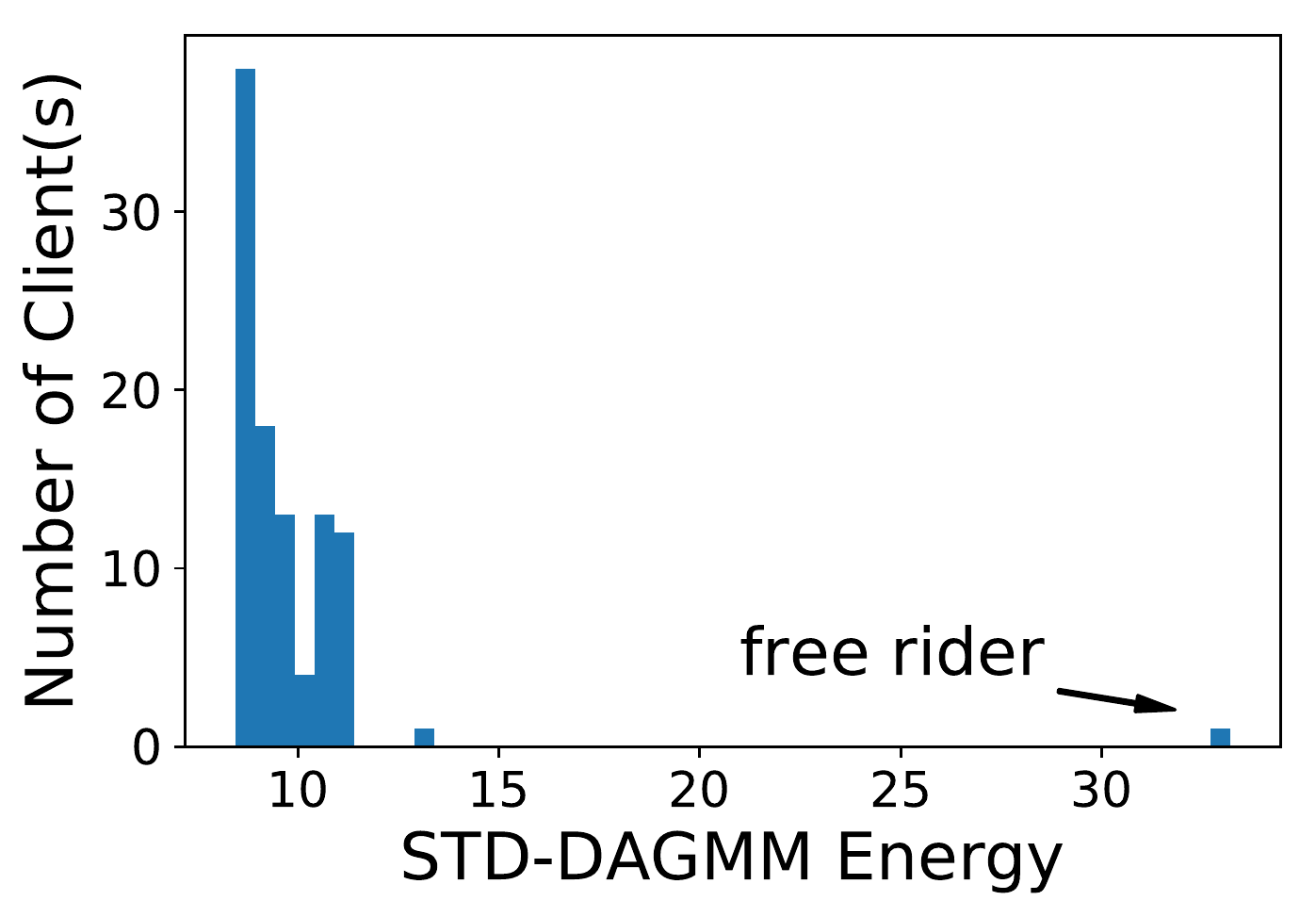}\hfill%
             \caption{Round 5}
         \end{subfigure}
         \begin{subfigure}[b]{0.49\linewidth}
             \includegraphics[width=\linewidth]{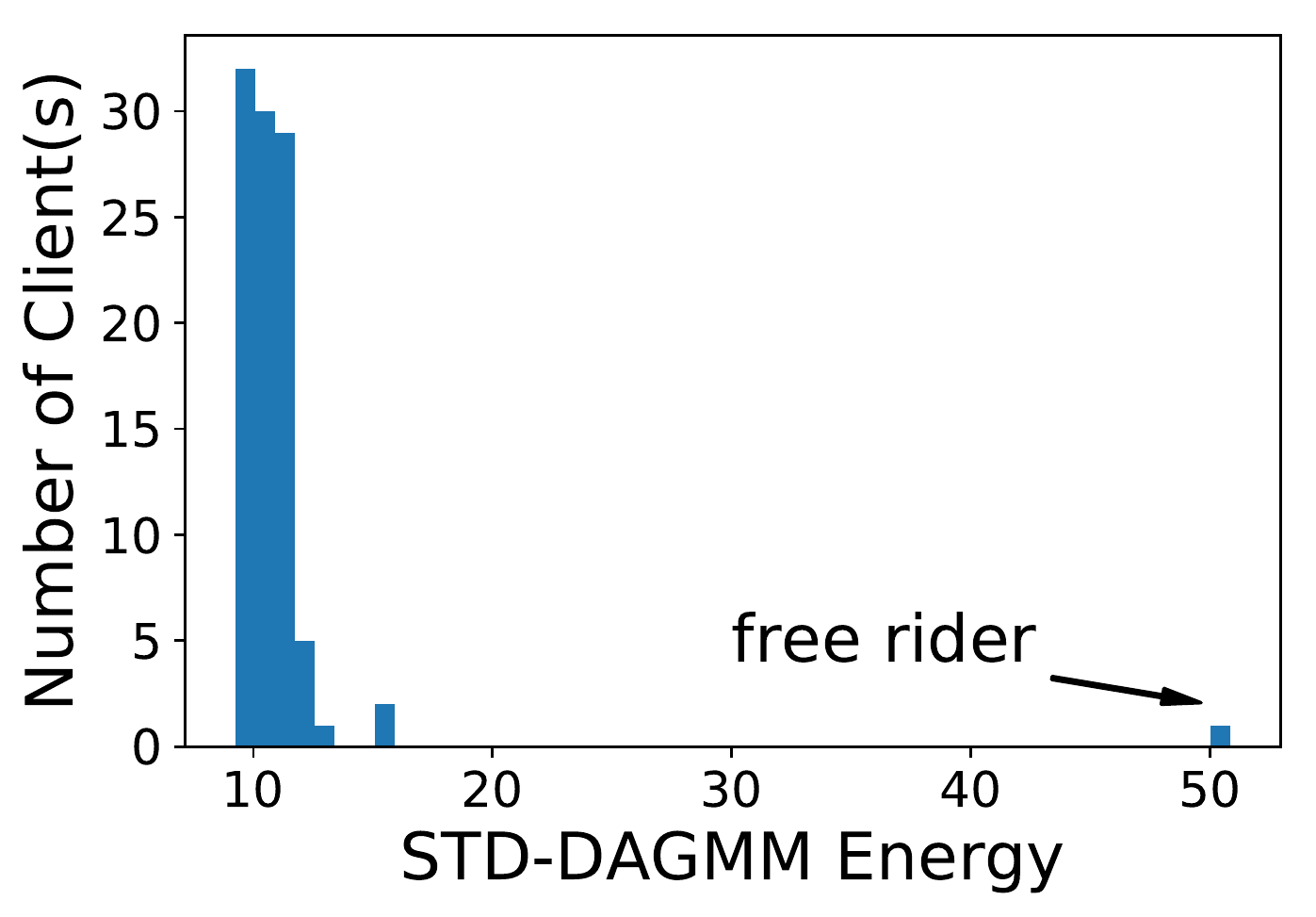}
             \caption{Round 80}
         \end{subfigure}
         \caption{STD-DAGMM - detected attack in Figure~\ref{fig:random10e-3-uneven}}
         \label{fig:random1e-3-uneven-stddagmm}
     \end{subfigure}
        \caption{Similar local data distribution, random weights attack}
        \label{fig:random-even-stddagmm}
\end{figure}

Finally, we apply STD-DAGMM method on previously evaluated random weights attack generation.
The results in Figure~\ref{fig:random-even-stddagmm} show that
STD-DAGMM is able to detect 
the random weights attack evaluated in Section~\ref{sec:random-eval} no matter whether clients have similar local data or not, which further 
demonstrates its generality in free-rider attack detection.

\section{Attack III: Advanced Delta Weights}
\label{sec:advanced-delta}
\subsection{Attack specification}
In this section, we further explore a more advanced attack a free rider may exploit, which is to add Gaussian noise onto the fake update matrix generated by delta weights attack presented in Section~\ref{sec:delta}.
A free rider may benefit from the Gaussian noise addition in the following two ways.
First, we have mentioned in Section~\ref{sec:delta-eval} that if multiple free riders all take the delta weights generation approach, the detection would be trivial since all such free riders have exactly the same weights. As a result, a smart attacker may attempt to avoid this by adding random Gaussian noise onto its generated delta weights. Second, note that in previous section, we have demonstrated that having lower STD than others is one important reason leading to the detection of the free rider. Therefore, a free rider may try to avoid detection by adding Gaussian noise with zero mean and a certain STD, such that the resulted gradient update matrix has similar STD with other clients.


\paragraph{Advanced delta weights attack.}
Following delta weights generation in Section~\ref{sec:delta}, the advanced delta weights attack takes a further step by adding Gaussian noise having mean 0 and standard deviation $\sigma$ to the delta weights. Specifically, a fake gradient update is generated by
\begin{equation}
G_{i, j}^{f}=\eta \cdot \frac{1}{n} \sum_{x=1}^{n} G_{x, j-1}+N(0, \sigma) \;\;\;\;\;\;\mathrm{(ref. Equation~\ref{eq:delta})}
\end{equation}

\subsection{Defense strategy}
We find that our proposed STD-DAGMM approach continuously works well for this kind of attack.

\subsection{Experimental validation}
\label{sec:advanced-delta-eval}
\begin{figure}[hbtp]
     \centering
     \begin{subfigure}[b]{\linewidth}
         \centering
         \includegraphics[width=\textwidth]{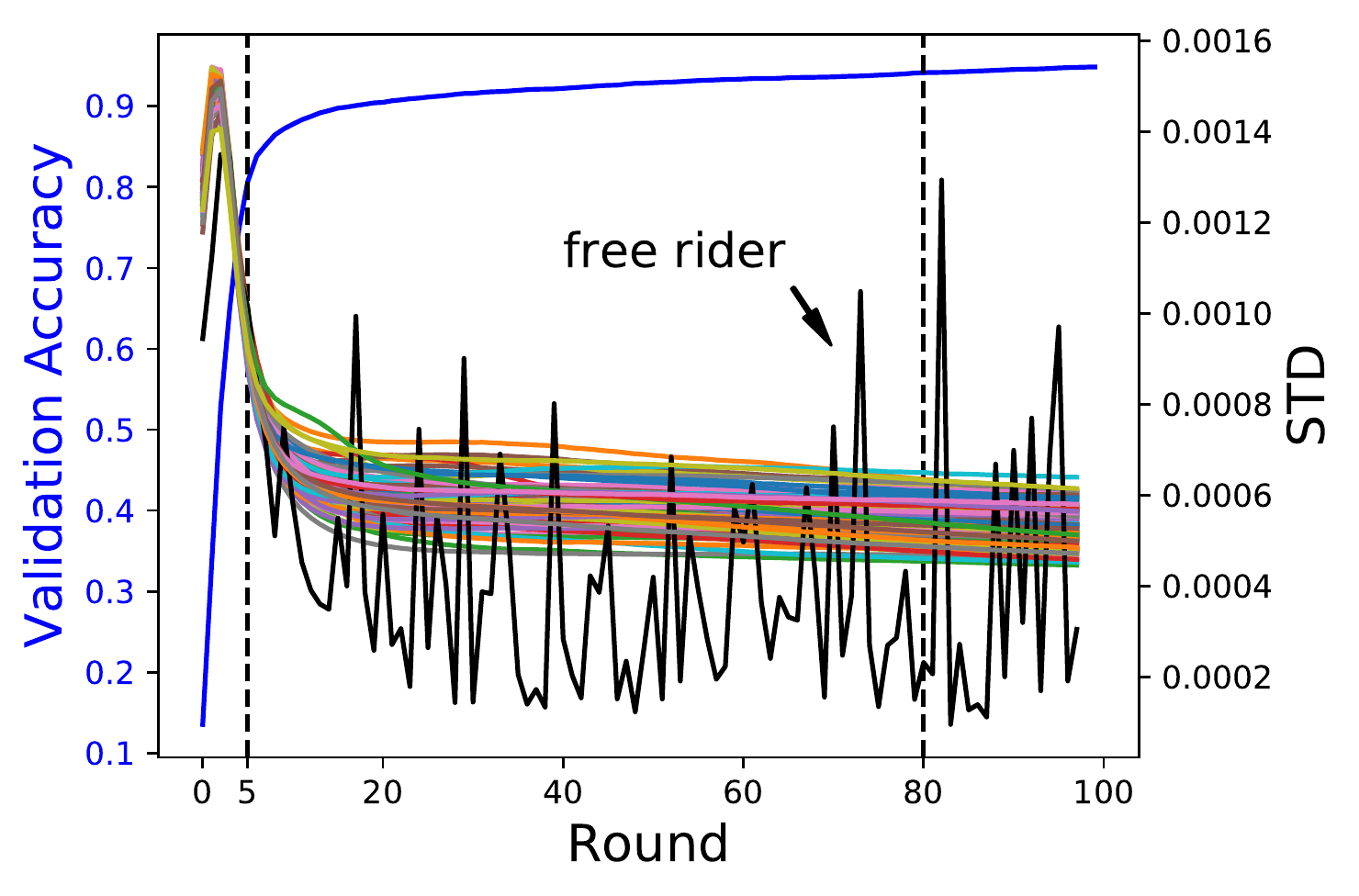}
         \caption{Standard deviation (STD) of the updates submitted by each client, with the change of training process}
         \label{fig:1e-3-even-std}
     \end{subfigure}
     \begin{subfigure}[b]{\linewidth}
         \centering
         \begin{subfigure}[b]{.48\textwidth}
             \includegraphics[width=\linewidth]{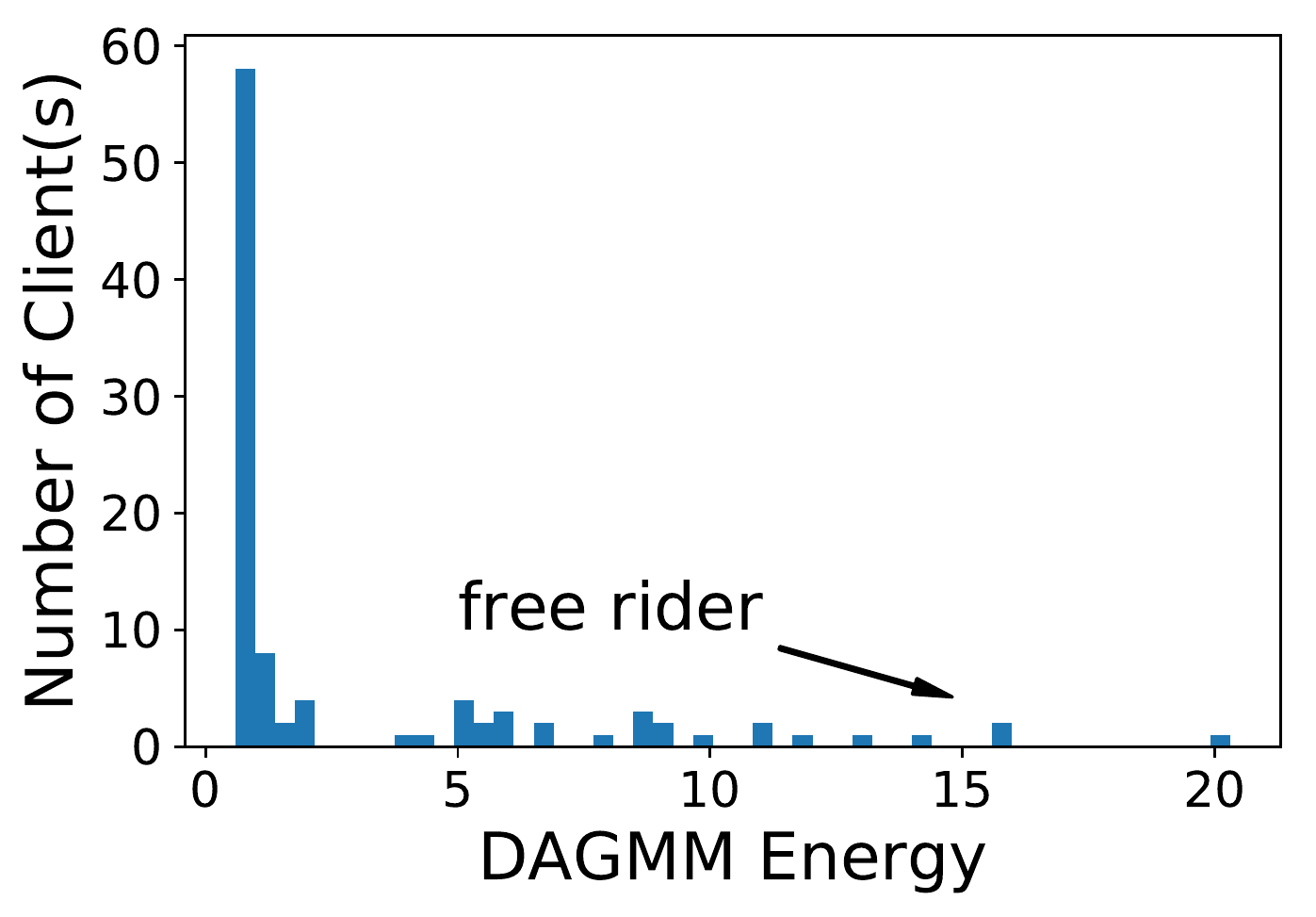}\hfill%
             \caption{Round 5}
         \end{subfigure}
         \begin{subfigure}[b]{.48\textwidth}
             \includegraphics[width=\linewidth]{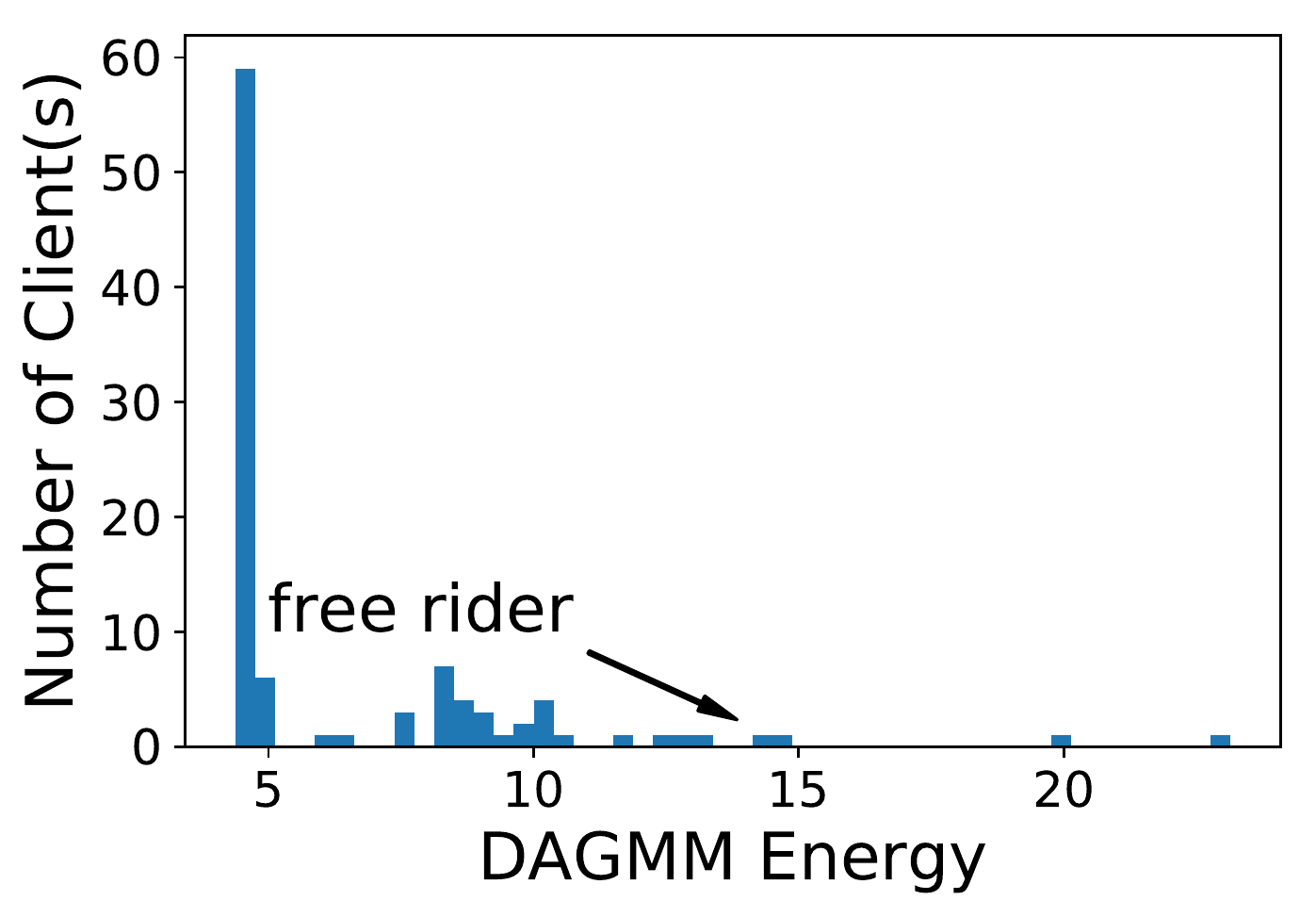}
             \caption{Round 80}
         \end{subfigure}
         \caption{DAGMM - not detected}
         \label{fig:1e-3-even-dagmm}
     \end{subfigure}
     \begin{subfigure}[b]{\linewidth}
         \centering
         \begin{subfigure}[b]{.48\textwidth}
             \includegraphics[width=\linewidth]{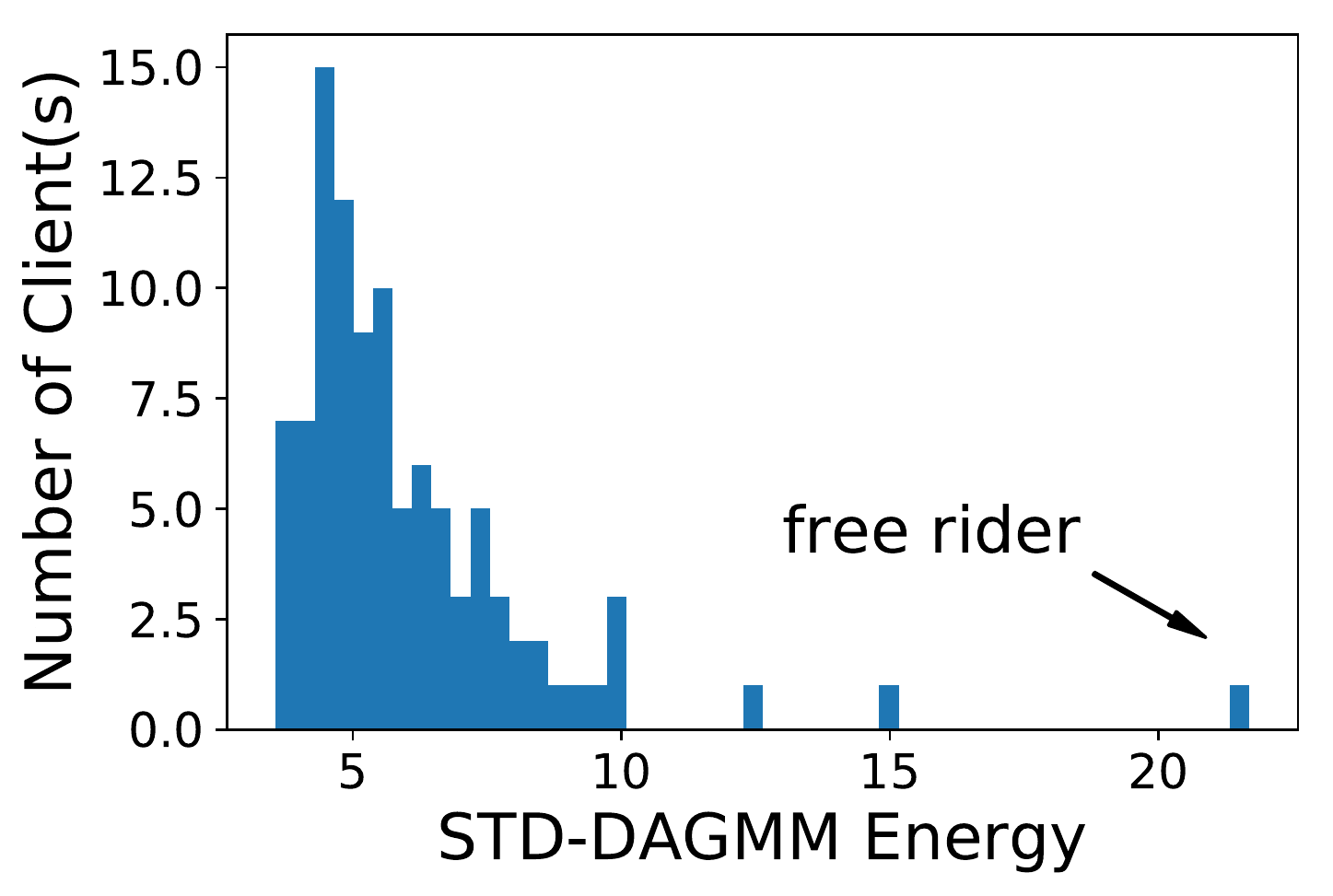}\hfill%
             \caption{Round 5}
         \end{subfigure}
         \begin{subfigure}[b]{.48\textwidth}
             \includegraphics[width=\linewidth]{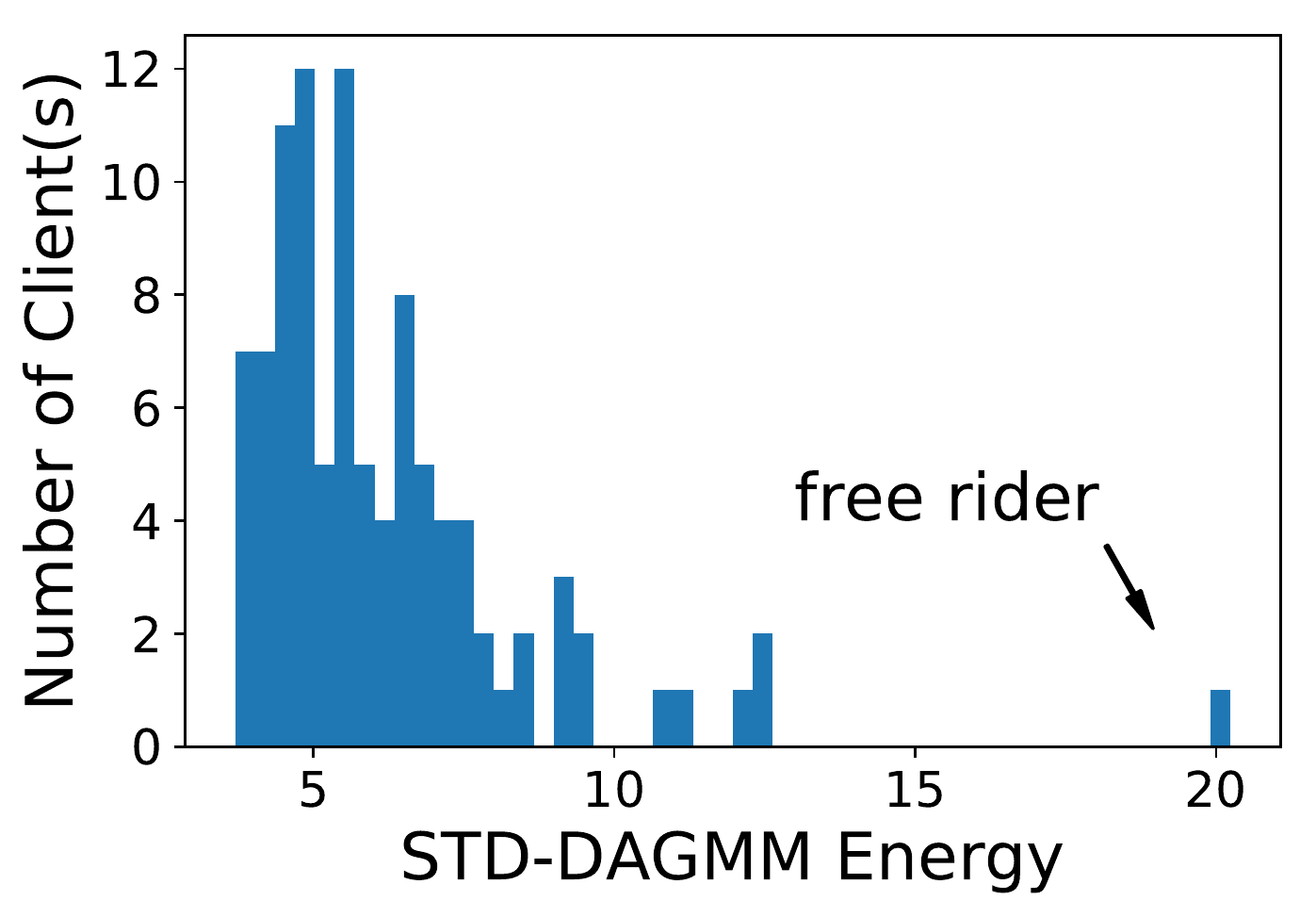}
             \caption{Round 80}
         \end{subfigure}
         \caption{STD-DAGMM - detected}
         \label{fig:1e-3-even-variance-dagmm}
     \end{subfigure}
        \caption{Similar local data distribution, advanced delta weights attach with $\sigma=10^{-3}$}
        \label{fig:deltaad10e-3-even}
\end{figure}

Intuitively, the effectiveness of the free rider detection method would be affected by the choice of $\sigma$ a free rider may choose.
Although it's hard to guess the standard deviation of other normal clients' gradient updates at each round, we assume that a strong attacker has previous knowledge of the training process so that he can submit weights with STD similar to benign clients' weights in most of the rounds. 
In our experiments, we tried different standard deviation for Gaussian noise, and find that $\sigma=10^{-3}$ is among the best choices for the free rider.
Specifically, as shown in Figure~\ref{fig:1e-3-even-std}, adding a Gaussian noise $\mathbb{N}(0, 10^{-3})$ to the delta weights of two previous global models will result in a fake gradient update having similar STD with other clients. 
Moreover, we show that this attack indeed makes the free rider harder to be detected, as in Figure~\ref{fig:1e-3-even-dagmm}. For DAGMM method, the energy value of the free rider is completely hidden among normal clients, for both round 5 and round 80. 
Nevertheless, our proposed STD-DAGMM method is still effective in detecting this type of advanced delta attack, as shown in Figure~\ref{fig:1e-3-even-variance-dagmm}, where the energy value for free rider is the biggest, and a wide range of threshold values suffice to separate the free rider from other clients.

\section{Extension}
In this section, we present some real-world settings and discuss how our proposed defense strategy would work under each setting.
In reality, many factors may affect the proposed attack and defense mechanisms.
For example, the number of free riders in the federated learning participants is unknown to the parameter server. It would be interesting to explore how 
STD-DAGMM method would work with more free riders present.
Recently, many privacy enhancing techniques have been proposed~\cite{DBLP:journals/corr/abs-1712-07557,dwork2006differential,DBLP:journals/corr/BonawitzIKMMPRS16}, in order to address the
increasing privacy concern in federated learning and machine learning models in general~\cite{DBLP:journals/corr/abs-1807-11655,DBLP:journals/corr/PapernotMSW16,DBLP:journals/corr/abs-1811-09712}.
These methods such as differential privacy, which works by applying random Gaussian noise to the model parameters, would influence the effectiveness of the proposed attack, and hence the detection mechanism as well.


\subsection{More free riders}
\label{sec:more-free-riders}

This section explores the effectiveness of our proposed defense methods, in particular, STD-DAGMM, in the face of more free riders.

\begin{figure}[hbtp]
     \centering
     \begin{subfigure}[b]{\linewidth}
         \centering
         \includegraphics[width=\linewidth]{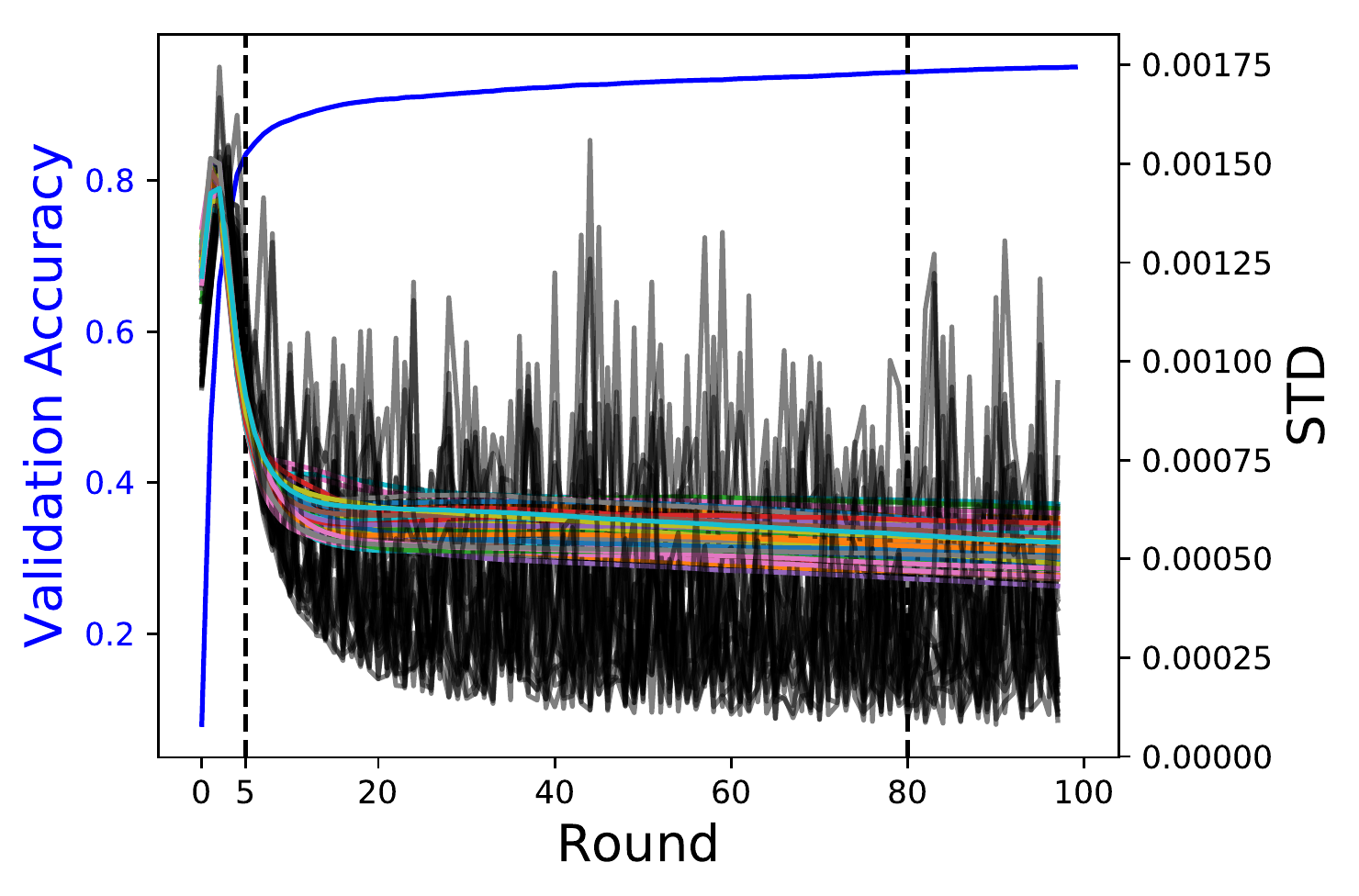}
         \caption{Standard deviation (STD) of the updates submitted by each client at different rounds}
         \label{fig:std-20fds}
     \end{subfigure}
     \begin{subfigure}[b]{\linewidth}
         \centering
         \begin{subfigure}[b]{.49\linewidth}
             \includegraphics[width=\linewidth]{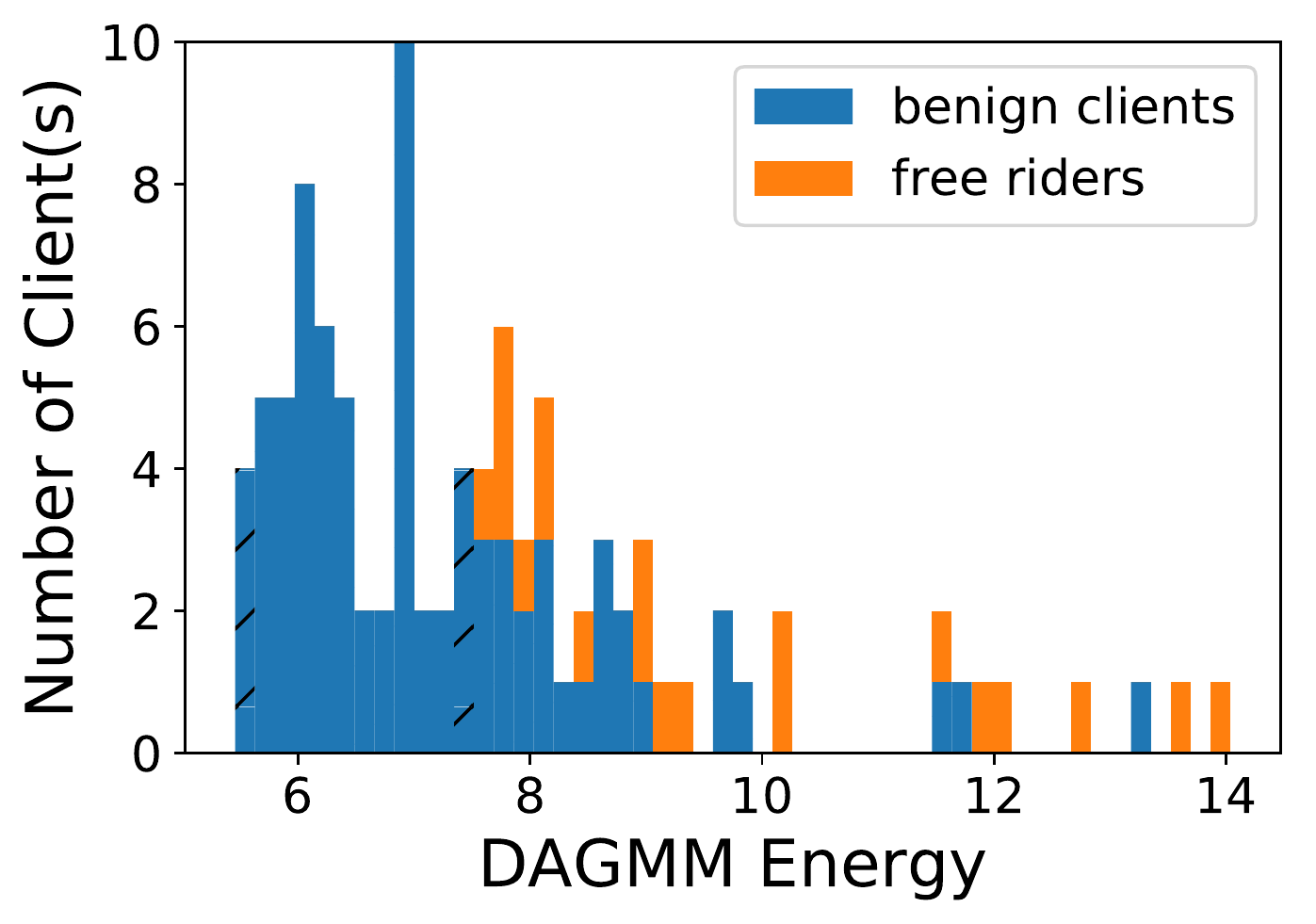}\hfill%
             \caption{Round 5 (AUC=0.8856)}
         \end{subfigure}
         \begin{subfigure}[b]{.49\linewidth}
             \includegraphics[width=\linewidth]{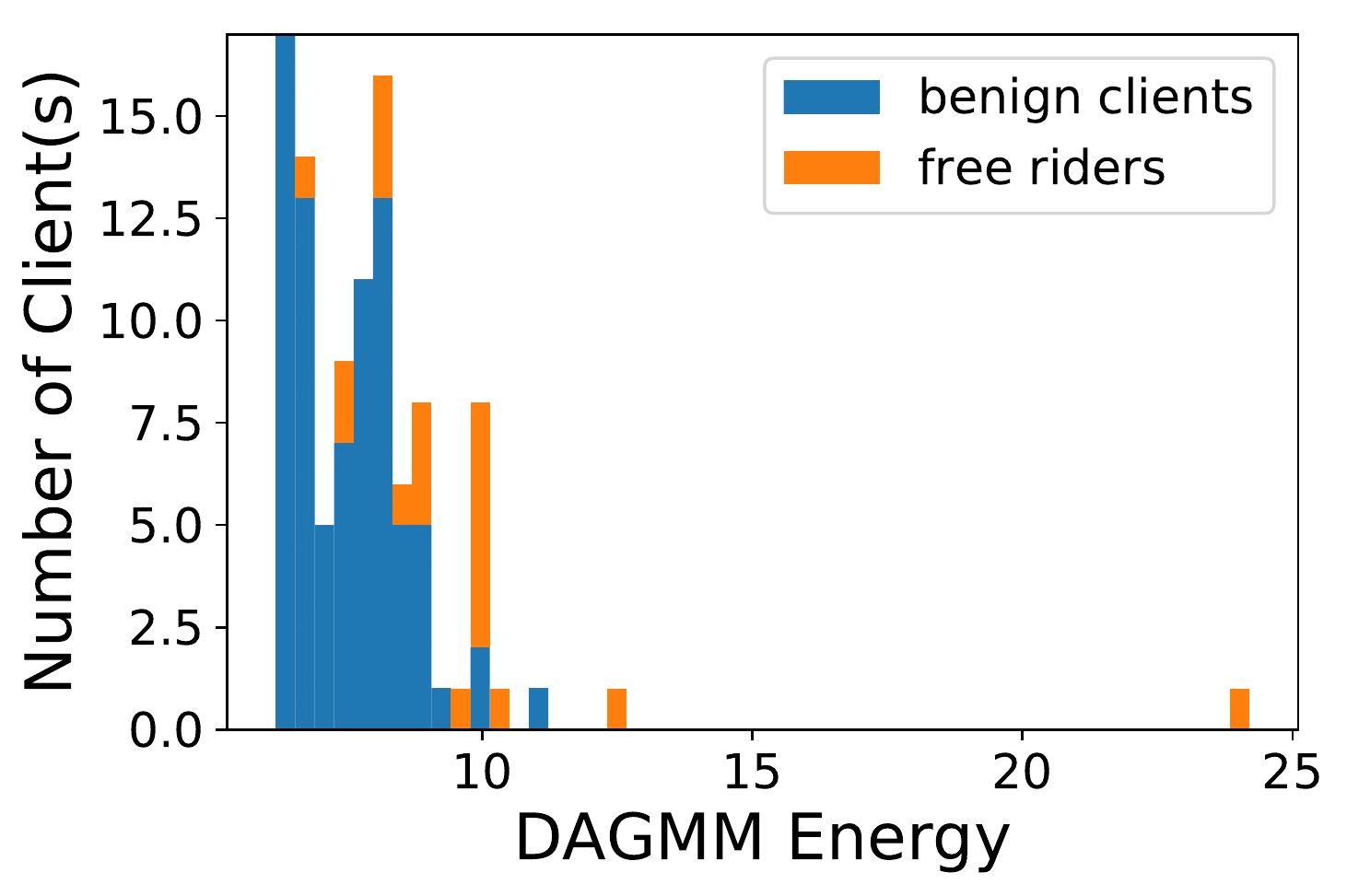}
             \caption{Round 80 (AUC=0.8487)}
         \end{subfigure}
         \caption{DAGMM}
         \label{fig:DAGMM-20fds}
     \end{subfigure}
     \begin{subfigure}[b]{\linewidth}
         \centering
         \begin{subfigure}[b]{.49\linewidth}
             \includegraphics[width=\linewidth]{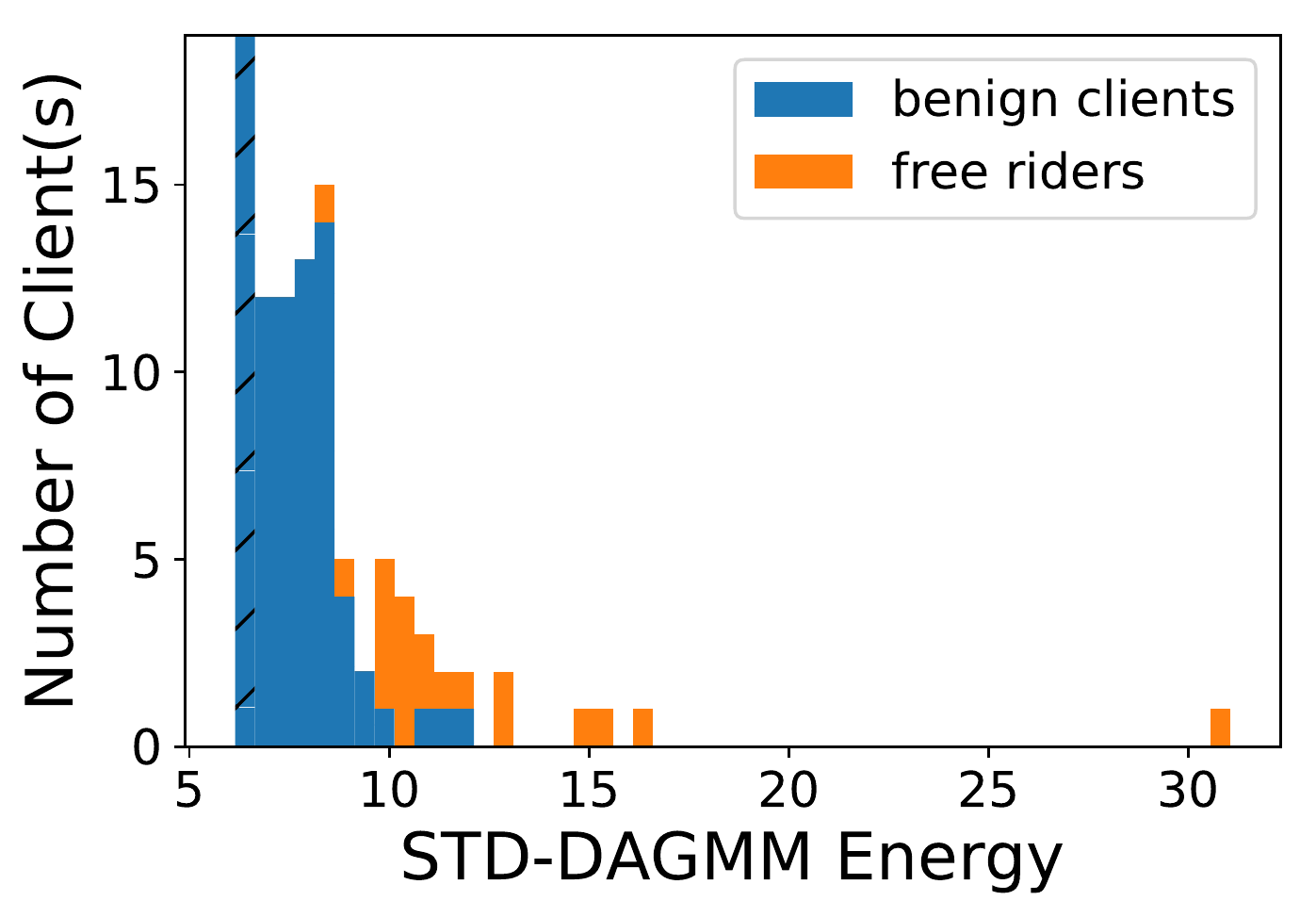}\hfill%
             \caption{Round 5 (AUC=0.9644)}
         \end{subfigure}
         \begin{subfigure}[b]{.49\linewidth}
             \includegraphics[width=\linewidth]{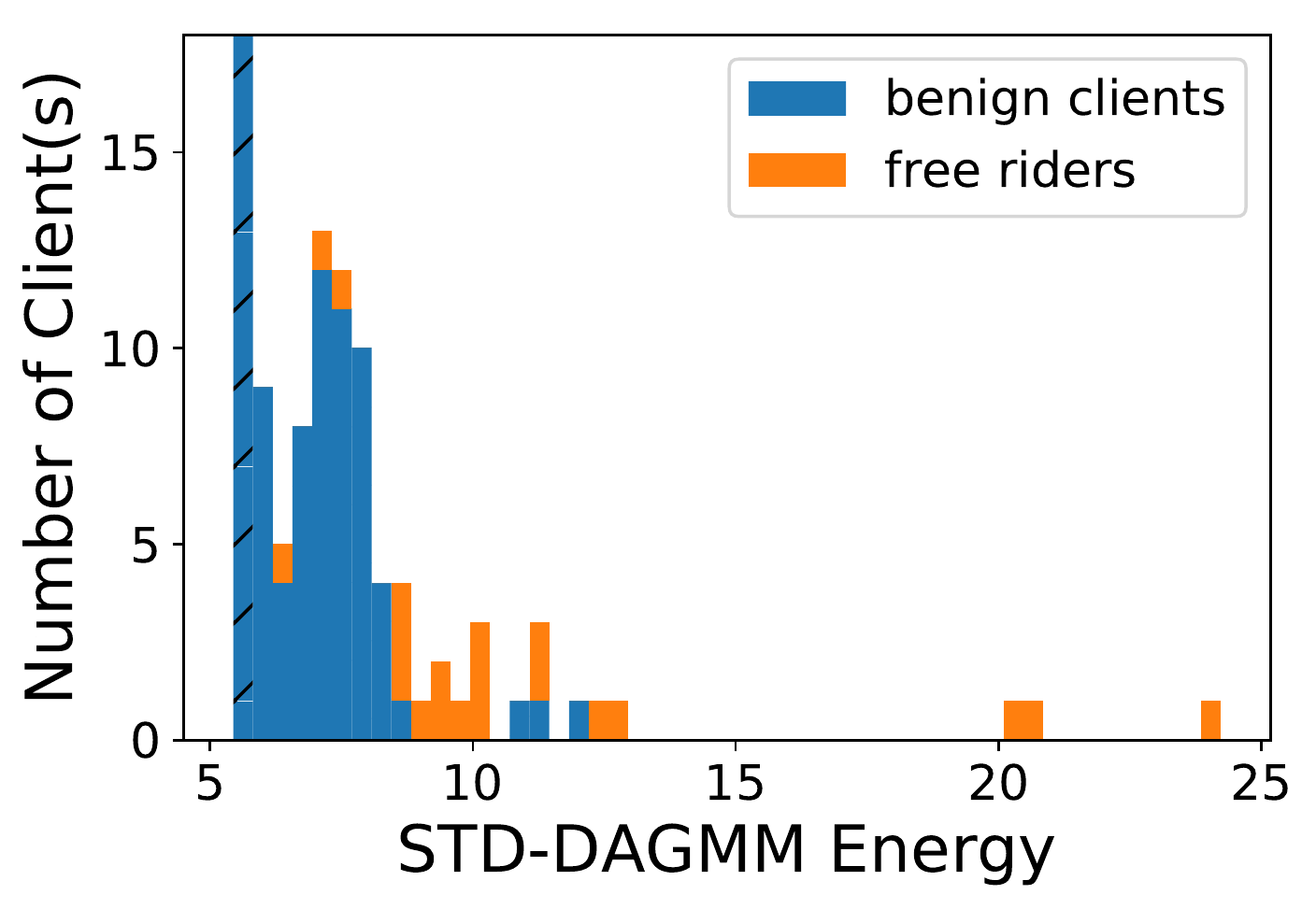}
             \caption{Round 80 (AUC=0.9088)}
         \end{subfigure}
         \caption{STD-DAGMM}
         \label{fig:STD-DAGMM-20fds}
     \end{subfigure}
     \caption{20 free riders, similar local data distribution, advanced delta weights attack with $\sigma=10^{-3}$}
\end{figure}

Without loss of generality,
we first consider a total of 20 free riders out of 100 clients, each of which explores the strongest attack presented earlier - advanced delta weights attack. In particular, each free rider first calculates a delta weights by subtracting the global models in two previous rounds, and adds that with a Gaussian random noise $N(0, \sigma)$ to make its standard deviation look similar to other normal clients.
For the MNIST dataset we explore, we find that $\sigma=10^{-3}$ is among the best choices in Section~\ref{sec:advanced-delta-eval}, which we adopt here.
As shown in Figure~\ref{fig:std-20fds}, the black lines are the 20 free riders while the colored lines represent the benign clients. The higher variation in STD of free riders across the training rounds are due to the random noise sampled from a Gaussian distribution. 
We present the detection results for DAGMM and STD-DAGMM in Figure~\ref{fig:DAGMM-20fds} and Figure~\ref{fig:STD-DAGMM-20fds} respectively. We use yellow bars to represent the histogram for free riders and blue bars for other normal clients. For the bins of energy values that are duplicated in the figures, we stack free riders' counts above normal ones. 
As in the figures, there is no single threshold that works to separate free riders and normal clients entirely, either for DAGMM or STD-DAGMM. However, we can observe that for STD-DAGMM, more free riders have larger energy values, compared with the ones for DAGMM.
To compare the performance more clearly, we compute the AUC score, which is the \textit{area under the curve} that plots the true positive rate change with the false positive rate, with respect to different thresholds to separate abnormal/normal samples.
The higher AUC score is, the clearer the separation is, and the better the detection method.
As the numbers shown under each figure in Figure~\ref{fig:DAGMM-20fds} and Figure~\ref{fig:STD-DAGMM-20fds}, STD-DAGMM works better than DAGMM for both round 5 (beginning of the training iteration) and round 80 (end of the training iteration.). 

\begin{figure}[hbtp]
         \centering
         \begin{subfigure}[b]{\linewidth}
             \includegraphics[width=\linewidth]{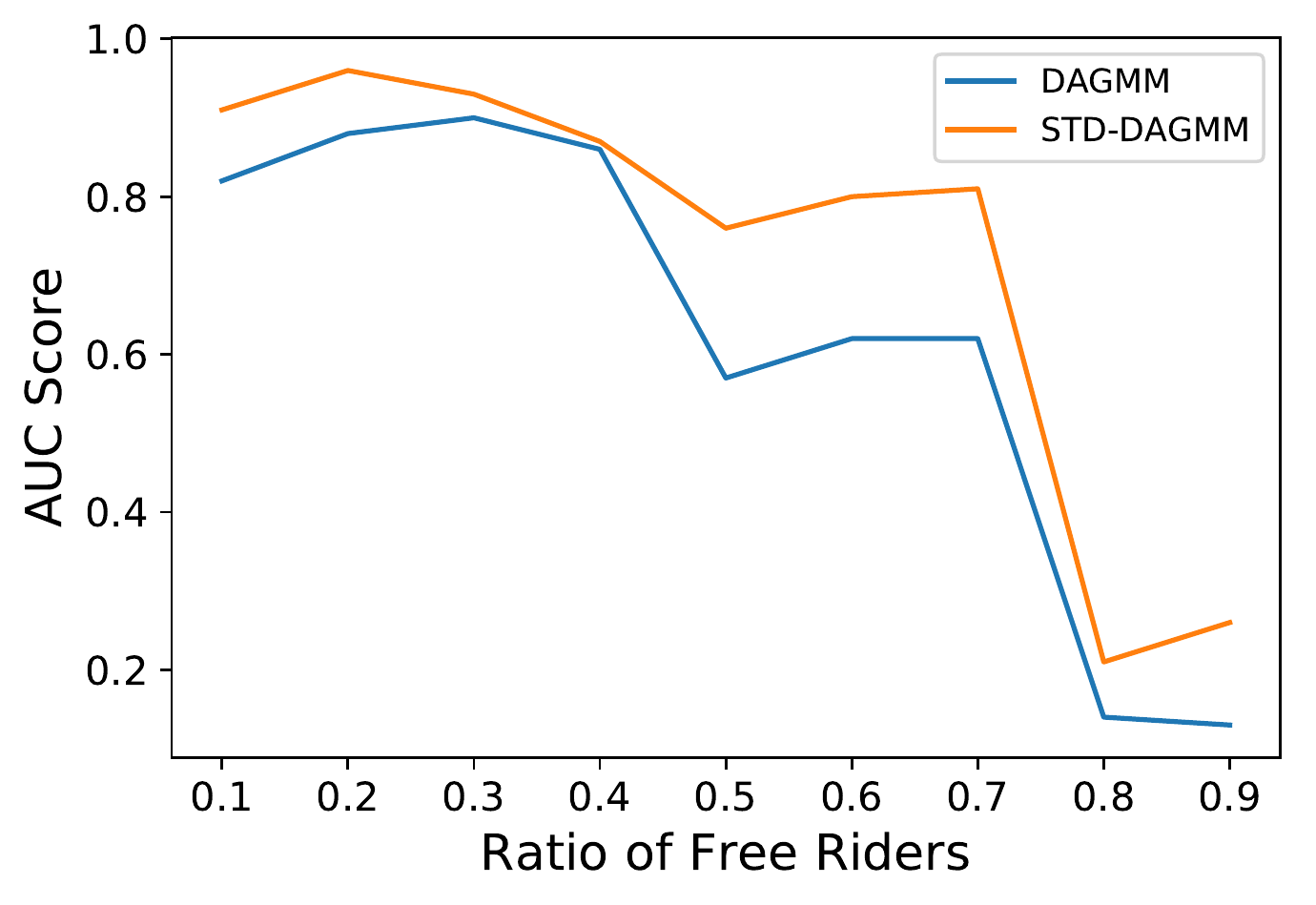}\hfill%
             \caption{Round 5}
         \end{subfigure}
         \begin{subfigure}[b]{\linewidth}
             \includegraphics[width=\linewidth]{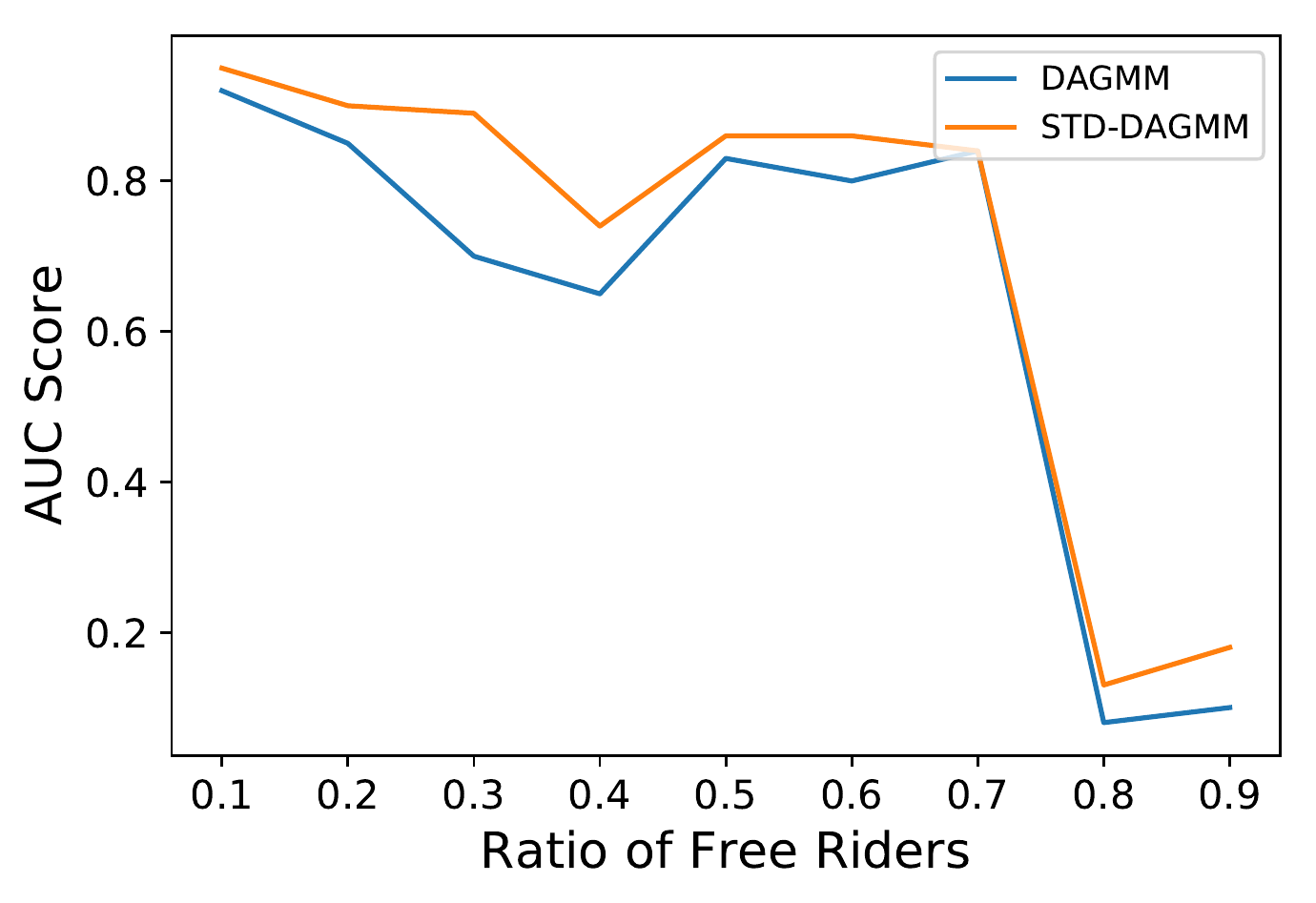}
             \caption{Round 80}
         \end{subfigure}
     \caption{AUC scores of different detection methods with varying ratios of free riders, 100 clients in total, similar local data distribution, advanced delta weights attack with $\sigma=10^{-3}$}
         \label{fig:ratio-fd}
\end{figure}

In reality, the defense mechanism wouldn't know if there are free riders or how many free riders there are in the system. Therefore, we further conduct an experiment to test the effectiveness of STD-GAGMM with varying ratios of free riders. The AUC scores for DAGMM and STD-DAGMM with increasing number of free riders are as shown in Figure~\ref{fig:ratio-fd}.
As one would imagine, as the ratio of free riders in the system becomes larger, it becomes more difficult for the defense mechanism to detect the free riders, validated by the decreasing curves in Figure~\ref{fig:ratio-fd}. Nevertheless, STD-DAGMM outperforms DAGMM regardless of free rider ratios.

We note that there is still room for improvement in order to detect all free riders without too many false positives, which we leave as our future work.



\subsection{With differential privacy}
\label{sec:ext-dp}
The concept of differential privacy (DP) was proposed in \cite{dwork2006differential} and later extended for deep learning in \cite{45428} and federated learning in \cite{DBLP:journals/corr/abs-1712-07557}.
In this section, we consider the proposed DP usage in federated learning setting \cite{DBLP:journals/corr/abs-1712-07557}, where each client clips its local gradient updates with a clipping bound, and the parameter server adds DP noise to the aggregated model, to form the new global model.
Although the clipping mechanism each client applies could potentially render local updates not that different with other clients, we find that for federated learning, another mechanism called \textit{privacy amplification} influences the defense mechanism the most, and could in fact favor the detection. 

The idea of privacy amplification is: in each round, instead of all clients participating in the training process, each client chooses whether to join with a ratio $q$. 
Remember that the goal of differential privacy is to hide the participance of each local client. 
If some adversary has to guess if a client joins a round of training, the probability he guesses it right degrades by $q$ under privacy amplification, which helps privacy.
With this, a free rider will only join the model training every $1/q$ rounds, 
and receive the global model when it's his turn.
If a free rider were to launch delta weights attack in this case, he could only subtract two previously received global model weights, and possibly divide that by the number $1/q$, to approximate other clients' updates.
In this case, the constructed free rider update by delta weights attack is more different than other clients' normal submissions, compared with the case where the delta weights are calculated by subtracting two adjacent previous rounds' global models. Thus the delta weights attack is easier 
to detect.

\begin{figure}[hbtp]
     \centering
     \begin{subfigure}[b]{\linewidth}
         \centering
         \includegraphics[width=\textwidth]{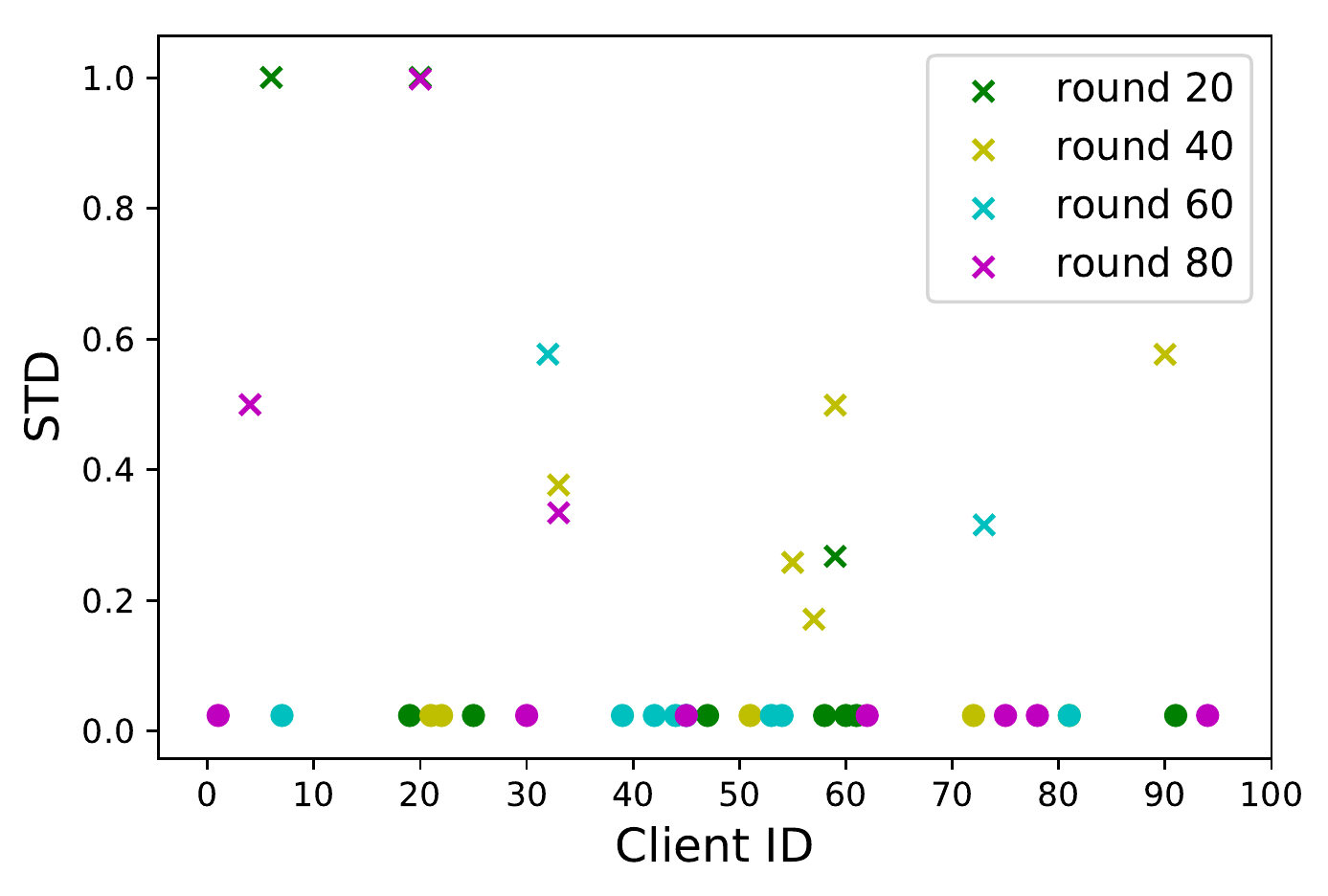}
         \caption{Standard deviation (STD) of the updates submitted by each client at different rounds}
         \label{fig:DP-std}
     \end{subfigure}
     \begin{subfigure}[b]{0.48\textwidth}
         \centering
         \begin{subfigure}[b]{.49\textwidth}
             \includegraphics[width=\linewidth]{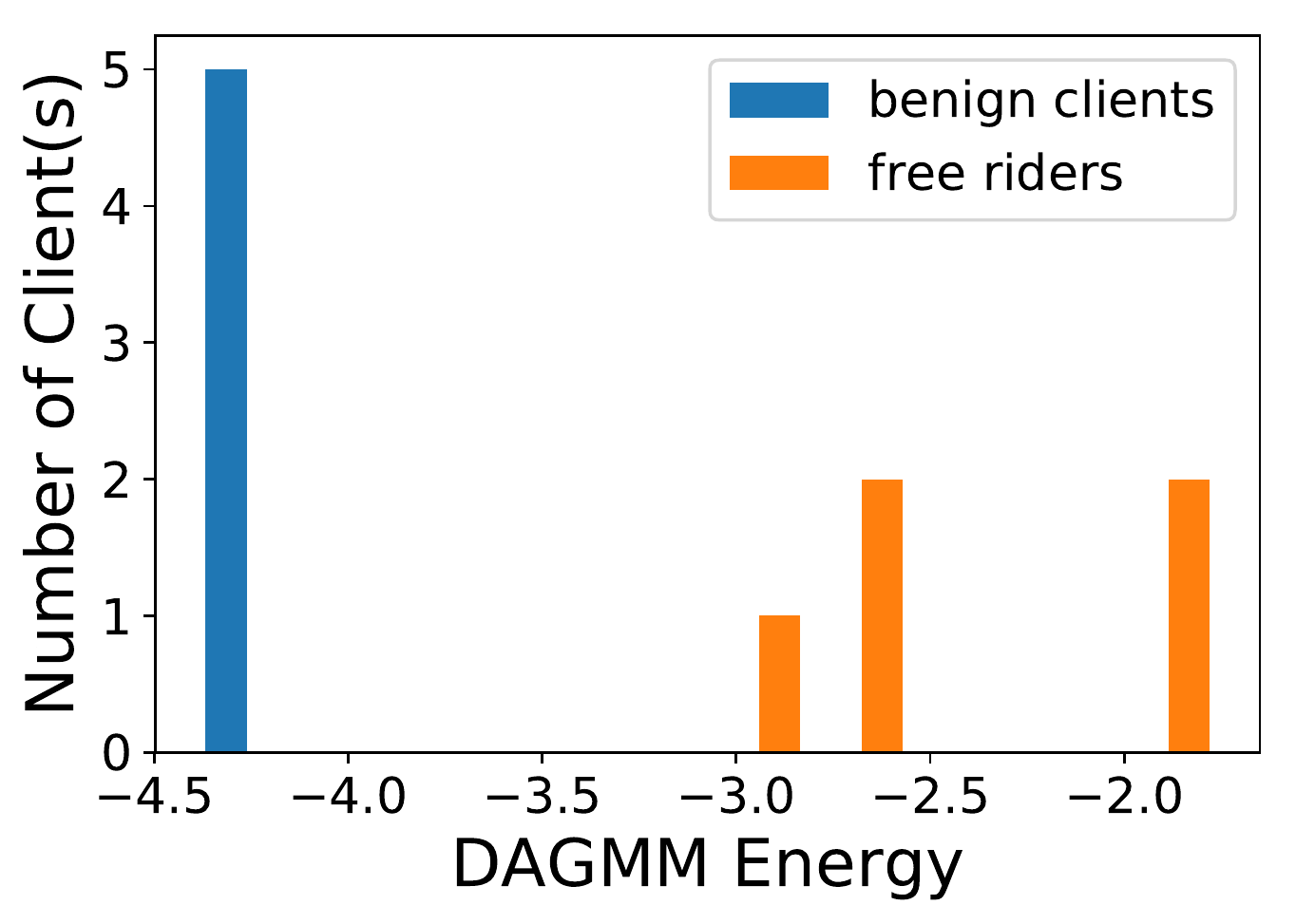}\hfill%
             \caption{Round 40}
         \end{subfigure}
         \begin{subfigure}[b]{.49\textwidth}
             \includegraphics[width=\linewidth]{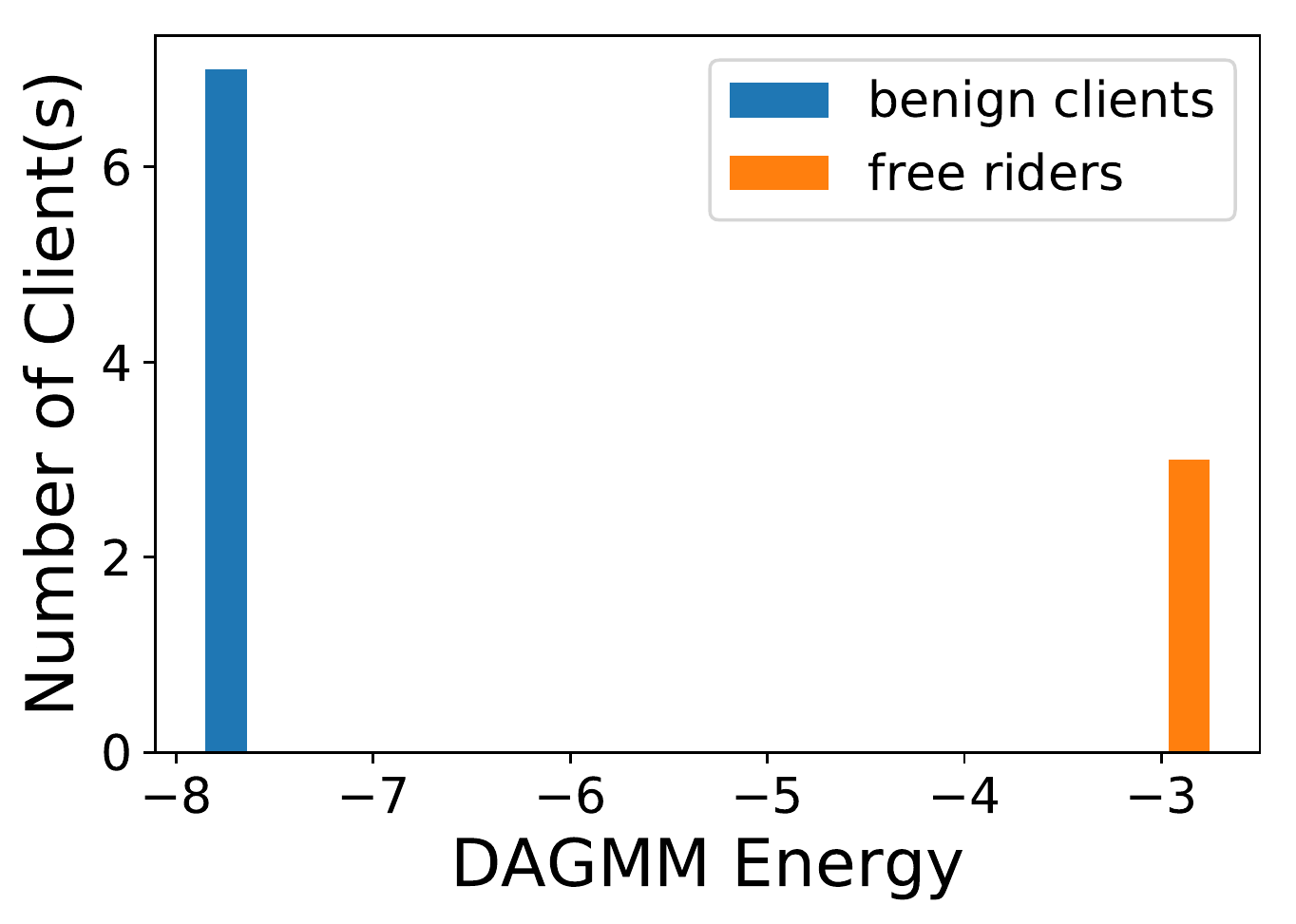}
             \caption{Round 80}
         \end{subfigure}
         \caption{DAGMM - detected}
         \label{fig:DP-DAGMM}
     \end{subfigure}
     \begin{subfigure}[b]{\linewidth}
         \centering
         \begin{subfigure}[b]{.49\linewidth}
             \includegraphics[width=\linewidth]{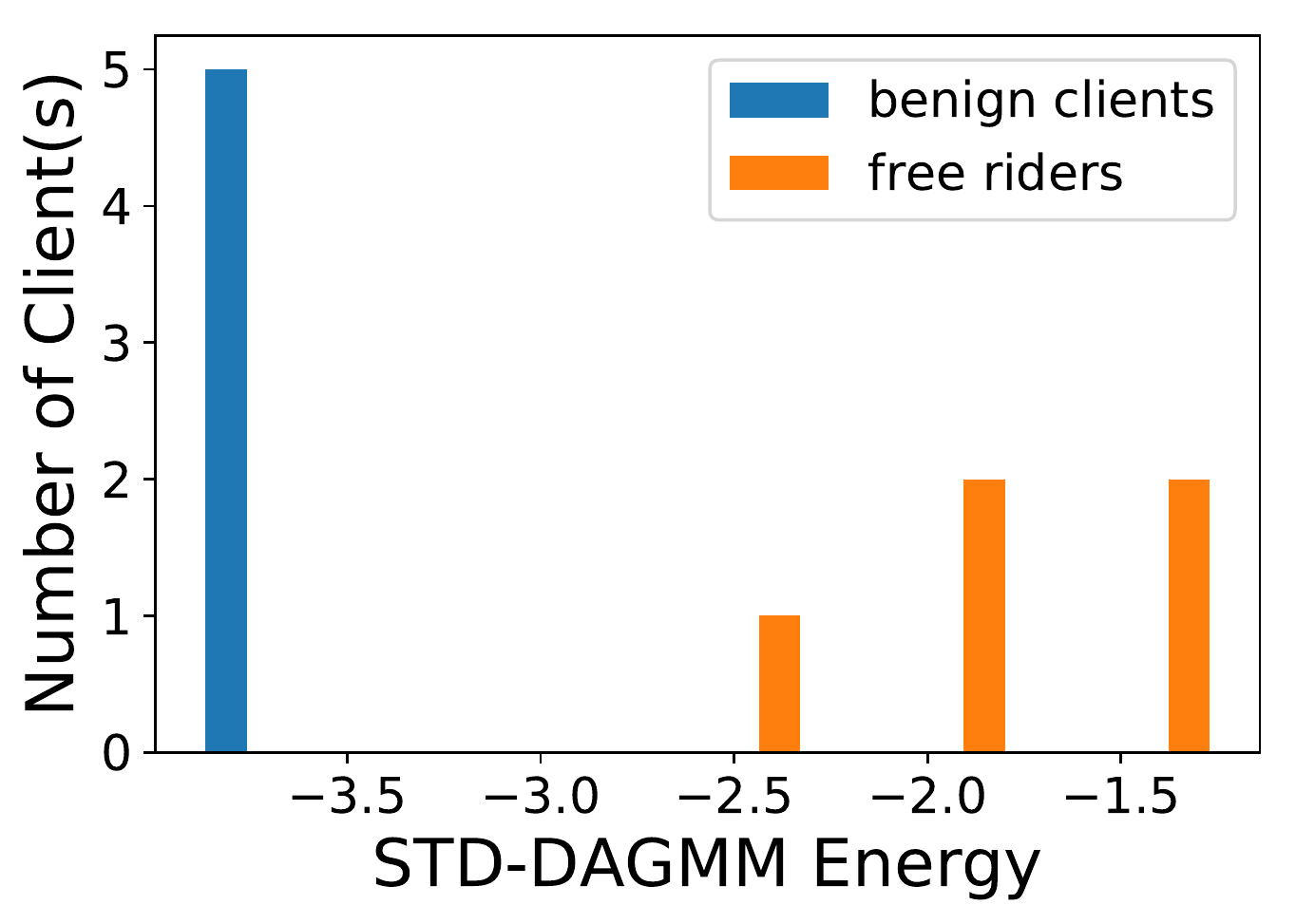}\hfill%
             \caption{Round 40}
         \end{subfigure}
         \begin{subfigure}[b]{.49\linewidth}
             \includegraphics[width=\linewidth]{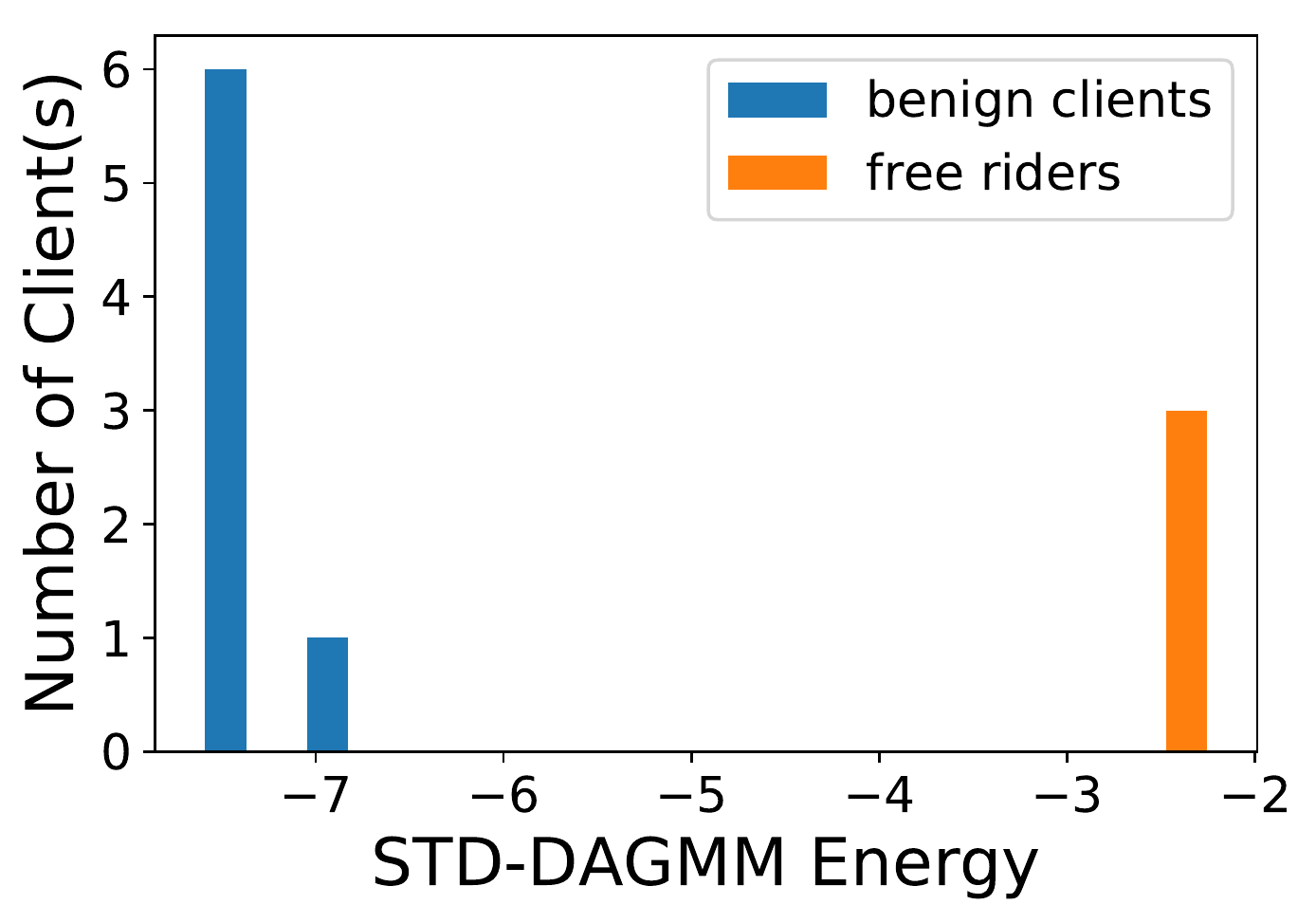}
             \caption{Round 80}
         \end{subfigure}
         \caption{STD-DAGMM - detected}
         \label{fig:STD-DP-DAGMM}
     \end{subfigure}
     \caption{With differential privacy, similar local data distribution, delta weights attack, 20 free riders}
\end{figure}

For this experiment, we assume there are 20 free riders and they randomly 
join a training round with other clients. 
For detection, we deliberately skip the first several rounds to make sure that free riders have accumulated enough knowledge (two global models) to generate a delta weight update. We perform the detection at round 20, 40, 60, and 80 respectively, Figure~\ref{fig:DP-std} shows that the STD of free riders (marked as crosses) are much higher than that of normal clients (marked as dots), and thus should be easier to be detected. Since free riders' updates are an aggregation of small gradient updates during $q$ rounds, it should be larger than benign clients' gradient updates in magnitude and standard deviation as well. As shown in Figure~\ref{fig:DP-DAGMM} and \ref{fig:STD-DP-DAGMM}, both DAGMM and STD-DAGMM are able to detect all free riders.

\section{Limitation}
Currently, MNIST dataset and a two-layer fully-connected model are used in all of our experiments. Although our attack and detection methods do not depend on the actual training data and model architecture, we admit that there are a large variety of datasets and models that may be used under the framework of federated learning. We leave it as future work to exploit the performance of different attack and defense methods on other datasets and models, e.g., 
advanced language models on natural language datasets.
Moreover, recent papers have proposed new paradigms for federated learning, for example, secure aggregation, which may affect the attack and defense strategies proposed here. For example, in secure aggregation where the central parameter server cannot see the updated matrices of the clients in plain texts, free rider attack detection could be challenging. More research work could be done targeting different federated learning solutions.
\section{Related work}



Since proposed in~\cite{DBLP:journals/corr/McMahanMRA16}, federated learning has 
attracted many research to make it more practical and more robust in real world usages.
Some of the schemes proposed for federated learning could potentially affect a free rider's weights generation strategy, or the effectiveness of the detection mechanism.

There have been many previous work proposed to enhance federated learning, e.g., to make it more communication efficient \cite{DBLP:journals/corr/KonecnyMYRSB16, chen2019communication}, or privacy-preserving \cite{DBLP:journals/corr/abs-1712-07557, Bonawitz:2017:PSA:3133956.3133982}.
The communication cost in federated learning could be reduced by reducing the size of updates, in ways such as cutting down decimal digits in each value of the model update \cite{DBLP:journals/corr/KonecnyMYRSB16, chen2019communication}.
This mechanism could potentially cause free riders easier to deceive since there are fewer digits to manipulate.
To address the privacy concerns~\cite{DBLP:journals/corr/HitajAP17, DBLP:journals/corr/abs-1812-00535}, client-level differential privacy \cite{DBLP:journals/corr/abs-1712-07557} is proposed to prevent leaking local clients' private data from global model parameters, and a secure aggregation mechanism is proposed for federated learning in \cite{Bonawitz:2017:PSA:3133956.3133982}.
For example, if secure aggregation \cite{Bonawitz:2017:PSA:3133956.3133982} is applied, where the parameter server is not supposed to view locally submitted model parameters, advances in cryptographic fields might be required to address the free rider attack problem.



Although the concept of free rider attack has not been explored in previous work, there have been related work on other attacks and defenses targeting federated learning domain.
For starters, \cite{Shen:2016:UDA:2991079.2991125} proposes a method to defend against poisoning attacks in federated learning, through clustering the local clients' updates and treating the majority as normal. This detection does not work for the delta weights attack in our setting because the attacker submitted updates are similar to those of normal clients.
Another type of backdoor attack is proposed in \cite{DBLP:journals/corr/abs-1807-00459}, where the attacker submits constructed large weights to overwrite the global model's parameter, such that it predicts a certain label for some images with special patterns.
They find that client-level DP~\cite{DBLP:journals/corr/abs-1710-06963} could degrade the effectiveness of such attack. Also, we note that our STD-DAGMM method is potentially effective since an attack constructed this way would have much larger STD than other normal clients.
Poisoning attacks for federated learning are further explored in~\cite{DBLP:journals/corr/abs-1811-12470}, where the adversarial objective is to cause the model to mis-classify chosen inputs with high confidence.
While no defense mechanisms are presented in the paper, and it will be interesting to explore whether our STD-DAGMM method could defend against the proposed attacks.
\cite{DBLP:journals/corr/abs-1808-04866} explores sybil-based label-flipping and backdoor poisoning attacks in federated learning,
and proposes to use cosine similarity as a detection metric to adjust the weight for each client update. However, 
such defense wouldn't work for our delta weights attack, since free riders' 
gradient updates would not show a big difference with other clients in terms of angles in weight updates.



\section{Conclusion}
In this paper, we present a potential problem in federated learning, that a client having no local data may construct a local gradient update, in order for rewards. We study different attacks an attacker may take, and possible detection mechanisms against such attacks. In particular, we propose a novel detection scheme: STD-DAGMM, that is effective in detecting most of the free riders in most explored settings. Further, we find that differential privacy, particularly the privacy amplification mechanism in federated learning, favors the detection of free riders.
For future work, the proposed STD-DAGMM method shows its effectiveness in anomaly detection for model weights, so studying its effectiveness in detecting other attacks for federated learning, e.g., poisoning attacks, is one interesting direction to pursue.
Finally, we note that there are much more to investigate in the field of free rider attacks and defenses, especially with constantly proposed federated learning enhancing strategies. 
We believe our initial exploration may shed light on future work in this area.

\bibliographystyle{IEEEtran}
\bibliography{ref}

%

\begin{IEEEbiography}[{\includegraphics[width=1in,height=1.25in,clip,keepaspectratio]{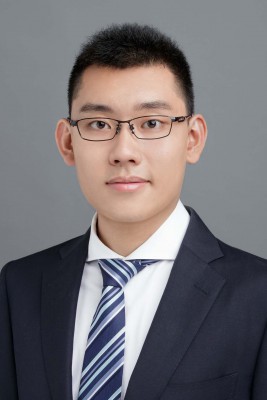}}]{Jierui Lin}
is a fourth-year undergraduate studying Computer Science and Applied Mathematics at UC Berkeley. His research interests include computer vision, robotics and machine learning security.
\end{IEEEbiography}

\begin{IEEEbiography}[{\includegraphics[width=1in,height=1.25in,clip,keepaspectratio]{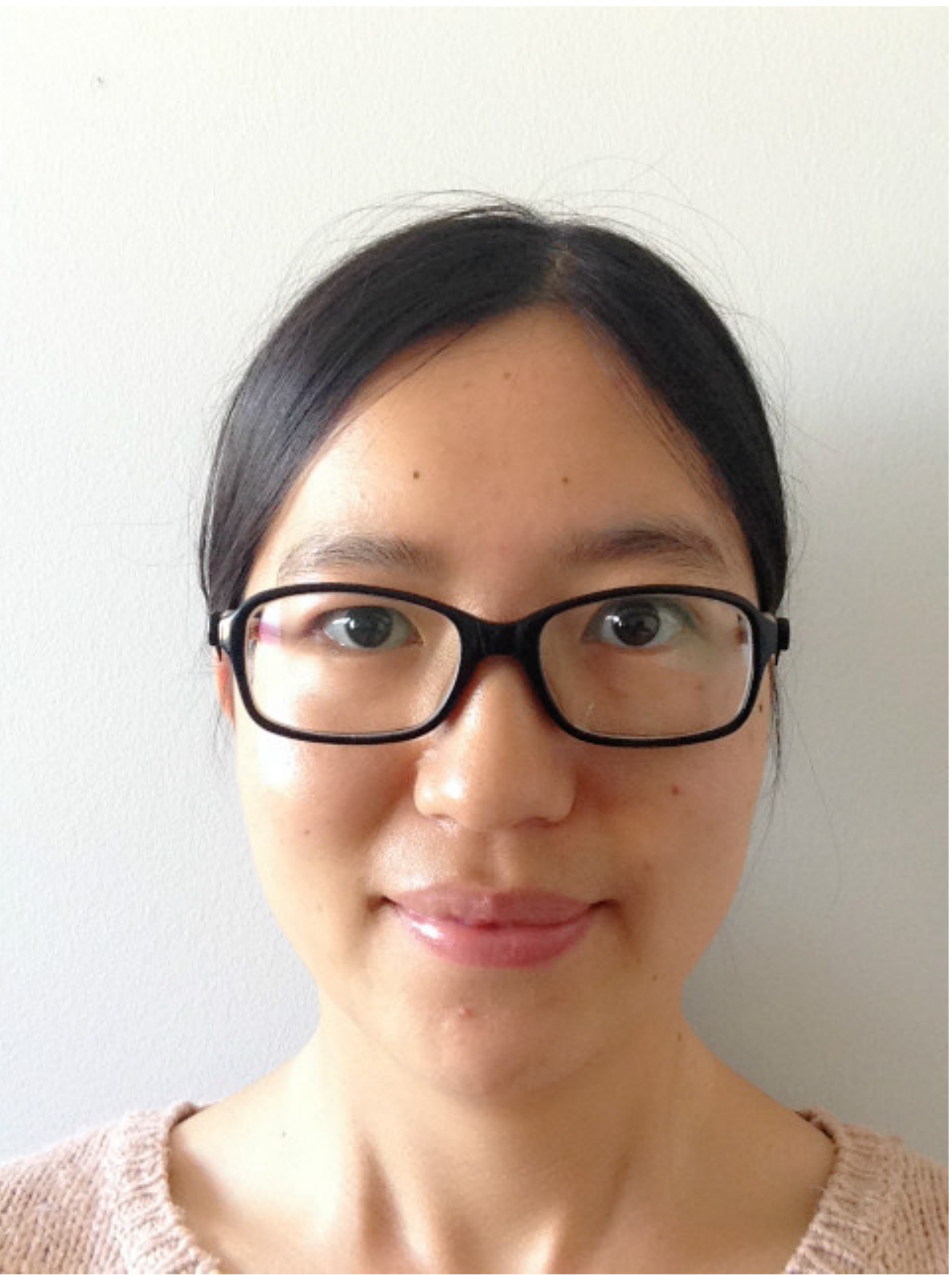}}]{Min Du}
received the PhD degree from the
School of Computing, University of Utah in 2018,
as well as the bachelor’s degree and the master's degree from Beihang University. She is currently a Postdoctoral scholar in EECS department, UC Berkeley. Her research interests include big data analytics and machine learning security.
\end{IEEEbiography}

\begin{IEEEbiography}[{\includegraphics[width=1in,height=1.25in,clip,keepaspectratio]{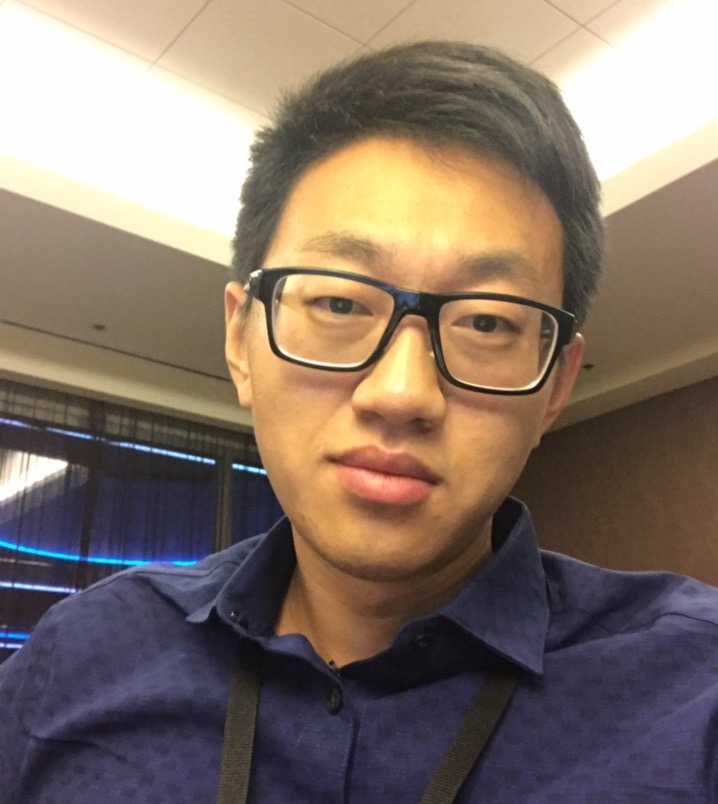}}]{Jian Liu}
has been a postdoctoral researcher since July 2018, at Department of Electrical Engineering and Computer Sciences , UC Berkeley, working with Prof. Dawn Song. Before joining UC Berkeley, he was a PhD candidate since the fall of 2014, at the Secure Systems Group, Department of Computer Science, Aalto University, working with Prof. N. Asokan.
His research is on Applied Cryptography and Blockchains. He is interested in building real-world systems that are provably secure, easy to use and inexpensive to deploy.
\end{IEEEbiography}








\end{document}